%% file: main.tex
\documentclass{article}
\ifdefined\pdfobjcompresslevel
\fi
\usepackage{iclr2027_conference,times}
\usepackage[utf8]{inputenc}
\usepackage[T1]{fontenc}
\usepackage{microtype}
\usepackage{graphicx,booktabs,array,makecell,multirow,wrapfig,tabularx,float}
\usepackage{enumitem,capt-of}
\usepackage{amsmath,amssymb,amsfonts,mathtools,amsthm}
\usepackage{xcolor}
\usepackage{colortbl}
\definecolor{codeblue}{HTML}{215CAF}
\usepackage{hyperref,url}
\usepackage[capitalize,noabbrev]{cleveref}
\let\cite\citep
\makeatletter
\let\spinsci@makecaption\@makecaption
\newcommand{\spinsciwrapcaption}{%
  \vspace{-\intextsep}%
  \setlength{\abovecaptionskip}{0pt}%
  \setlength{\belowcaptionskip}{4pt}%
  \long\def\@makecaption##1##2{%
    {\footnotesize\spinsci@makecaption{##1}{##2}}}%
}
\makeatother
\newtheorem{theorem}{Theorem}[section]
\newtheorem{proposition}[theorem]{Proposition}

\theoremstyle{remark}
\newtheorem{remark}[theorem]{Remark}
\iclrfinalcopy
\title{\raggedright SpinSCI: Mask-Conditioned Spatial-Temporal\\Spin Transformer for Video Snapshot\\Compressive Imaging}
\author{Seungki Hong\qquad Zhishuai Zhang\qquad Taekwang Jang\\
ETH AI Center\\
ETH Zurich, Switzerland}
\hypersetup{pdfborder={0 0 0.5},pdftitle={},pdfauthor={},pdfsubject={},pdfkeywords={},
  pdfcreator={},pdfproducer={}}
\ifdefined\pdfinfoomitdate\pdfinfoomitdate=1\fi
\ifdefined\pdfsuppressptexinfo\pdfsuppressptexinfo=15\fi
\ifdefined\pdftrailerid\pdftrailerid{}\fi
\begin{document}
\maketitle
\fancyhead[L]{Preprint}
\begin{abstract}
Video snapshot compressive imaging (SCI) recovers multiple frames from one spatiotemporally coded measurement. This paper proposes \textit{SpinSCI}, a mask-conditioned Spatial-Temporal Spin Transformer whose \textit{Mask-Conditioned Ising Refinement Block (MC-IRB)} combines attention logits and sensing information to form signed spin couplings. Finite mean-field updates gate residual feature refinement. Physics-guided initialization and final data consistency incorporate the measurement model. Compared with HiSViT$_9$, SpinSCI achieves average PSNR gains of \textbf{0.51~dB}, \textbf{0.23~dB}, and \textbf{0.04~dB} on \textit{small-scale gray, mid-scale color, and large-scale color} benchmarks, respectively, with \textbf{16.9\% fewer parameters, 21.0--21.6\% fewer FLOPs, 16.1--49.0\% lower latency, and 59.9--62.7\% lower peak memory}. Relative to DU-9stg, average PSNR differs by at most \textbf{0.13~dB} across these benchmarks, with \textbf{28.2--28.7\% fewer FLOPs, 64.3--71.9\% lower inference latency, and 30.1--32.7\% lower peak memory}. Qualitative comparisons on the \textit{real benchmark} further show recovery of fine details and textures. Code and reference checkpoints are available at \href{https://github.com/skethz/SpinSCI}{\textcolor{codeblue}{github.com/skethz/SpinSCI}}.

\end{abstract}

\section{Introduction}
\begin{figure}[H]
    \centering
    \vspace{-1em}
    \includegraphics[width=1.0\columnwidth]{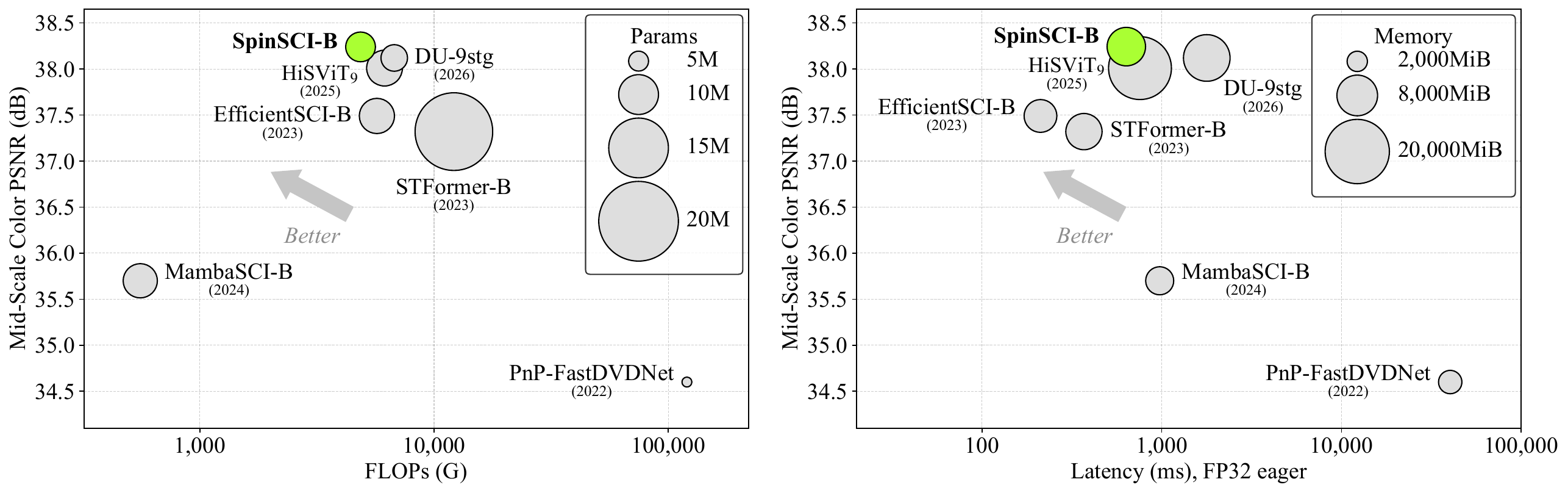}
    \vspace{-2em}
    \caption{Mid-scale color PSNR versus FLOPs (left) and FP32 eager inference latency (right) for all methods (\Cref{tab:color}). Circle diameter represents parameter count on the left and circle area represents peak allocated memory on the right.}
    \label{fig:efficiency}
\end{figure}

High-speed imaging captures rapid motion but faces bandwidth, storage, and cost constraints~\cite{intro}. Video snapshot compressive imaging (SCI) encodes multiple frames into one measurement using time-varying masks, as in Coded Aperture Compressive Temporal Imaging (CACTI)~\cite{cacti}. Reconstruction is therefore a severely underdetermined inverse problem related to compressed sensing~\cite{cs1,cs2}.

Prior model-based and learned methods~\cite{desci,e2ecnn,metasci,pnp,efficient,stformer,res2former,effipp,mamba,hisvit} improve reconstruction, yet spatial masking and temporal integration complicate detail recovery and frame separation~\cite{hisvit}. In attention-based reconstruction, logits encode feature affinity, while coding masks specify each frame's contribution to the measurement. These cues motivate token refinement conditioned on sensing information.

SpinSCI coordinates this refinement through \textit{Ising-model-inspired} interactions. The \textit{Mask-Conditioned Ising Refinement Block (MC-IRB)} derives signed couplings from pre-softmax attention logits and sensing inputs: the coding masks, normalized initial estimate, and mask normalizer. Local~fields encode each token's evidence; positive and negative couplings favor agreement and opposition between latent spins. Finite mean-field updates produce magnetizations that gate learned residual corrections to the attention output.

\begin{figure}[!t]
    \centering
    \includegraphics[width=0.80\columnwidth]{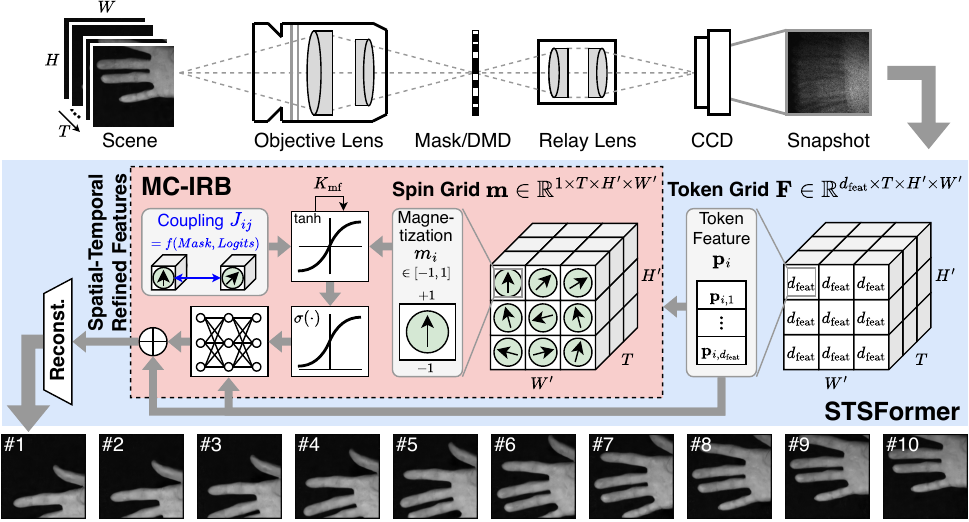}
    \vspace{-1em}
    \caption{Illustration of the SpinSCI video SCI pipeline. A single snapshot encoded by CACTI~\cite{cacti} is input to SpinSCI. In the Spatial-Temporal Spin Transformer (STSFormer), token features $\mathbf{p}_i$ are extracted, and the Mask-Conditioned Ising Refinement Block (MC-IRB) performs differentiable mean-field inference for $K_{\mathrm{mf}}$ steps to estimate magnetizations $m_i\!\in\![-1,1]$, which are used to refine the spatial and temporal attention features. After post-processing, the video is reconstructed. Here, $d_{\mathrm{feat}}=256$ denotes the token feature width. $\#$ denotes the frame number and a representative example from the Hand dataset~\cite{pnp} ($512\times512\times40$) in the \textbf{real benchmark} is used at a compression rate of $10$.}
    \label{fig:intro}
\end{figure}

MC-IRB refines the parallel spatial and temporal branches of a \textit{Spatial-Temporal Spin Transformer (STSFormer)}. Physics-guided initialization and final data consistency incorporate the sensing model (\Cref{fig:intro}).

This work makes the following key contributions:
\begin{itemize}[nosep, leftmargin=*]
  \item SpinSCI introduces MC-IRB, which uses \textit{mask-conditioned positive and negative couplings} to favor agreement and opposition between latent spins, respectively. Finite mean-field inference converts these interactions into spatial and temporal gates that guide residual feature corrections; removing the couplings or their signs lowers PSNR (\Cref{sec:ablation}).
  \item SpinSCI achieves competitive reconstruction quality on the \textit{small-scale gray, mid-scale color, and large-scale color benchmarks} and the \textit{real benchmark}, with \textit{low parameter counts, FLOPs, inference latency, and memory usage} (\Cref{fig:efficiency}).
\end{itemize}

\section{Related Work}\label{sec:related_work}

\noindent\textbf{Vision Transformers.} Self-attention~\cite{transformer} extends to images through ViT~\cite{vit} and Swin~\cite{swin}, and to video through local windows~\cite{videoswin} or space--time factorization~\cite{timesformer}. Building on non-local similarity~\cite{nonlocal2005,nonlocal2007,nonlocal2014}, STFormer~\cite{stformer} separates spatial and temporal attention into parallel branches. HiSViT~\cite{hisvit} uses sequential cross-scale attention with pooled keys/values, hierarchical branches, and a gated self-modulated feed-forward network (FFN). SpinSCI combines parallel branches and pooled-key/value attention with \textit{mask-conditioned spin interactions for residual feature refinement.}

\noindent\textbf{Mask-Aware Reconstruction and Mask Co-Design.} MetaSCI~\cite{metasci} uses exposure-normalized, mask-modulated inputs and per-mask parameter adaptation. Res2former~\cite{res2former} jointly learns structural masks and reconstruction to improve dynamic range under temporal multiplexing. SpinSCI requires neither per-mask parameter updates nor physical-mask optimization. It conditions refinement on the supplied mask, normalized initial estimate, and mask normalizer to \textit{guide mask-aware feature refinement during reconstruction.}

\noindent\textbf{Deep Unfolding and Further Related Work.} Differential Unfolding (DU)~\cite{du} alternates full reconstruction stages with lightweight stages that refine changes in intermediate representations; \Cref{tab:hisvit-measured} compares its published scores with our evaluation of the released nine-stage checkpoint. Appendix~\ref{app:related} discusses deep unfolding, other learned video SCI methods, unrolled mean-field inference, energy-based, signed, and gated attention, and mask-conditioned reconstruction, and states how SpinSCI differs from each.

\section{Method}\label{sec:method}

\subsection{Overview}\label{subsec:overview}
\noindent\textbf{Video SCI.} A snapshot $\mathbf{y}\in\mathbb{R}^{H\times W}$ encodes $T$ frames $\mathbf{x}_t\in\mathbb{R}^{H\times W}$ through masks $\mathbf{\Phi}_t$:
\begin{equation}
    \mathbf{y} 
    = \sum_{t=1}^{T} \mathbf{\Phi}_t \odot \mathbf{x}_t + \mathbf{n},
    \label{eq:sci_forward_overview}
\end{equation}
Here, $H,W$ are the spatial dimensions, $t$ indexes frames, $\odot$ denotes element-wise multiplication, and $\mathbf{n}$ is measurement noise.

\begin{figure}[!t]
    \centering
    \includegraphics[width=0.86\columnwidth]{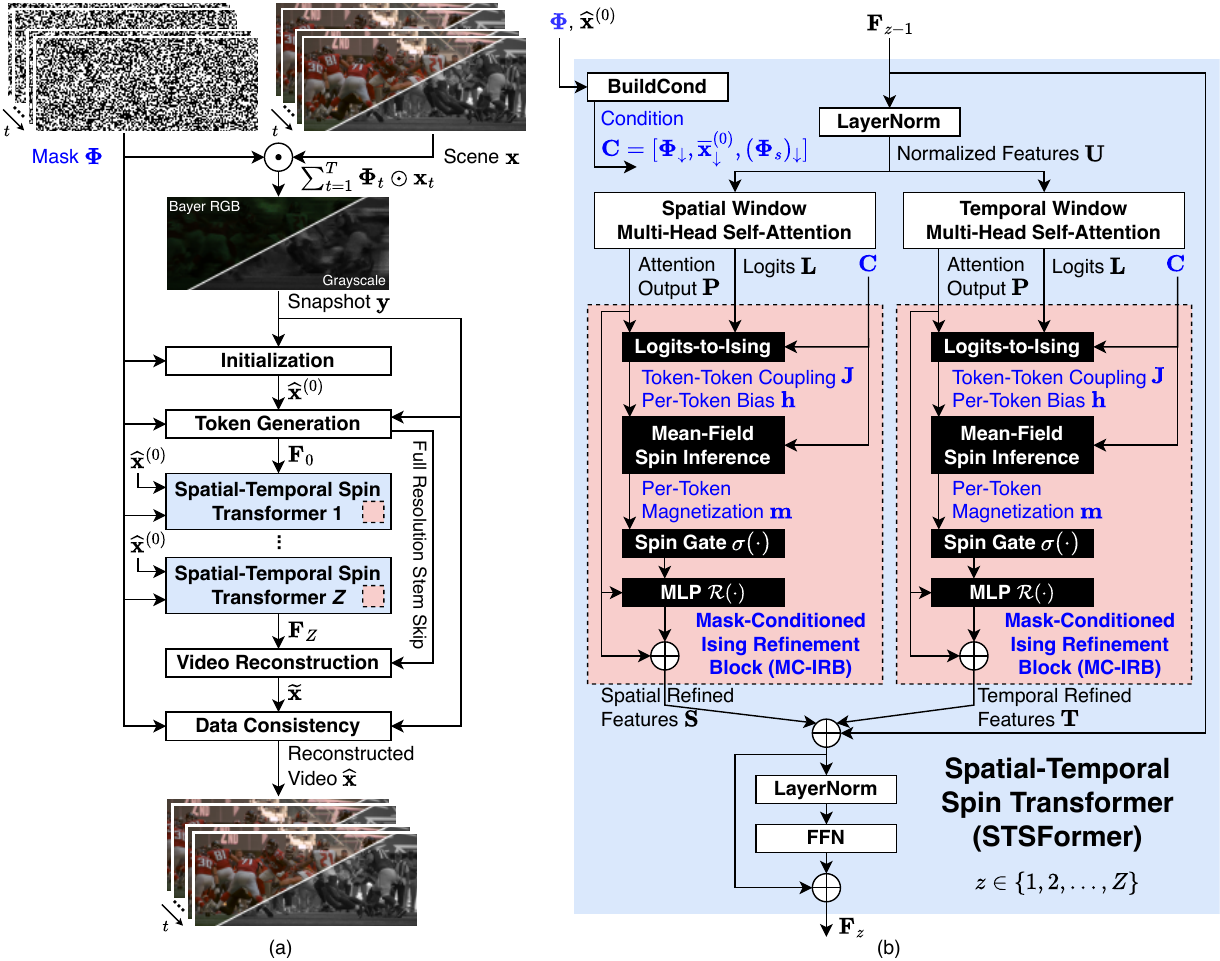}
    \vspace{-1em}
    \caption{
    Overall architecture of SpinSCI. (a) The snapshot $\mathbf{y}$ and coding masks $\mathbf{\Phi}$ provide a physics-guided initialization $\widehat{\mathbf{x}}^{(0)}$. Token generation is followed by $Z$ STSFormer blocks and video reconstruction, with a full-resolution stem skip supplied to the decoder. A final data-consistency correction maps the decoder output $\widetilde{\mathbf{x}}$ to the reconstructed video $\widehat{\mathbf{x}}$. (b) Spatial and temporal attention operate in parallel. MC-IRB uses the shared conditioning tensor $\mathbf{C}$ and attention logits $\mathbf{L}$ to form signed spin interactions and generate gates for residual feature corrections. The spatial branch combines native and pooled-key/value heads; only native-resolution heads supply its MC-IRB logits. The refined branch outputs are combined with the input features, followed by a residual feed-forward module.
    }
    \label{fig:arch}
\end{figure}

\noindent\textbf{Sensing and Learned Refinement.} SpinSCI uses a learned prior to resolve temporal ambiguity. Initialization and final correction act in the measurement domain; MC-IRB conditions feature refinement on $\mathbf{\Phi}$, the normalized initial estimate $\overline{\mathbf{x}}^{(0)}$, and mask normalizer $\mathbf{\Phi}_s$. Throughout, $\epsilon=10^{-8}$.

The initial estimate $\widehat{\mathbf{x}}^{(0)}$ and per-pixel mask normalizer $\mathbf{\Phi}_s$ are constructed as described in \Cref{subsec:architecture}. MC-IRB follows four steps: \textit{conditioning, signed-coupling construction, mean-field inference, and residual gating}. \Cref{subsec:mcirb,subsec:architecture} define the refiner and its backbone integration (\Cref{fig:arch}).

\subsection{Mask-Conditioned Ising Refinement Block (MC-IRB)}\label{subsec:mcirb}

Following the Ising model of statistical physics, each post-attention token $\mathbf{p}_i\in\mathbb{R}^{d_{\mathrm{feat}}}$ carries a latent binary spin. A refinement group contains $n=49$ spatial positions, including padding, or $n=T$ temporal positions. For spins $\mathbf{s}\in\{-1,+1\}^n$, pairwise couplings $J_{ij}$ and local fields $h_i$ define
\begin{equation}
E(\mathbf{s})=-\sum_{i<j} J_{ij}s_is_j-\sum_i h_is_i.
\label{eq:ising-energy}
\end{equation}
The energy is lower when spins joined by a positive coupling take the same sign, spins joined by a negative coupling take opposite signs, and each spin follows the sign of its field. Mean-field inference approximates each spin by its mean, the magnetization $m_i\in[-1,1]$, computed from its field and the magnetizations of its partners. These magnetizations determine refinement gates.

\noindent\textbf{Physical Conditioning (BuildCond).} Per-sample normalization controls the initial-estimate scale:
\begin{equation}
\overline{\mathbf{x}}^{(0)}=
\frac{\widehat{\mathbf{x}}^{(0)}}{\max\!\left(\operatorname{mean}_{t,h,w}|\widehat{\mathbf{x}}^{(0)}|,\epsilon\right)}.
\end{equation}
Nearest-neighbor resizing preserves mask values; bilinear resizing maps the estimate and exposure normalizer to the token grid. After resizing, the normalizer is repeated across frames and concatenated with the mask and normalized initial estimate to form
\begin{equation}
\mathbf{C}=\left[\mathbf{\Phi}_{\downarrow},\overline{\mathbf{x}}^{(0)}_{\downarrow},(\mathbf{\Phi}_s)_{\downarrow}\right],\qquad
\mathbf{c}_i=\left[\phi_i,u_i,d_i\right]\in\mathbb{R}^{3},
\end{equation}
Here, $\downarrow$ denotes token-grid resizing, $u_i=\bar{x}^{(0)}_i$, and $d_i=\phi_{s,i}$. Shared across blocks, $\mathbf{C}$ supplies each token with its coding-mask value, normalized initial estimate, and spatial exposure normalizer.

\noindent\textbf{Logits-to-Ising.} The two native spatial heads or all four temporal heads form $\mathcal{H}$. The head-averaged logits are
\begin{equation}
\overline{\mathbf{L}}=\frac{1}{|\mathcal{H}|}\sum_{a\in\mathcal{H}}\mathbf{L}^{(a)}.
\end{equation}
Reusing pre-softmax logits ties refinement to affinities already computed for attention. Differences in sensing information define a penalty applied before centering:
\begin{equation}
\Pi_{ij}=\lambda_{\phi}\lvert\phi_i-\phi_j\rvert
+\lambda_u\lvert u_i-u_j\rvert
+\lambda_d\lvert d_i-d_j\rvert.
\label{eq:sensing-penalty}
\end{equation}
For valid-pair indicators $v_{ij}=v_{ji}\in\{0,1\}$ with $v_{ii}=1$, conditioned logits and row means are
\begin{equation}
\widetilde{J}_{ij}=v_{ij}(\overline{L}_{ij}-\Pi_{ij}),\qquad
n_i=\sum_jv_{ij},\qquad
\mu_i=\frac{1}{n_i}\sum_j\widetilde{J}_{ij}.
\end{equation}
Centering measures relative affinity; remasking, symmetrization, and diagonal removal produce valid, reciprocal interactions without self-coupling:
\begin{equation}
J_{ij}^{\mathrm{ctr}}=v_{ij}(\widetilde{J}_{ij}-\mu_i),\qquad
J_{ij}=\gamma v_{ij}\mathbf{1}[i\ne j]\frac{J_{ij}^{\mathrm{ctr}}+J_{ji}^{\mathrm{ctr}}}{2}.
\label{eq:rowcenter}
\end{equation}
The learned weights $\lambda_\phi,\lambda_u,\lambda_d$ and scale $\gamma$ are clamped nonnegative. Diagonal entries enter the row means before removal. The resulting signed $\mathbf{J}$ is symmetric, with zeros on the diagonal and at invalid token pairs; its rows need not sum to zero.

\noindent\textbf{Mean-Field Spin Inference.} Learned affine projections encode token and sensing evidence into initial magnetization and local field:
\begin{equation}
m_i^{(0)}=\tanh\!\left(\mathbf{w}_m^\top[\mathbf{p}_i;\mathbf{c}_i]+b_m\right),\qquad
h_i=\mathbf{w}_h^\top[\mathbf{p}_i;\mathbf{c}_i]+b_h,
\end{equation}
Here, $[\cdot\,;\cdot]$ concatenates inputs, $\mathbf{w}_h,\mathbf{w}_m\in\mathbb{R}^{d_{\mathrm{feat}}+3}$, and $b_h,b_m\in\mathbb{R}$. The field retains unary evidence at each update; learned initialization supplies an input-dependent starting state for finite-step inference.

With $\mathbf{h}$ and $\mathbf{J}$ fixed, $K_{\mathrm{mf}}$ updates propagate interactions:
\begin{equation}
m_i^{(\ell+1)}=\tanh\!\left(\beta\left[h_i+\sum_jJ_{ij}m_j^{(\ell)}\right]\right),\qquad \ell=0,\ldots,K_{\mathrm{mf}}-1,
\end{equation}
The learned $\beta\ge0$ is clamped nonnegative. Positive couplings favor agreement between latent spins; negative couplings favor opposite signs. Invalid pairs contribute zero, and $\tanh$ bounds every iterate to $[-1,1]$. The fixed update budget coordinates gates through pairwise evidence with predictable computation.

\noindent\textbf{Spin Gate and Residual MLP.} Final magnetizations control correction strength through
\begin{equation}
g_i=\sigma\!\left(\kappa m_i^{(K_{\mathrm{mf}})}\right),\qquad
\sigma(-\kappa)\le g_i\le\sigma(\kappa),
\end{equation}
where $\sigma$ is sigmoid and the learned $\kappa\ge0$ is clamped nonnegative. The residual MLP $\mathcal{R}$ uses LayerNorm--Linear--GELU--Linear with $64$ hidden units. Zero initialization of its final affine layer makes MC-IRB initially the identity:
\begin{equation}
\mathbf{p}_i \leftarrow \mathbf{p}_i + g_i\,\mathcal{R}(\mathbf{p}_i).
\end{equation}
The MLP supplies a learned feature correction, and the gate regulates its strength. Within each branch, signed spin interactions make the gates of tokens in the same group depend on one another (\Cref{sec:ablation}). Residual addition preserves the attention output as the base. The correction magnitude $g_i\|\mathcal{R}(\mathbf{p}_i)\|$ depends on both gate and residual; $m_i=0$ gives $g_i=1/2$.

Appendix~\ref{app:formal} provides sensing and refinement proofs and specifies Lean verification coverage. Appendix~\ref{app:ablation-diagnostics} shows that, in a trained temporal block, the magnetizations nearly minimize the pairwise term of \Cref{eq:ising-energy}.

\subsection{SpinSCI Architecture}\label{subsec:architecture}
SpinSCI applies token generation, $Z=9$ STSFormer blocks, decoding, and final data consistency to produce $\widehat{\mathbf{x}}$ from $\widehat{\mathbf{x}}^{(0)}$.

\noindent\textbf{Initialization.} For $\mathcal{A}(\mathbf{x};\mathbf{\Phi})=\sum_t\mathbf{\Phi}_t\odot\mathbf{x}_t$, an exposure-aware back-projection initializes the video:
\begin{equation}
\begin{aligned}
\widehat{\mathbf{x}}^{(0)}&=\mathcal{A}^{\top}(\mathbf{y};\mathbf{\Phi})
=\{\mathbf{\Phi}_t\odot\mathbf{y}\}_{t=1}^{T},\\
\widehat{\mathbf{x}}^{(0)}&\leftarrow\widehat{\mathbf{x}}^{(0)}+
\mathcal{A}^{\top}\!\left(\frac{\mathbf{y}-\mathcal{A}(\widehat{\mathbf{x}}^{(0)};\mathbf{\Phi})}{\mathbf{\Phi}_s+\epsilon};\mathbf{\Phi}\right).
\end{aligned}
\end{equation}
The normalizer $\mathbf{\Phi}_s$ is $\sum_t\mathbf{\Phi}_t$ with zeros replaced by one; it equals mask energy at binary-mask exposed pixels. Back-projection and exposure-normalized correction supply a sensing-informed starting point for learned reconstruction. Bayer initialization applies this operation independently on the four RGGB sub-grids before re-interleaving.

\noindent\textbf{Token Generation.} A 3D stem embeds $[\widehat{\mathbf{x}}^{(0)},\mathbf{\Phi},\mathbf{y}/(\mathbf{\Phi}_s+\epsilon)]$, with the normalized measurement repeated across frames, into $128$-channel features. Spatial downsampling halves resolution and sets $d_{\mathrm{feat}}=256$:
\begin{equation}
\mathbf{F}_0\in\mathbb{R}^{d_{\mathrm{feat}}\times T\times H'\times W'},\qquad H'=\frac{H}{2},\; W'=\frac{W}{2}.
\end{equation}
Joint embedding exposes coding and measurement information before attention. Downsampling reduces attention cost while preserving temporal length; full-resolution stem features supply a decoder skip. One $3\times3\times3$ convolution in each of the stem, downsampler, and decoder uses rank-$32$ factorization: a $32$-channel convolution followed directly by a $1\times1\times1$ projection.

\noindent\textbf{STSFormer.} Following STFormer~\cite{stformer}, block $z$ supplies $\mathbf{U}=\mathrm{LayerNorm}(\mathbf{F}_{z-1})$ to parallel spatial and temporal branches. The branches model complementary within-frame and across-frame dependencies before feature fusion.

The spatial branch uses four heads with key/value pooling factors $(1,1,2,4)$~\cite{hisvit}. Across successive blocks, both native heads alternate between unshifted $7\times7$ windows and windows with three-token cyclic shifts to exchange information across window boundaries; $\mathbf{M}$ excludes wrap-around pairs~\cite{swin}. For native head $a$,
\begin{equation}
\mathbf{L}^{(a)}=\frac{\mathbf{Q}^{(a)}(\mathbf{K}^{(a)})^\top}{\sqrt{d_h^{\mathrm{sp}}}}+\mathbf{B}^{(a)}+\mathbf{M},\qquad
\mathbf{P}^{(a)}=\mathrm{softmax}(\mathbf{L}^{(a)})\mathbf{V}^{(a)},
\end{equation}
Here, $\mathbf{Q},\mathbf{K},\mathbf{V}$ are query/key/value projections and $\mathbf{B}$ is learned relative-position bias. The other heads retain queries at token-grid resolution and pool keys/values into $8\times8$ windows, broadening context with $64$ keys per query and no cyclic shifts. Concatenated head outputs are projected to $d_{\mathrm{feat}}$. Spatial MC-IRB uses only native-head logits, whose square matrices index the same tokens on both axes for pairwise spin interactions.

Temporal attention acts over $\mathbf{U}_{h,w}\in\mathbb{R}^{T\times d_{\mathrm{feat}}}$, using four heads with total inner width $d_{\mathrm{feat}}/2$ to reduce projection and attention cost. Thus $d_h^{\mathrm{sp}}=64$ and $d_h^{\mathrm{tmp}}=32$, with temporal logits
\begin{equation}
\mathbf{L}^{(a)}=\frac{\mathbf{Q}^{(a)}(\mathbf{K}^{(a)})^\top}{\sqrt{d_h^{\mathrm{tmp}}}}+\mathbf{B}_{\mathrm{tmp}}^{(a)}.
\label{eq:temporal-attention}
\end{equation}
The temporal relative-position bias $\mathbf{B}_{\mathrm{tmp}}^{(a)}\in\mathbb{R}^{T\times T}$ is indexed by frame displacement from $2T-1$ learned values per head; no shifted-window mask applies. Temporal MC-IRB uses logits averaged across all four heads and refines the attention output projected to $d_{\mathrm{feat}}$.

Refined branch outputs $\mathbf{S},\mathbf{T}$ fuse before the residual FFN:
\begin{equation}
\mathbf{F}_{\mathrm{st}}=\mathbf{F}_{z-1}+\mathbf{S}+\mathbf{T},\qquad
\mathbf{F}_{z}=\mathbf{F}_{\mathrm{st}}+\mathrm{FFN}\!\left(\mathrm{LayerNorm}(\mathbf{F}_{\mathrm{st}})\right).
\end{equation}
The inverted residual FFN~\cite{mobilenetv2} expands width to $2d_{\mathrm{feat}}$, applies depthwise $3\times3\times3$ mixing, and projects back, with GELU after expansion and depthwise convolution. Pointwise layers mix channels, while depthwise convolution adds local space--time interactions at low cost.

\noindent\textbf{Video Reconstruction.} A transposed 3D convolution upsamples the final features; concatenation with the stem skip recovers local detail before grayscale/RGB decoding into $\widetilde{\mathbf{x}}$. The reported models are trained with the final reconstruction loss in \Cref{eq:mse-loss} alone, without intermediate supervision.

\noindent\textbf{Data Consistency.} Grayscale reconstruction corrects the decoder's measurement mismatch:
\begin{equation}
\widehat{\mathbf{x}}=\widetilde{\mathbf{x}}+
\mathcal{A}^{\top}\!\left(\frac{\mathbf{y}-\mathcal{A}(\widetilde{\mathbf{x}};\mathbf{\Phi})}{\mathbf{\Phi}_s+\epsilon};\mathbf{\Phi}\right).
\end{equation}
For exposed binary-mask pixels, this multiplies the measurement residual by $\epsilon/(\mathbf{\Phi}_s+\epsilon)$ in exact arithmetic; zero-exposure pixels cannot be corrected.

For RGB output, $\mathcal{B}$ samples RGGB sites and $\mathcal{B}^{\top}$ scatters corrections back:
\begin{equation}
\begin{aligned}
\mathbf{r}&=\mathbf{y}-\mathcal{A}(\mathcal{B}\widetilde{\mathbf{x}}_{\mathrm{rgb}};\mathbf{\Phi}),\\
\widehat{\mathbf{x}}_{\mathrm{rgb}}&=\widetilde{\mathbf{x}}_{\mathrm{rgb}}+
\mathcal{B}^{\top}\mathcal{A}^{\top}\!\left(\frac{\mathbf{r}}{\mathbf{\Phi}_s+\epsilon};\mathbf{\Phi}\right).
\end{aligned}
\end{equation}
Only observed color sites change; other RGB values retain the decoder prediction.

\subsection{Computational Complexity}\label{subsec:complexity}

For $N=TH'W'$ tokens, spatial attention uses $49$ native or $64$ pooled keys per query. Temporal attention uses $T$ keys; each MC-IRB takes $K_{\mathrm{mf}}=3$ updates.

Sensing operations cost $O(THW)$. Fixed-width stem and decoder convolutions cost $O(Nd_{\mathrm{feat}}^2)$.

Attention, residual MLPs, and mean-field interactions give the $Z$-block cost
\begin{equation}
O\!\left(ZN\left[d_{\mathrm{feat}}^2+64d_{\mathrm{feat}}+d_{\mathrm{feat}}(49+64+T)+(1+K_{\mathrm{mf}})(49+T)\right]\right).
\end{equation}
For fixed windows, spatial cost is linear in pixel count, while temporal attention is quadratic in $T$. Counts include padding and full inference, excluding training-only heads; component costs appear in Appendix~\ref{app:implementation-details}.

\section{Experiments}\label{sec:experiments}

\subsection{Experimental Setup}\label{sec:exp_setup}

\noindent\textbf{Datasets.} Training uses DAVIS2017~\cite{davis}. Evaluation covers six \textbf{small-scale gray}, six \textbf{mid-scale color}, and four \textbf{large-scale color} sequences~\cite{pnp,efficient,stformer,mamba}. The \textbf{real benchmark} includes Hand~\cite{pnp}, Water Balloon, Domino~\cite{e2ecnn}, and Pendulum Ball. Dimensions and protocols are in Appendix~\ref{app:implementation-details}.

\noindent\textbf{Evaluation Metrics.} PSNR and SSIM~\cite{ssim} are computed using the evaluation code provided by~\citet{cactilib}; efficiency is assessed by parameter count, FLOPs, inference latency, and peak memory.

\noindent\textbf{Efficiency Measurement.} All methods use FP32 eager inference on an NVIDIA GH200 with batch size~$1$ and $T=8$. Latency is the CUDA-synchronized median for one complete snapshot: one full frame for gray and mid-scale color, and all $24$ tile forwards with output assembly for large-scale color. Peak allocated GPU memory is measured over the same calls. One multiply--accumulate counts as one FLOP. Appendix~\ref{app:compiled} reports the BF16 compiled comparison.

\noindent\textbf{Implementation Details.} PyTorch training uses four NVIDIA H200 GPUs. SpinSCI-B uses the architecture in \Cref{subsec:architecture}, $K_{\mathrm{mf}}=3$, and a $64$-unit residual MLP.

Training uses $200$ epochs, with $256\times256$ gray crops (global batch~$10$) and $512\times512$ color crops (global batch~$2$). Muon~\cite{muon} updates eligible matrix weights and Adam~\cite{adam} the remainder, with cosine learning-rate decay from $2\times10^{-4}$ to $10^{-6}$. The MSE loss~\cite{effipp} is
\begin{equation}\label{eq:mse-loss}
\mathcal{L}_{\mathrm{MSE}}
=\frac{1}{T\,C_{\mathrm{out}}\,H\,W}
\left\lVert\widehat{\mathbf{x}}-\mathbf{x}\right\rVert_F^2,
\end{equation}
where $C_{\mathrm{out}}=1$ for gray and $3$ for RGB. BF16/FP32 mixed precision~\cite{mixed} accelerates training; gradient clipping~\cite{gc} stabilizes optimization.

\subsection{Ablation Studies}
\label{sec:ablation}

Ablations test spatial/temporal refinement, interacting gates, and explicit mask conditioning. Appendix~\ref{app:ablation} provides additional comparisons, control definitions, and the repeated-run protocol.

\begin{wraptable}[10]{r}{0.62\textwidth}
\centering
\spinsciwrapcaption
\caption{Spatial and temporal MC-IRB ablation at $80$ epochs, seed $2447$. Efficiency uses $512\times512$ inputs, $T=8$, and BF16 compiled inference (Appendix~\ref{app:compiled}).}
\label{tab:ablation}
\setlength{\tabcolsep}{1.4pt}
\footnotesize
\begin{tabular*}{\linewidth}{@{\extracolsep{\fill}}ccccccc@{}}
\toprule
\makecell{Spatial\\MC-IRB} & \makecell{Temporal\\MC-IRB} &
\makecell{PSNR\\(dB)} & \makecell{Params\\(M)} &
\makecell{FLOPs\\(G)} & \makecell{Latency\\(ms)} &
\makecell{Memory\\(MiB)} \\
\midrule
- & - & 36.68 & 6.86 & 4528 & 117 & 4280 \\
- & \checkmark & 36.91 & 7.16 & 4690 & 134 & 4282 \\
\checkmark & - & 36.84 & 7.16 & 4694 & 135 & 4282 \\
\checkmark & \checkmark & 37.07 & 7.47 & 4856 & 150 & 4284 \\
\bottomrule
\end{tabular*}
\end{wraptable}

\noindent\textbf{Controlled Protocol.} Variants share DAVIS2017~\cite{davis}, $80$ epochs, batch~$20$, $128\times128$ crops, seed $2447$, loss, optimizer, schedule, initialization, stem skip, and final data consistency. EMA weights are tested on six \textbf{mid-scale color} sequences at compression rate~$8$. All models use nine blocks except the ten-block capacity control.

\begin{table}[!htbp]
\centering
\begingroup
\setlength{\abovecaptionskip}{3pt}
\setlength{\belowcaptionskip}{3pt}
\caption{Complete MC-IRB design comparisons at $80$ epochs.
\Cref{tab:app-confirmatory} reports sample SDs and
paired intervals for selected rows. $\Delta$ is
full-model PSNR minus variant PSNR, in dB. FLOPs follow the THOP convention
in Appendix~\ref{app:implementation-details}; equal counts do not imply
equal executed computation.}
\label{tab:ablation_irb}
\fontsize{8.5}{9.5}\selectfont
\renewcommand{\arraystretch}{0.96}
\setlength{\tabcolsep}{4pt}
\begin{tabular}{@{}lrrrr@{}}
\toprule
Variant & \makecell[r]{PSNR\\(dB)} & $\Delta$ &
\makecell[r]{Params\\(M)} & \makecell[r]{FLOPs\\(G)} \\
\midrule
Full MC-IRB, $K_{\mathrm{mf}}=3$ & 37.07 & -- & 7.47 & 4856 \\
Without MC-IRB & 36.68 & $+0.39$ & 6.86 & 4528 \\
Without MC-IRB, $Z=10$ & 36.87 & $+0.20$ & 7.53 & 4880 \\
\midrule
Residual MLP, fixed $g_i=1/2$ & 36.78 & $+0.29$ & 7.46 & 4851 \\
Conditioned gate without mean-field inference & 36.87 & $+0.20$ & 7.47 & 4856 \\
Sigmoid gate on $\mathbf{h}$ only & 36.84 & $+0.23$ & 7.47 & 4856 \\
$K_{\mathrm{mf}}=0$ & 36.81 & $+0.26$ & 7.47 & 4856 \\
No interactions, $\mathbf{J}=0$ & 36.86 & $+0.21$ & 7.46 & 4853 \\
Unsigned couplings, existing control & 36.93 & $+0.14$ & 7.47 & 4856 \\
Entrywise $|\mathbf{J}|$ control & 36.93 & $+0.14$ & 7.47 & 4856 \\
\midrule
$K_{\mathrm{mf}}=1$ & 36.98 & $+0.09$ & 7.47 & 4856 \\
$K_{\mathrm{mf}}=5$ & 37.07 & 0.00 & 7.47 & 4856 \\
\midrule
No pairwise conditioning in $\mathbf{J}$ & 36.90 & $+0.17$ & 7.47 & 4856 \\
No unary conditioning in $\mathbf{h}$ and $\mathbf{m}^{(0)}$ & 36.89 & $+0.18$ & 7.47 & 4856 \\
Mask penalty off, $\lambda_\phi=0$ & 36.96 & $+0.11$ & 7.47 & 4856 \\
Initial-estimate penalty off, $\lambda_u=0$ & 36.95 & $+0.12$ & 7.47 & 4856 \\
\midrule
No row-centering & 37.03 & $+0.04$ & 7.47 & 4856 \\
No symmetrization & 37.06 & $+0.01$ & 7.47 & 4856 \\
\bottomrule
\end{tabular}\par
\endgroup
\end{table}

\noindent\textbf{Spatial and Temporal Refinement.} Compared with the backbone without MC-IRB, spatial-only and temporal-only refinement improve PSNR by $0.16$ and $0.23$~dB; combining both yields a $0.39$~dB gain (\Cref{tab:ablation}). The full model uses $8.9\%$ more parameters and $7.2\%$ more FLOPs than the backbone.

\noindent\textbf{Interactions, Gates, and Capacity.} MC-IRB improves PSNR over fixed, conditioned, and field-only gates by $0.29$, $0.20$, and $0.23$~dB (\Cref{tab:ablation_irb}). Setting $\mathbf{J}=0$ retains the nonlinear field-to-gate mapping and reduces PSNR by $0.21$~dB; replacing $\mathbf{J}$ with $|\mathbf{J}|$ reduces it by $0.14$~dB while preserving entrywise coupling magnitudes. These comparisons support adaptive signed interactions. The ten-block backbone remains $0.20$~dB lower despite slightly higher parameter and FLOP counts. At the released full-training checkpoint, permuting the temporal couplings or replacing them with $|\mathbf{J}|$ lowers mid-scale color PSNR by $1.04$ and $0.80$~dB (Appendix~\ref{app:ablation-diagnostics}).

\noindent\textbf{Iteration Count.} Three mean-field updates exceed one by $0.09$~dB and match five at this training budget, supporting the selected $K_{\mathrm{mf}}=3$ (\Cref{tab:ablation_irb}).

\noindent\textbf{Sensing Context.} Removing pairwise or unary conditioning reduces PSNR by $0.17$ and $0.18$~dB; removing the mask or initial-estimate penalty gives $0.11$ and $0.12$~dB reductions. These comparisons support sensing context in both pairwise compatibilities and unary spin states.

\noindent\textbf{Centering and Symmetry.} Removing row-centering or symmetrization changes PSNR by only $0.04$ or $0.01$~dB, respectively. Their principal rationale remains structural: centering expresses affinity relative to a valid row's mean, while symmetrization defines reciprocal couplings for an undirected Ising energy.

\noindent\textbf{Training Budget.} Gains over the backbone without MC-IRB persist at $40$/$80$/$200$ epochs ($0.45$/$0.39$/$0.37$~dB) and are $0.52$/$0.41$/$0.39$~dB on the three benchmarks under the full training setting of \Cref{tab:gray,tab:color,tab:large} (Appendix~\ref{app:ablation-budget}).

\noindent\textbf{Explicit Mask Conditioning.} Removing mask channels reduces PSNR by $0.13$--$0.19$~dB on the training mask and $0.30$--$0.38$~dB on the unseen binary mask (\Cref{tab:ablation_mask}). The larger transfer gaps support explicit conditioning beyond the mask-dependent initialization and final correction retained by all controls.

\begin{table}[!htbp]
\centering
\begingroup
\setlength{\abovecaptionskip}{3pt}
\setlength{\belowcaptionskip}{3pt}
\caption{Mask-conditioning and operator comparisons, PSNR in dB. Rows give
the training configuration and columns the test mask. All channel-removal
variants retain the correct masks in initialization and final data
consistency. Appendix~\ref{app:ablation-masks} describes the operators.}
\label{tab:ablation_mask}
\fontsize{8.5}{9.5}\selectfont
\renewcommand{\arraystretch}{0.96}
\setlength{\tabcolsep}{4pt}
\begin{tabular}{@{}lrrr@{}}
\toprule
Configuration & \makecell[r]{Benchmark\\binary} & \makecell[r]{Unseen\\binary} & Gaussian \\
\midrule
Full model & 37.07 & 37.07 & 18.08 \\
$\mathbf{\Phi},\mathbf{\Phi}_s$ removed from $\mathbf{C}$ & 36.94 & 36.77 & -- \\
Also removed from the stem input & 36.88 & 36.69 & -- \\
Training with random binary masks & 37.10 & -- & -- \\
Training with per-sample Gaussian masks & 27.12 & 27.13 & 35.96 \\
\bottomrule
\end{tabular}\par
\endgroup
\end{table}

The full model keeps $37.07$~dB on the unseen binary mask, and each checkpoint performs best on the operator family it was trained with: $37.07$~dB for binary training on binary masks and $35.96$~dB for Gaussian training on Gaussian masks. Training with random binary masks improves PSNR by $0.03$~dB to $37.10$~dB.

\noindent\textbf{Repeatability.} Across five training seeds, the full model obtains $37.07\pm0.10$~dB (mean $\pm$ sample SD). Paired gains over field-only gating, $\mathbf{J}=0$, and $|\mathbf{J}|$ are $0.26$, $0.24$, and $0.19$~dB, with pointwise $95\%$ intervals above zero. Appendix~\ref{app:ablation-confirm} details the paired protocol and interval assumptions.

\noindent\textbf{Learned Temporal Interactions.} At the fixed checkpoint, removing temporal couplings reduces PSNR by $0.101$~dB, while replacing them with their entrywise absolute values or a validity-preserving permutation reduces PSNR by $0.795$ and $1.036$~dB, respectively (\Cref{tab:app-diagnostics}). The permutation preserves coupling symmetry and spectrum. These results support the functional contribution of coupling signs and token correspondence to temporal refinement.

\begin{table}[!t]
\centering
\vspace{-1em}
\caption{Reconstruction quality and computational cost on the \textbf{small-scale gray} benchmark at compression rate $8$. Latency and peak memory use FP32 eager inference for all methods. The best and second-best PSNR/SSIM are bold and underlined, respectively.}
\label{tab:gray}
\begingroup
\scriptsize
\setlength{\tabcolsep}{1pt}
\renewcommand{\arraystretch}{1.05}
\resizebox{\textwidth}{!}{%
\begin{tabular}{@{}lrrrr*{7}{rr}@{}}
\toprule
Method & Params & FLOPs & Latency & Memory & \multicolumn{2}{c}{Aerial} & \multicolumn{2}{c}{Crash} & \multicolumn{2}{c}{Drop} & \multicolumn{2}{c}{Kobe} & \multicolumn{2}{c}{Runner} & \multicolumn{2}{c}{Traffic} & \multicolumn{2}{c}{Average} \\
 & (M) & (G) & (ms) & (MiB) & PSNR & SSIM & PSNR & SSIM & PSNR & SSIM & PSNR & SSIM & PSNR & SSIM & PSNR & SSIM & PSNR & SSIM \\
\midrule
PnP-FastDVDNet & 2.47 & 29605 & 1265 & 618 & 27.98 & 0.897 & 27.32 & 0.925 & 41.82 & \underline{0.989} & 32.73 & 0.947 & 36.29 & 0.962 & 27.95 & 0.932 & 32.35 & 0.942 \\
EfficientSCI-B & 8.82 & 1424 & 49 & 1355 & 31.50 & 0.953 & 31.13 & 0.971 & 45.18 & \textbf{0.995} & 35.76 & 0.974 & 43.05 & \underline{0.988} & 32.30 & 0.968 & 36.48 & 0.975 \\
STFormer-B & 19.49 & 3060 & 82 & 1745 & 31.56 & 0.953 & 31.06 & 0.970 & 45.08 & \textbf{0.995} & 35.53 & 0.973 & 42.64 & \underline{0.988} & 32.15 & 0.967 & 36.34 & 0.974 \\
HiSViT$_9$ & 8.99 & 1536 & 176 & 4840 & 31.67 & 0.957 & 31.62 & \underline{0.976} & 45.55 & \textbf{0.995} & \underline{36.24} & \underline{0.976} & 43.84 & \textbf{0.991} & 33.06 & \underline{0.973} & 37.00 & \underline{0.978} \\
DU-9stg & 6.67 & 1690 & 437 & 2647 & \underline{32.27} & \underline{0.961} & \underline{32.47} & \textbf{0.978} & \textbf{45.99} & \textbf{0.995} & \textbf{36.51} & \textbf{0.978} & \textbf{44.75} & \textbf{0.991} & \underline{33.45} & \textbf{0.974} & \textbf{37.57} & \textbf{0.980} \\
\rowcolor{green!5}
\textbf{SpinSCI-B} & \textbf{7.47} & \textbf{1214} & \textbf{131} & \textbf{1850} & \textbf{33.48} & \textbf{0.967} & \textbf{32.83} & \textbf{0.978} & \underline{45.67} & \textbf{0.995} & 35.71 & 0.974 & \underline{43.87} & \textbf{0.991} & \textbf{33.50} & \underline{0.973} & \underline{37.51} & \textbf{0.980} \\
\bottomrule
\end{tabular}%
}
\endgroup
\par\vspace{2pt}
\begingroup
\normalsize
    \centering
    \includegraphics[width=\columnwidth]{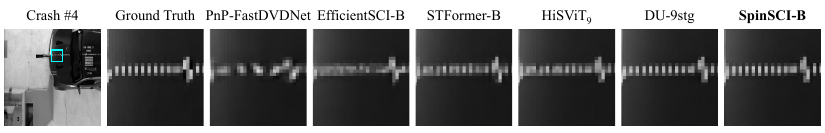}
    \vspace{-2.5em}
    \captionof{figure}{
    Qualitative comparison on frame 4 of the Crash sequence from the small-scale gray benchmark. The first panel shows the full ground-truth frame with the selected region marked in cyan. Subsequent panels show enlarged crops using nearest-neighbor interpolation.
    }
    \label{fig:gray}
\endgroup
\end{table}

\begin{table}[!t]
\centering
\vspace{-1em}
\caption{Reconstruction quality and computational cost on the \textbf{mid-scale color} benchmark at compression rate $8$. Latency and peak memory use FP32 eager inference for all methods. The best and second-best PSNR/SSIM are bold and underlined, respectively.}
\label{tab:color}
\scriptsize
\setlength{\tabcolsep}{1pt}
\renewcommand{\arraystretch}{1.05}
\resizebox{\textwidth}{!}{%
\begin{tabular}{@{}lrrrr*{7}{rr}@{}}
\toprule
Method & Params & FLOPs & Latency & Memory & \multicolumn{2}{c}{Beauty} & \multicolumn{2}{c}{Bosphorus} & \multicolumn{2}{c}{Jockey} & \multicolumn{2}{c}{Runner} & \multicolumn{2}{c}{ShakeNDry} & \multicolumn{2}{c}{Traffic} & \multicolumn{2}{c}{Average} \\
 & (M) & (G) & (ms) & (MiB) & PSNR & SSIM & PSNR & SSIM & PSNR & SSIM & PSNR & SSIM & PSNR & SSIM & PSNR & SSIM & PSNR & SSIM \\
\midrule
PnP-FastDVDNet & 2.48 & 120077 & 40317 & 2648 & 35.27 & 0.972 & 37.24 & 0.978 & 35.63 & 0.950 & 38.22 & 0.965 & 33.71 & \textbf{0.969} & 27.49 & 0.915 & 34.60 & 0.955 \\
EfficientSCI-B & 8.83 & 5702 & 211 & 5207 & 37.51 & 0.979 & 40.89 & 0.988 & 38.49 & 0.969 & 42.73 & 0.985 & 35.19 & 0.953 & 30.13 & 0.943 & 37.49 & 0.970 \\
MambaSCI-B & 8.59 & 557 & 974 & 3809 & 36.95 & 0.979 & 38.62 & 0.982 & 36.54 & 0.960 & 40.02 & 0.977 & 34.55 & 0.950 & 27.52 & 0.904 & 35.70 & 0.959 \\
STFormer-B & 19.49 & 12155 & 369 & 6368 & 37.37 & \underline{0.981} & 40.39 & 0.988 & 38.32 & 0.968 & 42.45 & 0.985 & 35.15 & 0.956 & 30.24 & 0.939 & 37.32 & 0.970 \\
HiSViT$_9$ & 8.99 & 6146 & 757 & 19156 & 37.75 & \underline{0.981} & 41.50 & \underline{0.990} & 39.29 & \underline{0.972} & 43.27 & \underline{0.986} & 35.49 & 0.958 & \textbf{30.76} & \textbf{0.946} & 38.01 & \underline{0.972} \\
DU-9stg & 6.67 & 6763 & 1781 & 10517 & \underline{37.90} & \underline{0.981} & \underline{41.65} & \underline{0.990} & \textbf{39.31} & \textbf{0.973} & \textbf{43.55} & \textbf{0.987} & \underline{35.72} & 0.960 & 30.56 & \underline{0.945} & \underline{38.12} & \textbf{0.973} \\
\rowcolor{green!5}
\textbf{SpinSCI-B} & \textbf{7.47} & \textbf{4856} & \textbf{635} & \textbf{7151} & \textbf{38.04} & \textbf{0.982} & \textbf{42.05} & \textbf{0.991} & \underline{39.30} & \textbf{0.973} & \underline{43.51} & \textbf{0.987} & \textbf{35.83} & \underline{0.961} & \underline{30.72} & \underline{0.945} & \textbf{38.24} & \textbf{0.973} \\
\bottomrule
\end{tabular}%
}
\end{table}

\begin{figure}[!t]
    \centering
    \vspace{-1em}
    \includegraphics[width=\columnwidth]{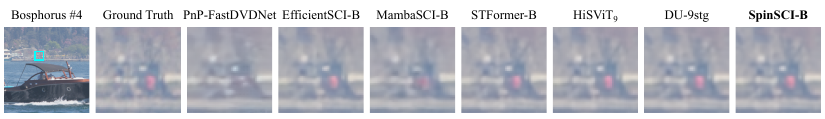}
    \vspace{-2.5em}
    \caption{
    Qualitative comparison on frame 4 of the Bosphorus sequence from the mid-scale color benchmark. The first panel shows the full ground-truth frame with the selected region marked in cyan. Subsequent panels show enlarged crops using nearest-neighbor interpolation.
    }
    \label{fig:color}
\end{figure}

\begin{table}[!t]
\centering
\vspace{-1em}
\caption{\textbf{Large-scale color} at compression rate $8$. All methods use FP32 eager inference. Latency and memory cover the complete $24$-tile reconstruction of one snapshot. Best and second-best PSNR/SSIM are bold and underlined.}
\label{tab:large}
\begingroup
\scriptsize
\setlength{\tabcolsep}{1pt}
\renewcommand{\arraystretch}{1.05}
\resizebox{\textwidth}{!}{%
\begin{tabular}{@{}lrrrr*{5}{rr}@{}}
\toprule
Method & Params & FLOPs & Latency & Memory & \multicolumn{2}{c}{Football} & \multicolumn{2}{c}{Hummingbird} & \multicolumn{2}{c}{Messi} & \multicolumn{2}{c}{Swinger} & \multicolumn{2}{c}{Average} \\
 & (M) & (G) & (ms) & (MiB) & PSNR & SSIM & PSNR & SSIM & PSNR & SSIM & PSNR & SSIM & PSNR & SSIM \\
\midrule
PnP-FastDVDNet & 2.48 & 956865 & 297970 & 1029 & 37.42 & 0.982 & 38.51 & 0.979 & 33.54 & 0.979 & 28.26 & 0.930 & 34.43 & 0.967 \\
EfficientSCI-B & 8.83 & 45100 & 1597 & 2046 & 40.33 & 0.989 & 40.41 & 0.974 & 34.85 & 0.974 & 29.45 & 0.940 & 36.26 & 0.969 \\
MambaSCI-B & 8.59 & 4405 & 7645 & 1587 & 35.91 & 0.974 & 31.67 & 0.641 & 29.09 & 0.923 & 27.01 & 0.900 & 30.92 & 0.859 \\
STFormer-B & 19.49 & 96071 & 2578 & 2478 & 40.64 & 0.990 & 40.43 & 0.974 & 34.87 & 0.976 & 29.59 & 0.943 & 36.38 & 0.971 \\
HiSViT$_9$ & 8.99 & 48977 & 8173 & 6715 & \underline{42.25} & \textbf{0.993} & \underline{41.48} & 0.989 & 36.39 & \underline{0.986} & \textbf{30.43} & \textbf{0.954} & 37.64 & \textbf{0.980} \\
DU-9stg & 6.67 & 53893 & 14832 & 4003 & \textbf{42.34} & \textbf{0.993} & \textbf{41.99} & \underline{0.991} & \underline{36.71} & \underline{0.986} & \underline{30.16} & \underline{0.950} & \textbf{37.80} & \textbf{0.980} \\
\rowcolor{green!5}
\textbf{SpinSCI-B} & \textbf{7.47} & \textbf{38402} & \textbf{4169} & \textbf{2695} & 41.68 & \underline{0.992} & \textbf{41.99} & \textbf{0.992} & \textbf{37.07} & \textbf{0.987} & 29.98 & 0.947 & \underline{37.68} & \underline{0.979} \\
\bottomrule
\end{tabular}%
}
\endgroup
\par\vspace{2pt}
\begingroup
\normalsize
    \centering
    \includegraphics[width=\columnwidth]{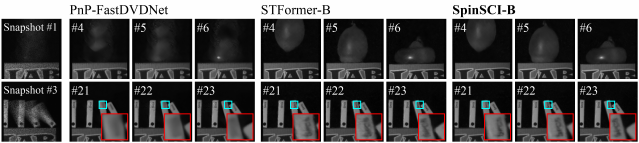}
    \vspace{-2em}
    \captionof{figure}{Qualitative comparison on Water Balloon~\cite{e2ecnn} ($512\times512\times10$) and Domino~\cite{e2ecnn} ($512\times512\times60$) in the \textbf{real benchmark} at compression rate $10$. $\#$ denotes the frame number. Only baselines with officially released real-data checkpoints are shown.}
    \label{fig:real}
\endgroup
\end{table}

\subsection{Comparison with State-of-the-Art Methods}\label{sec:comparison}
\textbf{Baselines.} Comparisons include PnP-FastDVDNet~\cite{pnp}, EfficientSCI-B~\cite{efficient}, MambaSCI-B~\cite{mamba}, STFormer-B~\cite{stformer}, HiSViT$_9$~\cite{hisvit}, and DU-9stg~\cite{du}. MambaSCI is designed for quad-Bayer measurements and has no reported grayscale scores.

\noindent\textbf{Quantitative and Qualitative Results.} \Cref{tab:gray,tab:color,tab:large} report PSNR/SSIM and computational cost. Small-scale gray and mid-scale color baseline scores come from the respective papers; evaluated on our masks, the official HiSViT$_9$ checkpoints reproduce their reported PSNR within $0.005$~dB. For DU-9stg on mid-scale color, the measured PSNR/SSIM are used because the measured mean PSNR is $0.18$~dB below the reported value. Large-scale color scores use the released checkpoints. Appendix~\ref{app:implementation-details} details the quality and profiling protocols.

Compared with HiSViT$_9$, SpinSCI-B achieves average PSNR gains of $0.51$ and $0.23$~dB on small-scale gray and mid-scale color, and comparable PSNR on large-scale color ($+0.04$~dB), with $16.9\%$ fewer parameters, $21.0$--$21.6\%$ fewer FLOPs, $16.1$--$49.0\%$ lower FP32 eager latency, and $59.9$--$62.7\%$ lower peak memory. Relative to DU-9stg, SpinSCI-B's average PSNR differs by at most $0.13$~dB across these benchmarks, with more parameters ($7.47$M versus $6.67$M), $28.2$--$28.7\%$ fewer FLOPs, $64.3$--$71.9\%$ lower inference latency, and $30.1$--$32.7\%$ lower peak memory.

\Cref{fig:gray,fig:color,fig:real} show representative reconstructions. Figures~\ref{fig:supp-gray}--\ref{fig:supp-large} in Appendix~\ref{app:qualitative} cover all gray/color sequences. The displayed DU-9stg reconstructions use the shared-mask evaluation described in Appendix~\ref{app:implementation-details}.

Full-frame reconstructed videos for datasets in the \textbf{real benchmark} at compression rates of $10$ and $50$ are shown as frame sequences in Appendices~\ref{app:real-cr10}--\ref{app:real-cr50}.

\noindent\textbf{Additional Analyses.} The appendices add magnetization and gate maps (Appendix~\ref{app:ablation-diagnostics}), mask calibration errors (Appendix~\ref{app:ablation-calibration}), baseline re-runs on official and benchmark masks (\Cref{tab:hisvit-measured}), BF16 compiled inference (Appendix~\ref{app:compiled}), other compression rates (Appendix~\ref{app:other-cr}), and first-order hardware energy estimates (Appendix~\ref{app:hw-ppa}).

\noindent\textbf{Limitations.} The current MC-IRB implementation materializes pre-softmax attention logits to construct spin couplings, limiting direct compatibility with FlashAttention-style kernels~\cite{flashattention}. However, SpinSCI-B achieves fewer FLOPs, lower inference latency, and lower peak memory than HiSViT$_9$ and DU-9stg with comparable PSNR.

\section{Conclusion}
\textit{SpinSCI} combines visual affinity and sensing information through signed spin interactions. Finite mean-field inference coordinates residual refinement within the spatial and temporal branches, while physics-guided initialization and final data consistency incorporate the sensing model. Formal analysis establishes coupling symmetry, bounded magnetizations and gates, and binary-mask measurement-residual reduction. Benchmark comparisons show higher PSNR than HiSViT$_9$ on small-scale gray and mid-scale color (comparable on large-scale color) with fewer parameters, comparable PSNR to DU-9stg, and fewer FLOPs, lower inference latency, and lower peak memory than both methods.

\subsection*{AI Use Statement}
Generative AI tools were used to assist with writing and checking formatting
compliance. The authors reviewed all AI-assisted changes and take responsibility
for the final content of this work, including text, claims, and artifacts
produced with the aid of generative AI.

\subsection*{Ethics Statement}
This work uses public video snapshot compressive imaging datasets and standard benchmark
protocols and does not involve human subjects or private data. The authors have
read and commit to the ICLR Code of Ethics.

\subsection*{Reproducibility Statement}
The model architecture, mask-conditioned refinement computations, and
data-consistency operations are specified in \Cref{sec:method}.
\Cref{sec:exp_setup} describes the datasets, training settings, loss,
evaluation metrics, and efficiency protocol; \Cref{sec:ablation,sec:comparison}
describe the ablation and comparison settings. Appendices~\ref{app:formal}--\ref{app:qualitative}
provide proofs and verification coverage, additional ablations and statistical
reporting, implementation details, and qualitative comparisons, respectively.
Appendix~\ref{app:related} extends the related-work discussion.
Code and reference checkpoints are available at
\href{https://github.com/skethz/SpinSCI}{\textcolor{codeblue}{github.com/skethz/SpinSCI}}.

\bibliography{ref,formal_refs,hw_ppa}
\bibliographystyle{iclr2027_conference}
\clearpage
\appendix
\setcounter{table}{0}
\setcounter{figure}{0}
\setcounter{equation}{0}
\input{sections/formal_proofs}
\clearpage
\input{sections/ablation_details}
\clearpage
\input{sections/implementation_details}
\clearpage
\input{sections/additional_results}
\clearpage
\input{sections/related_work_extended}
\clearpage
\input{sections/hardware_cost}
\end{document}

%% file: sections/formal_proofs.tex
\numberwithin{equation}{section}
\section{Mathematical Properties of SpinSCI}\label{app:formal}

This appendix analyzes the sensing operator, data-consistency correction, and
Mask-Conditioned Ising Refinement Block (MC-IRB). The analysis follows SpinSCI
with nine transformer blocks, conditioning on the back-projected initialization,
three mean-field updates per refiner, and a final data-consistency correction
with unit step size. All theorems concern exact real arithmetic.
Reconstruction quality remains an empirical property of the learned prior.

The companion Lean development~\cite{lean4,mathlib} checks the identities and
conditional results identified in \Cref{app:coverage}. Mathematical assumptions
are stated separately from implementation observations.

\paragraph{Notation.}
The derivations omit boldface from vector and tensor symbols used in the main
paper. The operators $\mathcal A_\Phi x$ and $\mathcal A_\Phi^\top y$ denote
$\mathcal A(\mathbf x;\mathbf\Phi)$ and
$\mathcal A^\top(\mathbf y;\mathbf\Phi)$, respectively.
The initial estimate is written as $x^{(0)}=\widehat{\mathbf x}^{(0)}$, and
its normalized version as $u=\overline{\mathbf x}^{(0)}$.
For the current binary masks, the supplied normalizer $d$ corresponds to
$\mathbf\Phi_s$; $D$ denotes the mask energy before the zero-exposure safeguard.
The generic correction $z\mapsto z^+$ specializes to the main paper's final
grayscale update with $z=\widetilde{\mathbf x}$,
$z^+=\widehat{\mathbf x}$, and $\alpha=1$.
For color reconstruction, $X$ and $X^+$ correspond to
$\widetilde{\mathbf x}_{\mathrm{rgb}}$ and
$\widehat{\mathbf x}_{\mathrm{rgb}}$.

\subsection{Sensing, Adjoint, and Mask Normalization}\label{app:sensing}

Let $\Omega$ be the spatial pixel set and $t\in\{1,\ldots,T\}$ index frames.
A grayscale or mosaic video is $x\in\mathbb R^{T\times|\Omega|}$ and its
measurement is $y\in\mathbb R^{|\Omega|}$. For fixed masks $\Phi$,
\begin{equation}
(\mathcal A_\Phi x)_p=\sum_t\Phi_{t,p}x_{t,p},\qquad
(\mathcal A_\Phi^\top y)_{t,p}=\Phi_{t,p}y_p,\qquad
D_p=\sum_t\Phi_{t,p}^{2}.
\end{equation}

\begin{proposition}[Adjoint and diagonal measurement operator]
Under the standard Euclidean inner products,
\begin{equation}
\langle\mathcal A_\Phi x,y\rangle
 =\langle x,\mathcal A_\Phi^\top y\rangle,\qquad
(\mathcal A_\Phi\mathcal A_\Phi^\top y)_p=D_p y_p.
\end{equation}
Furthermore, $D_p\ge0$, and for binary masks $D_p=\sum_t\Phi_{t,p}$.
\end{proposition}
\begin{proof}
The first identity follows by interchanging the finite sums:
$\sum_p(\sum_t\Phi_{t,p}x_{t,p})y_p
=\sum_{p,t}x_{t,p}(\Phi_{t,p}y_p)$.
Substituting the adjoint into the forward operator gives
$\sum_t\Phi_{t,p}(\Phi_{t,p}y_p)=D_p y_p$.
The remaining statements follow from nonnegativity of squares and
$b^2=b$ for $b\in\{0,1\}$.
\end{proof}

The implementation computes the sum of mask values and replaces a zero sum
by one. For binary masks this supplied normalizer is
\begin{equation}
 d_p=\begin{cases}D_p,&D_p>0,\\1,&D_p=0.\end{cases}
\end{equation}
Thus the code's normalizer is distinguished from the true energy $D$ at
unexposed pixels. For nonbinary masks, using the sum of squared mask values
with the same zero-exposure safeguard preserves the unit-step residual
factor derived below. The results first allow a supplied $d$ with positive
denominators and then specialize to binary masks.

\subsection{Data Consistency and Initialization}\label{app:dc}

For $\epsilon>0$ and a supplied normalizer satisfying $d_p+\epsilon>0$
at every pixel, define a correction of an arbitrary estimate $z$ by
\begin{equation}\label{eq:app-dc}
 z^+=z+\alpha\mathcal A_\Phi^\top
       \left(\frac{y-\mathcal A_\Phi z}{d+\epsilon}\right).
\end{equation}
Division is pixelwise. This is a measurement-domain preconditioned correction;
it is also a gradient step for the weighted quadratic with weight
$\operatorname{diag}((d+\epsilon)^{-1})$.

\begin{proposition}[Exact residual update]\label[proposition]{prop:dc}
For $r=y-\mathcal A_\Phi z$,
\begin{equation}\label{eq:app-residual}
 r_p^+=\left(1-\frac{\alpha D_p}{d_p+\epsilon}\right)r_p.
\end{equation}
If $0\le\alpha D_p/(d_p+\epsilon)\le2$ at every pixel, then
$|r_p^+|\le|r_p|$, and hence $\|r^+\|_2\le\|r\|_2$.
With binary masks and the current $\alpha=1$, exposed pixels satisfy
\begin{equation}
 r_p^+=\frac{\epsilon}{D_p+\epsilon}r_p,
 \qquad 0<\frac{\epsilon}{D_p+\epsilon}<1.
\end{equation}
At $D_p=0$, the measurement residual is unchanged.
\end{proposition}
\begin{proof}
Applying $\mathcal A_\Phi$ to \Cref{eq:app-dc} and using
$\mathcal A_\Phi\mathcal A_\Phi^\top=\operatorname{diag}(D)$ yields
\Cref{eq:app-residual}. For $q\in[0,2]$, $|1-q|\le1$; multiplication
by $|r_p|$ gives the pointwise inequality. Squaring and summing gives the
Euclidean inequality. If $D_p>0$, $d_p=D_p$ and
$1-D_p/(D_p+\epsilon)=\epsilon/(D_p+\epsilon)\in(0,1)$.
When $D_p=0$, the multiplier in \Cref{eq:app-residual} equals one.
\end{proof}

\begin{remark}[Scope and bounded nonbinary masks]\label[remark]{rem:bounded-masks}
For $0\le\Phi_{t,p}\le1$, $\Phi_{t,p}^2\le\Phi_{t,p}$ gives
$0\le D_p\le d_p$ under mask-sum normalization with zero sums replaced by one.
Thus $\alpha=1$ yields $0\le D_p/(d_p+\epsilon)<1$, and
\Cref{prop:dc} ensures residual nonincrease. The guarantee concerns the
explicit correction; $\epsilon>0$ need not give exact measurement
consistency, and residual reduction does not certify clean-video recovery,
particularly under measurement noise. Positive learned step sizes alone
do not ensure the proposition's upper bound; SpinSCI uses fixed $\alpha=1$.
\end{remark}

The initializer uses $z=\mathcal A_\Phi^\top y$ followed by a unit correction.
Its residual is therefore
\begin{equation}\label{eq:app-init}
 y_p-(\mathcal A_\Phi x^{(0)})_p
 =\left(1-\frac{D_p}{d_p+\epsilon}\right)(1-D_p)y_p.
\end{equation}
This follows by inserting $\mathcal A_\Phi z=D y$ into
\Cref{eq:app-residual}. In the idealized case $\epsilon=0$, with
$d_p=D_p>0$, the initializer reduces to
$x^{(0)}_{t,p}=\Phi_{t,p}y_p/D_p$. Neither form can
identify the temporal decomposition uniquely without additional prior information.

\subsection{Bayer Sampling and RGB Correction}\label{app:bayer}

Let $c(p)\in\{R,G,B\}$ specify the RGGB color sampled at pixel $p$.
Define the color-sampling operator $\mathcal B$ and its adjoint by
\begin{equation}
 (\mathcal B X)_{t,p}=X_{c(p),t,p},\qquad
 (\mathcal B^\top\delta)_{c,t,p}
   =\mathbf1[c=c(p)]\delta_{t,p}.
\end{equation}
Consequently $\mathcal B\mathcal B^\top=I$ on mosaic videos. For a prior RGB
output $X$ with mosaic $z=\mathcal B X$, let $z^+$ be its corrected mosaic and
set $X^+=X+\mathcal B^\top(z^+-z)$.

\begin{proposition}[Observed-channel correction]\label[proposition]{prop:bayer}
The corrected RGB estimate satisfies $\mathcal B X^+=z^+$, and
$X^+_{c,t,p}=X_{c,t,p}$ whenever $c\ne c(p)$. Its measurement residual under
$\mathcal A_\Phi\mathcal B$ obeys \Cref{eq:app-residual}.
\end{proposition}
\begin{proof}
Sampling the scattered correction returns $z^+-z$ because exactly one color
is selected at each pixel. Thus $\mathcal B X^+=z+(z^+-z)=z^+$.
The scatter is zero in all other channels. Applying the forward operator to
this equality gives the residual claim.
\end{proof}
The four-subgrid initializer is the same pixelwise initialization after a
permutation of pixel indices; it does not mix the RGGB subgrids. The current
implementation requires even spatial dimensions for this rearrangement.

\subsection{Conditioning and Coupling Construction}\label{app:couplings}

For the active back-projection conditioning mode, define the per-sample scale
and conditioning signal before spatial interpolation as
\begin{equation}
 a=\max\left(\frac{1}{T|\Omega|}\sum_{t,p}|x^{(0)}_{t,p}|,\epsilon\right),
 \qquad u_{t,p}=x^{(0)}_{t,p}/a.
\end{equation}
Token features have width $d_{\mathrm{feat}}=256$, and $\mathbf C$ denotes
the main paper's conditioning tensor.
After nearest-neighbor resizing of $\Phi$ and bilinear resizing of $u,d$, a
token receives $c_i=[\phi_i,u_i,d_i]\in\mathbb R^3$.
The order is relevant to the learned projections. The variable called
\texttt{lambda\_res} in the code weights differences of $u$ in this mode.

Let $n$ denote the number of token positions in one refinement group.
The active configuration uses $n=49$ in each spatial window (including
padding where needed) and $n=T$ in each temporal sequence.
For one such group, let $v_{ij}\in\{0,1\}$ be the valid-pair indicator, with
$v_{ij}=v_{ji}$ and $v_{ii}=1$. The active configuration has signed coupling,
symmetrization enabled, and top-$k$ sparsification disabled. Spatial logits
are averaged over the two native-resolution heads; pooled-head logits are
not used to construct the spatial Ising matrix. Temporal logits are averaged
over all four temporal heads. Writing this head average as $\overline L$,
\begin{align}
 \Pi_{ij}&=\lambda_\phi|\phi_i-\phi_j|
       +\lambda_u|u_i-u_j|+\lambda_d|d_i-d_j|,\\
 \widetilde J_{ij}&=v_{ij}(\overline L_{ij}-\Pi_{ij}),\qquad
 n_i=\sum_j v_{ij},\qquad \mu_i=\frac{1}{n_i}\sum_j\widetilde J_{ij},\\
 J_{ij}^{\mathrm{ctr}}&=v_{ij}(\widetilde J_{ij}-\mu_i),\qquad
 J_{ij}=\gamma v_{ij}\mathbf{1}[i\ne j]\frac{J_{ij}^{\mathrm{ctr}}+J_{ji}^{\mathrm{ctr}}}{2}.
 \label{eq:app-coupling}
\end{align}
The parameters $\lambda_\phi,\lambda_u,\lambda_d,\gamma$ denote effective
nonnegative values obtained by clamping learned scalars. The diagonal is
included in centering and removed subsequently. A finite negative attention
mask sentinel is replaced by zero on invalid pairs before centering.

\begin{proposition}[Coupling invariants]\label[proposition]{prop:couplings}
Each centered intermediate row satisfies $\sum_j J_{ij}^{\mathrm{ctr}}=0$.
The final matrix has $J_{ii}=0$, $J_{ij}=0$ if $v_{ij}=0$, and $J=J^\top$.
Its final row sums need not vanish.
\end{proposition}
\begin{proof}
Since $n_i\ge1$, summation over valid pairs subtracts exactly $n_i\mu_i$,
which equals the row sum of $\widetilde J$. The two final indicators set the diagonal
and invalid pairs to zero. Symmetry follows from $v_{ij}=v_{ji}$ and
$J_{ij}^{\mathrm{ctr}}+J_{ji}^{\mathrm{ctr}}=J_{ji}^{\mathrm{ctr}}+J_{ij}^{\mathrm{ctr}}$.
For a counterexample to zero final row sums,
consider three mutually valid tokens, $\overline L=0$, $\gamma=\lambda_u=1$, other
penalty weights zero, and $u=(0,0,1)$. Direct evaluation gives
\begin{equation}
 J=\begin{pmatrix}0&1/3&-1/2\\1/3&0&-1/2\\-1/2&-1/2&0\end{pmatrix},
 \qquad\sum_j J_{1j}=-1/6.
\end{equation}
This matrix and the nonzero row sum are also checked in Lean.
\end{proof}

\begin{remark}[Interpretation of the pairwise penalty]
Before symmetrization, the change due to a penalty $\Pi_{ij}$ is relative to
its valid-row average $\mu_i^\Pi$. Let $\mu_i^L$ denote the corresponding
average of $\overline L_{ij}$. Then
$(\overline L_{ij}-\Pi_{ij})-(\mu_i^L-\mu_i^\Pi)
=(\overline L_{ij}-\mu_i^L)-(\Pi_{ij}-\mu_i^\Pi)$.
Thus a larger raw disagreement term does not imply that every final signed
coupling becomes smaller in magnitude. The example above even creates a
positive coupling from zero logits. The construction supports attractive
and repulsive latent-spin interactions.
\end{remark}

\subsection{Ising Interpretation and Mean-Field Stationarity}\label{app:ising}

For fixed $h,J$ during one refiner, define spins $s_i\in\{-1,+1\}$,
energy and Gibbs law
\begin{equation}
 E(s)=-\sum_i h_i s_i-\frac12\sum_{i,j}J_{ij}s_i s_j,
 \qquad p(s)=\mathcal Z^{-1}e^{-\beta E(s)},\quad\beta\ge0.
\end{equation}
Since $J$ is symmetric with zero diagonal,
$\tfrac12\sum_{i,j}J_{ij}s_is_j=\sum_{i<j}J_{ij}s_is_j$,
so this energy equals the main paper's Hamiltonian.
The partition function $\mathcal Z=\sum_{s\in\{-1,+1\}^n}e^{-\beta E(s)}$
is positive since it is a finite sum of positive terms.
For a factorized distribution $q(s)=\prod_iq_i(s_i)$ with interior
means $m_i\in(-1,1)$, let
\begin{equation}
 q_i(\pm1)=\frac{1\pm m_i}{2},\qquad
 \psi(t)=\frac{1+t}{2}\log\frac{1+t}{2}
        +\frac{1-t}{2}\log\frac{1-t}{2}.
\end{equation}
Expanding $\operatorname{KL}(q\Vert p)$ yields, up to $\log\mathcal Z$,
\begin{equation}\label{eq:app-free-energy}
 \mathcal F(m)=-\beta\sum_i h_i m_i
    -\frac{\beta}{2}\sum_{i,j}J_{ij}m_i m_j+\sum_i\psi(m_i).
\end{equation}
Indeed, factorization gives
$\mathbb E_q[\log q]=\sum_i\psi(m_i)$ and
$\mathbb E_q[s_i s_j]=m_i m_j$ for $i\ne j$; the diagonal terms are absent.
This finite Gibbs-to-variational expansion is a written derivation. The
Lean coverage concerns the scalar stationarity result below and the
symmetry identity used in its interaction field.

\begin{proposition}[Mean-field self-consistency]\label[proposition]{prop:ising}
Every interior stationary point of \Cref{eq:app-free-energy} satisfies
\begin{equation}
 m_i=\tanh\!\left(\beta\left(h_i+\sum_jJ_{ij}m_j\right)\right).
\end{equation}
For fixed other means, this is the unique stationary value of coordinate $i$.
\end{proposition}
\begin{proof}
Symmetry and a zero diagonal make the coordinate derivative
\begin{equation}
 \frac{\partial\mathcal F}{\partial m_i}
 =-\beta\left(h_i+\sum_jJ_{ij}m_j\right)
     +\frac12\log\frac{1+m_i}{1-m_i}.
\end{equation}
Setting it to zero and solving the log-odds equation gives the displayed
$\tanh$ identity. The second coordinate derivative is
$1/(1-m_i^2)>0$ on $(-1,1)$, so there is only one stationary coordinate
value. Alternatively, the probabilities proportional to $e^x,e^{-x}$
have mean $\tanh x$ and log odds $2x$, where
$x=\beta(h_i+\sum_jJ_{ij}m_j)$.
\end{proof}
A simultaneous update of every coordinate is the synchronous mean-field
iteration implemented by MC-IRB. Coordinate stationarity does not imply
that each synchronous iteration decreases the variational objective.

\subsection{Finite-Step Bounds and Residual Gates}\label{app:bounds}

Given fixed feature and conditioning vectors $[p_i;c_i]$, the refiner uses
\begin{align}
 h_i&=w_h^\top[p_i;c_i]+b_h,
 &m_i^{(0)}&=\tanh(w_m^\top[p_i;c_i]+b_m),\\
 m_i^{(\ell+1)}&=\tanh\!\left(\beta(h_i+\sum_jJ_{ij}m_j^{(\ell)})\right),
 &g_i&=\sigma(\kappa m_i^{(K_{\rm mf})}),
\end{align}
where $\sigma$ denotes the sigmoid function, the effective $\beta,\kappa\ge0$
are clamped learned scalars, and
$K_{\rm mf}=3$ in the current configuration.

\begin{proposition}[Bounds and identity initialization]\label[proposition]{prop:bounds}
Every finite real mean-field iterate lies in $(-1,1)^n$. In particular,
\begin{equation}
 \sigma(-\kappa)\le g_i\le\sigma(\kappa),\qquad
 m_i=0\ \Longrightarrow\ g_i=1/2.
\end{equation}
For the token-wise residual MLP $\mathcal R(p_i)$ and any norm homogeneous over real
scalars,
\begin{equation}
 \|p_i+g_i\mathcal R(p_i)-p_i\|=g_i\|\mathcal R(p_i)\|.
\end{equation}
When the final affine layer of $\mathcal R$ has zero weights and zero bias, the
refinement block is exactly the identity.
\end{proposition}
\begin{proof}
$\tanh$ maps every finite real argument into $(-1,1)$, covering the initializer
and all subsequent iterations by induction. Multiplying $m_i\in[-1,1]$ by
$\kappa\ge0$ gives $-\kappa\le\kappa m_i\le\kappa$; monotonicity of
$\sigma$ gives the gate interval. Its value at zero is $1/2$.
The norm equality follows by homogeneity and positivity of $g_i$.
A zero final affine map gives $\mathcal R(p_i)=0$ regardless of the gate.
\end{proof}
At initialization, $\kappa=1$ gives approximate bounds $0.269$ and $0.731$.
These numerical limits are not fixed throughout training because $\kappa$
is learned. A neutral magnetization does not close the gate, and ordering
gates alone does not order correction magnitudes across different tokens.

\subsection{A Sufficient Condition for Mean-Field Convergence}\label{app:stability}

For the fixed mean-field map
$f(m)=\tanh(\beta(h+Jm))$, with coordinatewise $\tanh$ and the induced
Euclidean operator norm $\|J\|_2$,
\begin{equation}
 \|f(m)-f(\widetilde m)\|_2
 \le\beta\|J\|_2\,\|m-\widetilde m\|_2.
\end{equation}

\begin{theorem}[Conditional convergence]\label{thm:contraction}
If $q=\beta\|J\|_2<1$, then $f$ has a unique fixed point $m^\star$ and
its iterates converge to $m^\star$ from every initialization. Moreover,
\begin{equation}
 \|f^{K}(m^{(0)})-m^\star\|_2
 \le q^{K}\|m^{(0)}-m^\star\|_2.
\end{equation}
\end{theorem}
\begin{proof}
The scalar derivative $\tanh'(a)=1/\cosh^2(a)$ lies in $[0,1]$, so scalar
$\tanh$ is 1-Lipschitz. Squaring and summing the coordinate bounds gives
1-Lipschitz continuity in the Euclidean norm. Composition with the affine
map yields the preceding bound. The contraction mapping theorem on the
complete finite-dimensional Euclidean space gives existence, uniqueness,
and convergence. Iterating the Lipschitz inequality against the fixed
point yields the geometric error bound.
\end{proof}
Appendix~\ref{app:ablation-diagnostics} evaluates this condition on the trained checkpoints.
Regardless of $q$, finite mean-field outputs remain bounded as in
\Cref{prop:bounds}.

\subsection{Formal Verification Coverage}\label{app:coverage}

\begin{center}
\small
\begin{tabular}{p{0.19\linewidth}p{0.72\linewidth}}
\toprule
Lean module & Checked content \\
\midrule
\texttt{Physics} & Finite sensing/adjoint identities; mask energy; exact residual update; pointwise nonincrease; unit-step residual factor; initialization residual; selected-color scatter and sampling.\\
\texttt{Couplings} & Valid-row centering; zero diagonal; invalid-pair exclusion; symmetry; centered-penalty algebra; exact three-token counterexamples.\\
\texttt{Ising} & Binary probabilities and $\tanh$ mean; entropy derivative; scalar stationary-point equivalence; symmetric interaction-field identity.\\
\texttt{Bounds} & Magnetization/gate bounds; neutral gate; gated-residual norm; zero-residual identity.\\
\texttt{Stability} & Scalar/vector Lipschitz bounds; Euclidean matrix action; finite-step error bound; unique fixed point and convergence under $\beta\|J\|_2<1$.\\
\bottomrule
\end{tabular}
\end{center}

The proof project pins Lean 4.33.1 and mathlib 4.33.1. Its verification script
builds every module and audits the transitive axioms of every named theorem,
allowing only the standard foundations \texttt{propext},
\texttt{Classical.choice}, and \texttt{Quot.sound}. The archive contains the
source, dependency manifest, verification script, and audit report.
The full finite Gibbs-to-variational reduction, full-image summation of the
pointwise sensing bounds, and coordinatewise assembly of the Bayer operator
are explained in the written proofs; their primitive identities are checked
as described in the table.

Separate local CPU probes instantiate the copied grayscale and color source,
check operator identities, and confirm masked-coupling invariants. These
probes use random tensors and the Conv3d feed-forward path. They do not
verify trained checkpoints, reconstruction accuracy, CUDA/Triton kernels,
mixed-precision numerical behavior, or the complete training program.

%% file: sections/ablation_details.tex
\section{Ablation Results and Confirmatory Protocol}\label{app:ablation}
\numberwithin{table}{section}
\numberwithin{figure}{section}

This appendix presents additional training-budget and mask comparisons,
control definitions, the repeated-run protocol, and fixed-checkpoint
mechanism diagnostics.

\subsection{Training Budget}\label{app:ablation-budget}

The MC-IRB design comparisons and shared protocol are reported in
\Cref{sec:ablation}. \Cref{tab:app-budget-pilot} compares the full model
and the backbone without MC-IRB at three training budgets of the controlled
protocol and under the full training setting of
\Cref{tab:gray,tab:color,tab:large}.

\begin{table}[H]
\centering
\caption{Full-model and no-MC-IRB PSNR under the controlled protocol and
the full training setting. Controlled rows follow \Cref{sec:ablation}
($128\times128$ crops, mid-scale color), varying only the number of epochs.
Full-setting rows train the backbone without MC-IRB for $200$ epochs with
the configuration of the released SpinSCI-B checkpoints ($256\times256$ gray
and $512\times512$ color crops) and compare it with SpinSCI-B in
\Cref{tab:gray,tab:color,tab:large}. $\Delta$ is full-model PSNR minus
no-MC-IRB PSNR, in dB.}
\label{tab:app-budget-pilot}
\small
\begin{tabular}{@{}lrrr@{}}
\toprule
Setting & Full PSNR & No-MC-IRB PSNR & $\Delta$ \\
\midrule
Controlled, $40$ epochs & 35.27 & 34.82 & $+0.45$ \\
Controlled, $80$ epochs & 37.07 & 36.68 & $+0.39$ \\
Controlled, $200$ epochs & 38.11 & 37.74 & $+0.37$ \\
\midrule
Full, small-scale gray & 37.51 & 36.99 & $+0.52$ \\
Full, mid-scale color & 38.24 & 37.83 & $+0.41$ \\
Full, large-scale color & 37.68 & 37.29 & $+0.39$ \\
\bottomrule
\end{tabular}
\end{table}

Under the controlled protocol, the full-model advantage is $0.45$, $0.39$,
and $0.37$~dB at $40$, $80$, and $200$ epochs (\Cref{tab:app-budget-pilot}).
Under the full training setting, SpinSCI-B leads the backbone without
MC-IRB by $0.52$, $0.41$, and $0.39$~dB on the three benchmarks. These runs
change only the MC-IRB switch of the released configurations. All rows are
evaluated with the released FP32 evaluator.

\subsection{Mask Conditioning and Operator Shift}\label{app:ablation-masks}

The mask-conditioning comparisons are reported in \Cref{tab:ablation_mask}
(Section~\ref{sec:ablation}).
Removing explicit channels does not remove all mask information: the initial
estimate and the final correction still depend on the coding masks.
Consequently, these variants test the incremental value of explicit channels
beyond the retained sensing operations.

\Cref{tab:ablation_mask} separates a new mask realization from a new
operator distribution. Both checkpoints change by at most $0.01$~dB
between the benchmark binary mask and the unseen binary mask, and the
full model keeps $37.07$~dB on both. Each checkpoint performs best on the
operator family it was trained with: binary training gives the best
binary results, and Gaussian training gives $35.96$~dB on the Gaussian
mask.

\noindent\textbf{Nonbinary Operators.}
For nonbinary masks, the mask energy is $D_p=\sum_t\Phi_{t,p}^2$.
The Gaussian-mask training sampler clips its samples to $[0,1]$ and uses
the temporal mask sum as $d_p$, with zero sums replaced by one.
This construction satisfies $0\le D_p\le d_p$, so the unit-step
data-consistency correction retains residual nonincrease under the
existing mask-sum normalization (\Cref{rem:bounded-masks}).
Equality between the normalizer and the mask energy is therefore
unnecessary for this guarantee.

\noindent\textbf{Calibrated Masks and the Real Benchmark.}
The \textbf{real benchmark} is reconstructed with separate checkpoints for
compression rates $10$ and $50$, each trained with the calibrated real mask
(Appendix~\ref{app:implementation-details}). This follows
\Cref{tab:ablation_mask}, where each checkpoint performs best on the
operator family it was trained with. As a deliberate operator mismatch,
applying the calibrated mask to reference videos with the checkpoint
trained on binary benchmark masks gives $19.04$~dB; this floating-point
operator never appeared in its training. The naturally captured
measurements lack aligned references, so \Cref{fig:real} and
Appendices~\ref{app:real-cr10}--\ref{app:real-cr50} assess them
qualitatively.

\subsection{Controls That Isolate the Mechanism}\label{app:ablation-controls}

The following definitions distinguish residual capacity, local gating, and
pairwise interactions. Matching PSNR or parameter counts alone does not
establish that two controls implement the same forward operation.

\noindent\textbf{Residual Capacity Without Adaptive Gating.}
The residual MLP, its normalization, and its zero-initialized last layer
are retained, while the gate is fixed to $g_i=1/2$ for every token.
This matches the neutral-magnetization gate value and tests whether
additional token-wise residual capacity is sufficient to explain the
gain. A comparison only against removing MC-IRB would remove this
residual path and the spin mechanism together.

\noindent\textbf{Exact No-Interaction Gate.}
The full MC-IRB retains its field projection, gate scale, residual MLP,
normalization, and initialization, but its coupling matrix is set to zero.
For $K_{\mathrm{mf}}\ge1$, the update then gives
\begin{equation}
m_i^{(K_{\mathrm{mf}})}=\tanh(\beta h_i),
\qquad
g_i=\sigma\!\left(\kappa\tanh(\beta h_i)\right).
\end{equation}
This control retains the full model's nonlinear field-to-gate mapping.
It differs from a bare $\sigma(h_i)$ gate and from $K_{\mathrm{mf}}=0$,
which gates the learned initial magnetization.
The confirmatory protocol requires training each variant from scratch.

\noindent\textbf{Unsigned Couplings With Preserved Magnitudes.}
The entrywise-absolute control applies
$J_{ij}^{\mathrm{abs}}=|J_{ij}|$ after the full coupling construction,
without recentering afterward. This preserves symmetry, the zero diagonal,
invalid-pair exclusion, and each coupling's absolute magnitude, while
removing repulsive interactions. A comparison
between independently trained variants measures the effect of the sign
constraint on the learned model; it is distinct from modifying signs only
at test time. The existing unsigned control and the entrywise-absolute
control are reported separately in \Cref{tab:ablation_irb}.

\noindent\textbf{Capacity and Cost.}
Parameters and FLOPs for the three additional controls are counted with
batch size $1$, $T=8$, and $512\times512$ inputs. Unused refiner
parameters and computations are omitted from the fixed-gate and exact
no-interaction implementations. The entrywise-absolute control retains the
full parameterization. FLOPs follow the main paper's THOP convention,
which omits functional attention and mean-field matrix products.
The ten-block backbone is a near match in parameters and FLOPs. The profiling protocol
uses the complete inference path, bypasses disabled computations in
simplified gates, and distinguishes instantiated parameters from those
that contribute to the output.

\noindent\textbf{Iteration Count.}
SpinSCI uses $K_{\mathrm{mf}}=3$; the comparisons in
\Cref{tab:ablation_irb} also consider $0$, $1$, and $5$ updates.
Increasing the update count need not monotonically improve PSNR.
Any further selection requires validation quality and measured inference
cost under the same $80$-epoch schedule. Test results are reserved for
evaluating the selected configuration.

\subsection{Confirmatory Training and Statistical Reporting}\label{app:ablation-confirm}

The confirmatory protocol uses five training seeds, fixed
before their outcomes are inspected.
\Cref{tab:app-confirmatory} reports their means, sample SDs, and paired intervals.
The full model obtains $37.07\pm0.10$~dB PSNR (mean $\pm$ sample SD).

These runs use the same ablation settings as
\Cref{sec:ablation}: DAVIS2017~\cite{davis}, $80$ training epochs,
batch size $20$, and $128\times128$ crops, with the same optimizer, loss,
learning-rate schedule, test inputs, and metric aggregation. The endpoint
is the exponential-moving-average checkpoint at $80$ epochs. This
ablation protocol is separate from the $200$-epoch final training on four
NVIDIA H200 GPUs, which uses $256\times256$ gray crops with global batch size $10$
and $512\times512$ color crops with global batch size $2$.

The paired control group comprises no MC-IRB at nine blocks, the
ten-block capacity control, the fixed-gate residual MLP, the existing
conditioned gate, the existing field-only gate, and the exact
$\mathbf{J}=0$ control. Their mean differences and confidence intervals
are reported in \Cref{tab:app-confirmatory}.
Signed-coupling and explicit-conditioning comparisons,
together with the spatial-only and temporal-only configurations, use the
same reporting rules.

The paired-comparison protocol specifies matched initialization for shared
layers of equal shape and controls data-order and augmentation randomness
independently of model construction. This preserves the paired minibatch
sequence when modules differ. Training settings and the checkpoint rule
are fixed before the runs; validation data determine model settings.
Test PSNR is not used to select checkpoints or training budgets.

Let $r_{s,v}$ denote the benchmark-average PSNR for training seed $s$ and
variant $v$, using the same aggregation rule as the main benchmark.
For a comparison with the full model, define
\begin{equation}
\delta_{s,v}=r_{s,\mathrm{full}}-r_{s,v},\qquad
\overline{\delta}_v=\frac{1}{S}\sum_{s=1}^{S}\delta_{s,v},\qquad
s_{\delta,v}=
\sqrt{\frac{1}{S-1}\sum_{s=1}^{S}
(\delta_{s,v}-\overline{\delta}_v)^2}.
\end{equation}
Model scores are reported as mean $\pm$ sample SD, with $S=5$. A paired
Student-$t$ interval for the mean PSNR difference is
\begin{equation}
\overline{\delta}_v
\ \pm\ t_{0.975,S-1}\frac{s_{\delta,v}}{\sqrt{S}}.
\end{equation}
This interval treats independent training seeds as the sampling units
and assumes that their paired differences are approximately normally
distributed. With five seeds it is a small-sample summary; individual
seed differences are not reported here. The interval is conditional on
the fixed benchmark and mask set, not a population interval over all
possible videos or sensors. Frames, patches, sequences, and masks from
the same trained model are not additional independent training replicates.

The comparison of full MC-IRB with the sigmoid gate on $\mathbf{h}$ only
is the primary accuracy contrast. The other reported intervals are also
pointwise. A joint superiority assessment requires two-sided paired tests
with Holm adjustment across the prespecified contrasts. No fixed PSNR
threshold, including $0.05$~dB, substitutes for measured uncertainty or
practical relevance.

\begin{table}[htbp]
\centering
\caption{Repeated-run evaluation of selected variants from
\Cref{tab:ablation_irb,tab:ablation_mask}. PSNR is reported as mean $\pm$
sample SD over five training seeds, at $80$ epochs.
$\Delta$ is calculated as the full-model mean minus the variant mean on
the same evaluation masks. Intervals are pointwise paired $95\%$
Student-$t$ confidence intervals.}
\label{tab:app-confirmatory}
\footnotesize
\setlength{\tabcolsep}{4pt}
\begin{tabular}{@{}lccc@{}}
\toprule
Variant & \makecell{PSNR (dB)\\mean $\pm$ sample SD} &
\makecell{Paired $\Delta$\\(dB)} & \makecell{$95\%$ interval\\for $\Delta$} \\
\midrule
Full MC-IRB, $K_{\mathrm{mf}}=3$ & $37.07\pm0.10$ & -- & -- \\
Without MC-IRB & $36.66\pm0.09$ & $+0.41$ & $[+0.226,\,+0.591]$ \\
Without MC-IRB, $Z=10$ & $36.90\pm0.08$ & $+0.17$ & $[-0.023,\,+0.355]$ \\
Residual MLP, fixed $g_i=1/2$ & $36.72\pm0.10$ & $+0.35$ & $[+0.133,\,+0.567]$ \\
Conditioned gate without mean-field inference & $36.86\pm0.07$ & $+0.21$ & $[+0.079,\,+0.350]$ \\
Sigmoid gate on $\mathbf{h}$ only & $36.81\pm0.09$ & $+0.26$ & $[+0.061,\,+0.452]$ \\
No interactions, $\mathbf{J}=0$ & $36.83\pm0.08$ & $+0.24$ & $[+0.062,\,+0.425]$ \\
Entrywise $|\mathbf{J}|$ control & $36.88\pm0.10$ & $+0.19$ & $[+0.025,\,+0.355]$ \\
\midrule
\multicolumn{4}{@{}l}{\textit{Unseen binary mask (variants from \Cref{tab:ablation_mask})}} \\
$\mathbf{\Phi},\mathbf{\Phi}_s$ removed from $\mathbf{C}$ & $36.81\pm0.08$ & $+0.26$ & $[+0.054,\,+0.470]$ \\
Also removed from the stem input & $36.72\pm0.09$ & $+0.35$ & $[+0.251,\,+0.454]$ \\
\bottomrule
\end{tabular}
\end{table}

The full-model SD describes variation across seeds; paired intervals
quantify uncertainty in the mean advantage over each control.
The primary field-only-gate comparison yields a $0.26$~dB gain with a
$95\%$ interval of $[0.061,0.452]$~dB. The $\mathbf{J}=0$ and
$|\mathbf{J}|$ comparisons give $0.24$ and $0.19$~dB gains, with intervals
of $[0.062,0.425]$ and $[0.025,0.355]$~dB, respectively. Their positive
lower bounds support pairwise and signed interactions under the stated
analysis. All listed intervals are pointwise.

\subsection{Mechanism Diagnostics and Their Interpretation}\label{app:ablation-diagnostics}

\Cref{tab:app-diagnostics} reports fixed-checkpoint measurements over all
192 frames in the six mid-scale color sequences. Spatial and temporal
statistics pool valid positions across nine blocks; spatial padding is
excluded. The reference reconstruction reproduces the Table~\ref{tab:color} average of $38.2430$~dB before rounding. Selected frames and the permutation seed
are fixed before inspecting the resulting error maps.

\noindent\textbf{Interaction Activity and Feature Correction.}
Negative entries constitute $51.95\%$ and $55.46\%$ of nonzero valid
spatial and temporal couplings, so both branches learn signed interactions.
The learned scales concentrate pairwise interaction in the temporal branch:
spatial $\gamma$ is zero in seven blocks, the third block has $\kappa=0$,
and the first has $\kappa=0.009$, so spatial coupling interventions leave
PSNR unchanged to four decimal places at this checkpoint. Spatial MC-IRB
acts here through its sensing-conditioned gates and residual MLP; in
\Cref{tab:ablation}, adding spatial MC-IRB to the backbone improves PSNR
by $0.16$~dB. The mean relative gated-residual norms are $0.219$ spatially
and $0.242$ temporally, so both branches apply substantial feature
corrections.

\noindent\textbf{Reliance on Temporal Pair Structure.}
Setting temporal $\mathbf{J}=0$, replacing it with $|\mathbf{J}|$, or
permuting it reduces PSNR by $0.101$, $0.795$, and $1.036$~dB,
respectively. Each intervention changes only the named branch.
The fixed permutation preserves valid-pair masks, padding membership,
symmetry, and spectrum while leaving features and local fields in place.
These results show that reconstruction relies on the learned temporal
interactions.

\begin{figure}[H]
\centering
\includegraphics[width=\columnwidth]{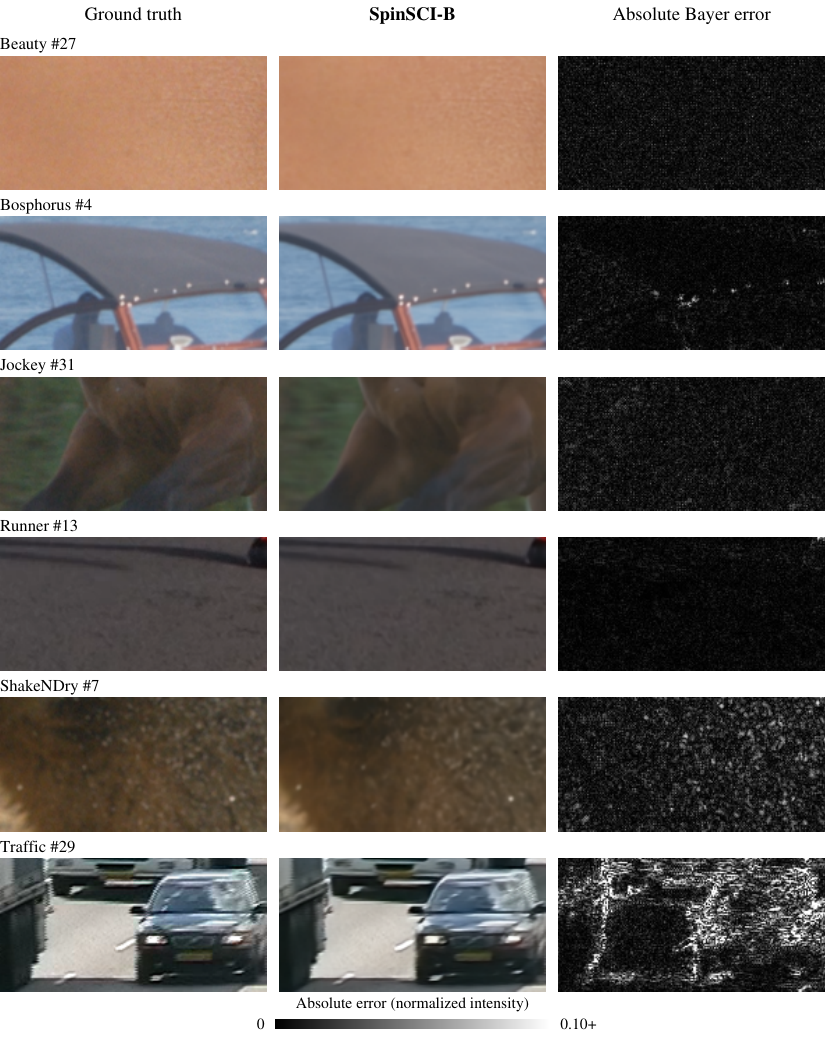}
\caption{Ground truth, \textbf{SpinSCI-B} reconstructions, and absolute
Bayer-domain reconstruction errors for the six fixed frames in the
\textbf{mid-scale color} benchmark; $\#$ denotes the frame number.
Each panel uses the same central $192\times96$ crop. Error maps use a
shared intensity scale from $0$ to $0.10$, with larger errors shown in
white. Errors are computed before display clipping; PSNR in
\Cref{tab:app-diagnostics} is averaged over all frames and full images
of each sequence.}
\label{fig:supp-b6-errors}
\end{figure}

\begin{figure}[t]
\centering
\includegraphics[width=0.9\columnwidth,trim=0 6.5 0 8,clip]{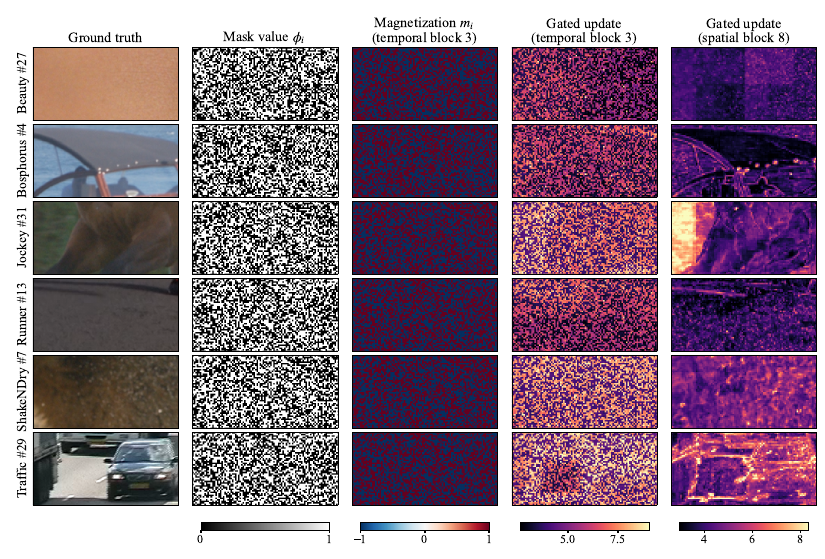}
\vspace{-0.8em}
\caption{MC-IRB states of \textbf{SpinSCI-B} on the six fixed
\textbf{mid-scale color} frames of \Cref{fig:supp-b6-errors}, using the same
central $192\times96$ crop. Columns show the ground truth, the mask value
$\phi_i$ on the token grid ($1$ for a measured token), the final magnetization
$m_i$ and the gated update $g_i\|\mathcal{R}(\mathbf{p}_i)\|_2$ of temporal
block~3, and the gated update of spatial block~8. In temporal block~3,
$m_i\approx1-2\phi_i$, so the gate $g_i=\sigma(\kappa m_i)$ is $0.41$ for
measured and $0.59$ for unmeasured tokens. Each column uses one color scale
for all six frames; update scales span the 1st to 99th percentiles.}
\label{fig:supp-mcirb-maps}
\end{figure}

\noindent\textbf{Magnetization and Gate Maps.}
\Cref{fig:supp-mcirb-maps} shows MC-IRB states for the six fixed frames of
\Cref{fig:supp-b6-errors}. Temporal block~3 learns a large mask weight,
$\lambda_\phi=3.39$, and its couplings are organized by exposure state: over
all 192 frames, the mean coupling is $+0.35$ between frames that share a
token's mask value and $-0.53$ between frames that do not. The magnetizations
settle into the matching two-group configuration, $m_i\approx-1$ for measured
tokens ($\phi_i=1$) and $m_i\approx+1$ for unmeasured tokens ($\phi_i=0$),
with Pearson correlation $-1.00$ between $m_i$ and $\phi_i$. Their signs agree
with couplings that carry $99.9\%$ of $\sum_{i,j}|J_{ij}|$, so the mean-field
state nearly minimizes the pairwise term of \Cref{eq:ising-energy}. The block
therefore identifies, in every frame, the tokens that lack a measurement and
refines them more strongly: their gates are $0.59$ rather than $0.41$, and
their gated updates are $1.93\times$ larger. The small-scale gray checkpoint
learns the same organization in temporal block~2, with $99.9\%$ of the
coupling weight satisfied and $1.56\times$ larger updates for unmeasured
tokens. Mask transitions, in contrast, do not organize the gates: as a score for
tokens whose mask value changes between consecutive frames, the gates of
these two blocks reach an AUC of $0.49$, and the gates of all blocks stay
within $0.45$--$0.54$, whereas the same gates identify unmeasured tokens
with an AUC of $1.00$. Deeper blocks direct their updates to image structure: in spatial
block~8, the Spearman correlation between gated updates and ground-truth
gradient magnitude is $0.41$ for mid-scale color and $0.73$ for small-scale
gray, and \Cref{fig:supp-mcirb-maps} shows these updates concentrating on
object contours and textured regions.

\noindent\textbf{Reconstruction Errors.}
\Cref{fig:supp-b6-errors} shows ground-truth and full-model reconstruction
crops from the six fixed frames, together with absolute Bayer-domain
error maps on a shared scale. \Cref{tab:app-diagnostics} reports the
corresponding per-sequence PSNR.

\noindent\textbf{Effect of Final Data Consistency.}
For $\rho(z)=\|\mathbf{y}-\mathcal{A}_{\Phi}z\|_2/
(\|\mathbf{y}\|_2+\epsilon)$, the final correction reduces mean $\rho$
from $3.475\times10^{-3}$ to $5.307\times10^{-8}$, consistent with the
binary-mask sensing identity up to numerical precision. PSNR increases
from $38.2111$ to $38.2430$~dB. Both changes are measured directly. Applying the same correction to the official HiSViT$_9$ checkpoints on the
benchmark masks raises their PSNR by only $0.04$~dB on small-scale gray
($37.00\to37.03$~dB) and $0.03$~dB on mid-scale color ($38.01\to38.04$~dB),
so SpinSCI-B leads by $0.48$ and $0.20$~dB when both methods use it.

\noindent\textbf{Convergence on Trained Checkpoints.}
\Cref{app:formal} establishes sensing identities, coupling symmetry,
finite-step bounds, and a sufficient convergence condition
(\Cref{thm:contraction}). Evaluated over every refinement group of every
test snapshot of the released gray and color checkpoints, the condition
$\beta\|\mathbf{J}\|_2<1$ holds throughout in $9$ of the $12$ temporal
blocks with active coupling, so their mean-field updates are contractions
with a unique fixed point. In the other three, the Jacobian of the update
at the reached magnetizations has spectral norm below one in more than
$99.999\%$ of groups. These three include the strongly coupled first
temporal block of each model, whose three-step magnetizations satisfy the
fixed-point equation to within $0.02$ in every group and form the
two-group configuration of \Cref{fig:supp-mcirb-maps}.

\begin{table}[H]
\centering
\caption{Fixed-checkpoint mechanism diagnostics on the six mid-scale color
sequences. Statistics pool valid positions across blocks and examples;
coupling fractions exclude diagonal and zero entries. Parameter ranges
follow the listed order, and SDs describe token variation. Each intervention
changes only the named branch. Per-sequence PSNR values are listed for Beauty,
Bosphorus, Jockey, Runner, ShakeNDry, and Traffic, in that order.
\Cref{fig:supp-b6-errors} shows reconstruction
crops and absolute error maps.}
\label{tab:app-diagnostics}
\footnotesize
\setlength{\tabcolsep}{4pt}
\begin{tabular}{@{}p{70mm}rr@{}}
\toprule
\multicolumn{3}{@{}l}{\textit{Internal MC-IRB statistics}} \\
Measurement & Spatial & Temporal \\
\midrule
Effective $\beta,\gamma,\kappa$ (range for each) & \makecell[r]{$[0.168,\,1.594]$\\$[0.000,\,0.928]$\\$[0.000,\,3.576]$} & \makecell[r]{$[0.179,\,1.203]$\\$[0.000,\,0.359]$\\$[0.052,\,3.718]$} \\
Effective $\lambda_\phi,\lambda_u,\lambda_d$ (range for each) & \makecell[r]{$[0.000,\,3.638]$\\$[0.000,\,0.753]$\\$[0.000,\,0.011]$} & \makecell[r]{$[0.000,\,5.455]$\\$[0.000,\,0.630]$\\$[0.000,\,0.000]$} \\
Negative-coupling fraction (\%) & 51.95 & 55.46 \\
$\operatorname{RMS}(\mathbf{h})$ & 4.540 & 1.650 \\
$\operatorname{RMS}(\mathbf{J}\mathbf{m}^{(K_{\mathrm{mf}}-1)})$ & 24.720 & 1.103 \\
Final magnetization $m_i$ (mean $\pm$ SD) & $0.455\pm0.699$ & $0.099\pm0.690$ \\
Gate $g_i$ (mean $\pm$ SD) & $0.659\pm0.179$ & $0.573\pm0.188$ \\
Fraction with $|m_i|>0.95$ (\%) & 22.91 & 21.29 \\
Fraction with $g_i<0.05$ (\%) & 0.00 & 0.00 \\
Fraction with $g_i>0.95$ (\%) & 1.95 & 0.86 \\
Mean $\|\mathcal{R}(\mathbf{p}_i)\|_2$ & 3.985 & 7.110 \\
Mean $g_i\|\mathcal{R}(\mathbf{p}_i)\|_2$ & 2.829 & 4.075 \\
Mean $g_i\|\mathcal{R}(\mathbf{p}_i)\|_2/(\|\mathbf{p}_i\|_2+\epsilon)$ & 0.219 & 0.242 \\
\midrule
\multicolumn{3}{@{}l}{\textit{Fixed-checkpoint coupling interventions}} \\
Evaluation condition & \makecell[r]{Spatial change\\PSNR (dB)} & \makecell[r]{Temporal change\\PSNR (dB)} \\
\midrule
Unmodified checkpoint (shared reference) & 38.2430 & 38.2430 \\
$\mathbf{J}=0$ & 38.2430 & 38.1423 \\
Entrywise $|\mathbf{J}|$ & 38.2430 & 37.4478 \\
$P\mathbf{J}P^\top$ (validity-preserving permutation) & 38.2430 & 37.2072 \\
\midrule
\multicolumn{3}{@{}l}{\textit{Full-model reconstruction quality}} \\
Model & \multicolumn{2}{r@{}}{Per-sequence PSNR (dB)} \\
\midrule
Full model & \multicolumn{2}{r@{}}{\makecell[r]{38.04 / 42.05 / 39.30\\43.51 / 35.83 / 30.72}} \\
\midrule
\multicolumn{3}{@{}l}{\textit{Final data-consistency correction}} \\
Output & PSNR (dB) & Mean $\rho$ \\
\midrule
Decoder prediction $\widetilde{\mathbf{x}}$ & 38.2111 & $3.475\times10^{-3}$ \\
Corrected video $\widehat{\mathbf{x}}$ & 38.2430 & $5.307\times10^{-8}$ \\
\bottomrule
\end{tabular}
\end{table}

\subsection{Mask Calibration Error}\label{app:ablation-calibration}

In this inference-only test, on the \textbf{mid-scale
color} benchmark (CR $8$, six videos), each measurement is formed with
the true binary mask $\mathbf{\Phi}$, and each method reconstructs with a
miscalibrated mask $\mathbf{\Phi}'$ wherever it uses the mask
(initialization, network input, data consistency and the mask sum).
S$\delta$ shifts $\mathbf{\Phi}$ horizontally by $\delta$ pixels
(bilinear interpolation, circular boundary), and N$\sigma$ uses
$\mathrm{clip}(\mathbf{\Phi}+\sigma\mathbf{n},0,1)$ with one fixed
standard Gaussian array $\mathbf{n}$ for every method. All methods use
their official checkpoints in FP32. C0 values are our re-runs with the
true mask, so baselines can differ from the published scores in
\Cref{tab:color}; SpinSCI-B and HiSViT$_9$ reproduce \Cref{tab:color}
within $0.001$~dB.

Under sub-pixel shifts, SpinSCI-B degrades like EfficientSCI-B, HiSViT$_9$ and DU-9stg and keeps the
highest PSNR among them (\Cref{tab:app-calibration}: S$0.25$ 24.62 vs
24.51--24.57~dB, S$0.5$ 18.76 vs 18.04--18.32~dB). Under noise on the mask values
it loses more than these three (N$0.05$: 34.13 vs 34.24--34.77~dB). Its final data-consistency step
enforces $\mathbf{\Phi}'$ exactly; disabling it raises PSNR by
$0.34$--$0.41$~dB under the other perturbations but changes it by
$0.01$~dB at N$0.05$, so the step does not explain the loss at N$0.05$. STFormer-B and
MambaSCI-B fall below $23$~dB under any nonbinary $\mathbf{\Phi}'$. Consistent with
\Cref{tab:ablation_mask}, every learned method needs the reconstruction
mask to match the sensing operator; for the real benchmark SpinSCI uses
checkpoints trained with the calibrated mask.

\begin{table}[H]
\centering
\caption{Average PSNR (dB) under mask calibration error on the
\textbf{mid-scale color} benchmark (CR $8$). The measurement is formed
with the true mask; reconstruction uses the perturbed mask. C0: true
mask. S$\delta$: horizontal shift by $\delta$ pixels. N$\sigma$: additive
Gaussian noise of standard deviation $\sigma$ on the mask values, clipped
to $[0,1]$.}
\label{tab:app-calibration}
\footnotesize
\setlength{\tabcolsep}{4pt}
\begin{tabular}{@{}lrrrrr@{}}
\toprule
Method & C0 & S$0.25$ & S$0.5$ & N$0.05$ & N$0.2$ \\
\midrule
PnP-FastDVDNet & 35.31 & 16.44 & 11.52 & 25.89 & 19.90 \\
EfficientSCI-B & 37.24 & 24.57 & 18.32 & 34.24 & 26.60 \\
MambaSCI-B & 35.70 & 16.65 & 14.46 & 19.86 & 18.33 \\
STFormer-B & 37.53 & 14.98 & 12.08 & 22.82 & 17.31 \\
HiSViT$_9$ & 38.01 & 24.51 & 18.04 & 34.77 & 26.92 \\
DU-9stg & 38.12 & 24.55 & 18.30 & 34.71 & 27.49 \\
\rowcolor{green!5}
\textbf{SpinSCI-B} & 38.24 & 24.62 & 18.76 & 34.13 & 26.40 \\
SpinSCI-B, final DC off & 38.21 & 24.98 & 19.10 & 34.12 & 26.81 \\
\bottomrule
\end{tabular}
\end{table}

%% file: sections/implementation_details.tex
\section{Implementation and Evaluation Details}\label{app:implementation-details}

This appendix specifies dataset dimensions, implementation details, and the evaluation protocol.

\noindent\textbf{Datasets.} Training uses DAVIS2017~\cite{davis}. The evaluation uses common video SCI datasets~\cite{pnp,efficient,stformer,mamba}. The \textbf{small-scale gray} benchmark comprises six sequences (Aerial~\cite{admm}, Crash~\cite{admm}, Drop~\cite{desci}, Kobe~\cite{gmm}, Runner~\cite{desci}, and Traffic~\cite{gmm}) at $256\times256$ resolution and compression rate $8$. The \textbf{mid-scale color} benchmark comprises six sequences~\cite{pnp} (Beauty, Bosphorus, Jockey, Runner, ShakeNDry, and Traffic) of size $512\times512\times3\times32$ (RGB, 32 frames). The \textbf{large-scale color} benchmark comprises four sequences~\cite{pnp}: Football ($822\times1920\times3\times48$), Hummingbird ($1080\times1920\times3\times40$), Messi ($1080\times1920\times3\times48$), and Swinger ($1080\times1920\times3\times40$). The \textbf{real benchmark} uses Hand~\cite{pnp} ($512\times512\times40$; Figure~\ref{fig:intro}), Water Balloon~\cite{e2ecnn} ($512\times512\times10$) and Domino~\cite{e2ecnn} ($512\times512\times60$) (Figure~\ref{fig:real}), and Pendulum Ball ($512\times512\times40$; Appendix~\ref{app:real-cr10}), all at a compression rate of $10$. Appendices~\ref{app:real-cr10}--\ref{app:real-cr50} provide the reconstructions at compression rates of $10$ and $50$.

\noindent\textbf{Evaluation Metrics.} PSNR and SSIM~\cite{ssim} are computed using the evaluation code provided by~\citet{cactilib}; efficiency is assessed by parameter count, FLOPs, inference latency, and peak memory.

\noindent\textbf{Attention and Feed-Forward Implementation.} The pooled spatial heads use $8\times8$ key windows with pooling factors two and four, corresponding to $16\times16$ and $32\times32$ query windows. They retain queries at token resolution and do not use native-window cyclic shifts. Spatial MC-IRB operates within native $7\times7$ windows using logits from the two unpooled heads. Bottom and right zero padding completes attention and refinement windows; padded positions participate in both operations, and $v_{ij}$ excludes only shifted-window wrap-around pairs. Temporal projections use inner width $d_{\mathrm{feat}}/2$ and return to $d_{\mathrm{feat}}$ before refinement. The inverted residual FFN comprises pointwise expansion to $2d_{\mathrm{feat}}$, a depthwise $3\times3\times3$ convolution, and pointwise projection, with GELU after the first two operations.

\noindent\textbf{Bayer Implementation.} RGGB selectors implement the sampling and correction operators in \Cref{app:bayer}, preserving the selected-color update.

\noindent\textbf{Component Costs.} Spatial attention products cost $O\!\left(N(d_{\mathrm{feat}}/4)(2\cdot49+2\cdot64)\right)$, ignoring boundary-padding overhead: two heads use $49$ native keys per query and two use $64$ pooled keys. The linear projections cost $O(Nd_{\mathrm{feat}}^2)$. Temporal attention adds $O(NTd_{\mathrm{feat}})$ for its attention products. Spatial and temporal MC-IRBs together cost $O\!\left(64Nd_{\mathrm{feat}}+N(1+K_{\mathrm{mf}})(49+T)\right)$, with $64$ hidden units in each residual MLP. The inverted residual FFN costs $O(Nd_{\mathrm{feat}}^2)$ for its fixed expansion ratio and convolution kernel size.

\noindent\textbf{Efficiency Protocol.} All cross-method runtime comparisons in the main paper (\Cref{fig:efficiency} and \Cref{tab:gray,tab:color,tab:large}) use FP32 eager inference on an NVIDIA GH200 (144\,GB), batch size~$1$, and $T=8$. cuDNN benchmarking and matrix-multiplication TF32 are disabled; cuDNN TF32 is enabled. Each native forward is timed from its GPU-resident input using a CUDA-synchronized wall clock. HiSViT$_9$ starts from its prepared network input. Peak memory is the maximum GPU memory allocated after warm-up, including weights and resident inputs. Checkpoint loading and file I/O are excluded. The MC-IRB ablation in \Cref{tab:ablation}, which compares SpinSCI variants only, uses BF16 compiled inference (Appendix~\ref{app:compiled}).

Gray and mid-scale color use $256\times256$ and $512\times512$ frames. EfficientSCI-B, STFormer-B, HiSViT$_9$, and MambaSCI-B use three warm-ups, followed by $40$ timed calls on gray or $20$ on mid-scale color where evaluated. SpinSCI-B uses three warm-ups and ten timed calls, DU-9stg five and twenty, and PnP-FastDVDNet one and three. Large-scale color reconstructs a $1080\times1920$ snapshot, padded to height $1088$, as $24$ sequential $272\times320$ tiles whose outputs are assembled on the GPU. Latency and peak memory cover this complete reconstruction, with the inputs and assembled output resident. On large-scale color, EfficientSCI-B, STFormer-B, HiSViT$_9$, and MambaSCI-B use three warm-ups and ten timed calls.

Parameters count all instantiated model parameters. FLOPs count one THOP module-level multiply--accumulate as one FLOP, omitting unregistered functional attention products and elementwise operations. SpinSCI uses the equivalent cuDNN depthwise convolution for counting and the Triton depthwise implementation for timing. Large-scale FLOPs cover a full frame; for HiSViT$_9$ and PnP-FastDVDNet, they cover the complete tiled reconstruction.

PnP-FastDVDNet is timed over its complete iterative reconstruction: $90$ ADMM iterations for grayscale, or $90$ GAP iterations with CPU demosaicing for color. Its FLOPs use the PyTorch operator-level counter over the same complete reconstruction, again counting one multiply--accumulate as one FLOP. At $256\times256$, this includes $90$ denoiser passes of $329$~G each; a per-pass figure would understate total cost by that factor.

\noindent\textbf{Checkpoint and Quality Provenance.} Small-scale gray and mid-scale color baseline PSNR/SSIM are taken from the respective papers, except as noted below for DU-9stg. SpinSCI uses the benchmark masks and test videos distributed with the official HiSViT release; the files have identical SHA-256 hashes. With these masks, the official HiSViT$_9$ checkpoints, evaluated with the same code and FP32 eager settings as SpinSCI-B, reproduce the reported PSNR within $0.005$~dB (\Cref{tab:hisvit-measured}). SpinSCI-B and HiSViT$_9$ are therefore compared on identical snapshot measurements. \Cref{tab:hisvit-measured} extends this check to every small- and mid-scale baseline, with the mask file of its official test configuration and with the benchmark masks. Only the grayscale configurations of PnP-FastDVDNet and EfficientSCI-B and the DU-9stg repository use other mask files; with them, only the grayscale DU-9stg result matches its published value. MambaSCI is designed for quad-Bayer color measurements and its paper does not report grayscale results, so it is omitted from the grayscale benchmark. Large-scale baseline quality uses official-checkpoint evaluations at compression rate $8$.

SpinSCI-B uses the released non-EMA weights and FP32 eager inference on the simulated benchmarks. The mean PSNR/SSIM is $37.51/0.980$ for small-scale gray (160 frames), $38.24/0.973$ for mid-scale color (192 frames), and $37.68/0.979$ for large-scale color (176 frames). Gray and mid-scale color use the benchmark masks. The large-scale binary masks use seed $2447$ and are shared by SpinSCI-B, HiSViT$_9$, and DU-9stg. Color metrics use the RGGB mosaic of unclipped SpinSCI-B predictions. Frames are averaged within each sequence, then sequences are weighted equally. Large-scale SpinSCI-B quality uses full-frame inference.

\begin{table}[!htbp]
\centering
\caption{Published versus measured average PSNR (dB) of the small- and mid-scale baselines. Measurements use the official checkpoints, FP32 eager inference, and the test videos and metrics of SpinSCI-B. Official mask: the mask file of the method's official test configuration; $=$: identical to the benchmark mask file (same SHA-256 hash). Parentheses: difference from the published value. Benchmark mask: the masks of \Cref{tab:gray,tab:color}, distributed with the official HiSViT release. PnP-FastDVDNet uses the CACTI toolbox implementation; its grayscale mask is the one stored in the six test files. The DU-9stg color configuration draws a random mask, so its official mask is the color mask file supplied with the repository. Under the CACTI SSIM implementation applied to SpinSCI-B, HiSViT$_9$ obtains $0.977$ (small-scale gray) and $0.972$ (mid-scale color); the official HiSViT test script gives $0.978$ on small-scale gray.}
\label{tab:hisvit-measured}
\small
\setlength{\tabcolsep}{6pt}
\begin{tabular}{@{}llccc@{}}
\toprule
Benchmark & Method & Published & Official mask & Benchmark mask \\
\midrule
Small-scale gray & PnP-FastDVDNet & 32.35 & 31.79 ($-0.56$) & 32.53 ($+0.18$) \\
 & EfficientSCI-B & 36.48 & 36.39 ($-0.09$) & 36.35 ($-0.13$) \\
 & STFormer-B & 36.34 & $=$ & 36.25 ($-0.09$) \\
 & HiSViT$_9$ & 37.00 & $=$ & 37.00 (0.00) \\
 & DU-9stg & 37.57 & 37.57 (0.00) & 37.52 ($-0.05$) \\
\midrule
Mid-scale color & PnP-FastDVDNet & 34.60 & $=$ & 35.31 ($+0.71$) \\
 & EfficientSCI-B & 37.49 & $=$ & 37.24 ($-0.25$) \\
 & MambaSCI-B & 35.70 & $=$ & 35.70 (0.00) \\
 & STFormer-B & 37.32 & $=$ & 37.53 ($+0.21$) \\
 & HiSViT$_9$ & 38.01 & $=$ & 38.01 (0.00) \\
 & DU-9stg & 38.30 & 38.12 ($-0.18$) & 38.12 ($-0.18$) \\
\bottomrule
\end{tabular}
\end{table}

\noindent\textbf{DU-9stg Results.} \Cref{tab:hisvit-measured} compares reported DU-9stg PSNR with FP32 eager checkpoint measurements under the shared evaluation protocol. Reported gray PSNR/SSIM are retained because the average PSNR difference is below $0.1$~dB; measured mid-scale color PSNR/SSIM are used because the difference exceeds this threshold. The color gap persists with the supplied mask and remains unexplained. Large-scale scores ($37.80$~dB) are measured because this four-sequence benchmark is unreported.

\noindent\textbf{Real-Benchmark Protocol.} Baseline comparisons use only officially released checkpoints for real-data reconstruction. SpinSCI uses EMA checkpoints trained with the calibrated real mask, at $260{,}000$ updates for compression rate $10$ and $130{,}000$ updates for compression rate $50$, with FP32 eager inference.

\noindent\textbf{Training Details.} Muon~\cite{muon} updates eligible matrix weights; AdamW with zero weight decay updates the remaining parameters, yielding the Adam~\cite{adam} update. Mixed-precision training uses BF16 computation and FP32 accumulation. Earlier video SCI work~\cite{effipp} reports only minor PSNR reduction under FP16/FP32 mixed precision.

\subsection{BF16 Compiled Inference}\label{app:compiled}
BF16 compiled inference wraps each model's forward pass in \texttt{torch.compile} with its default Inductor backend and runs it under \texttt{torch.autocast} with BF16.

\begin{figure}[H]
\centering
\includegraphics[width=.78\linewidth]{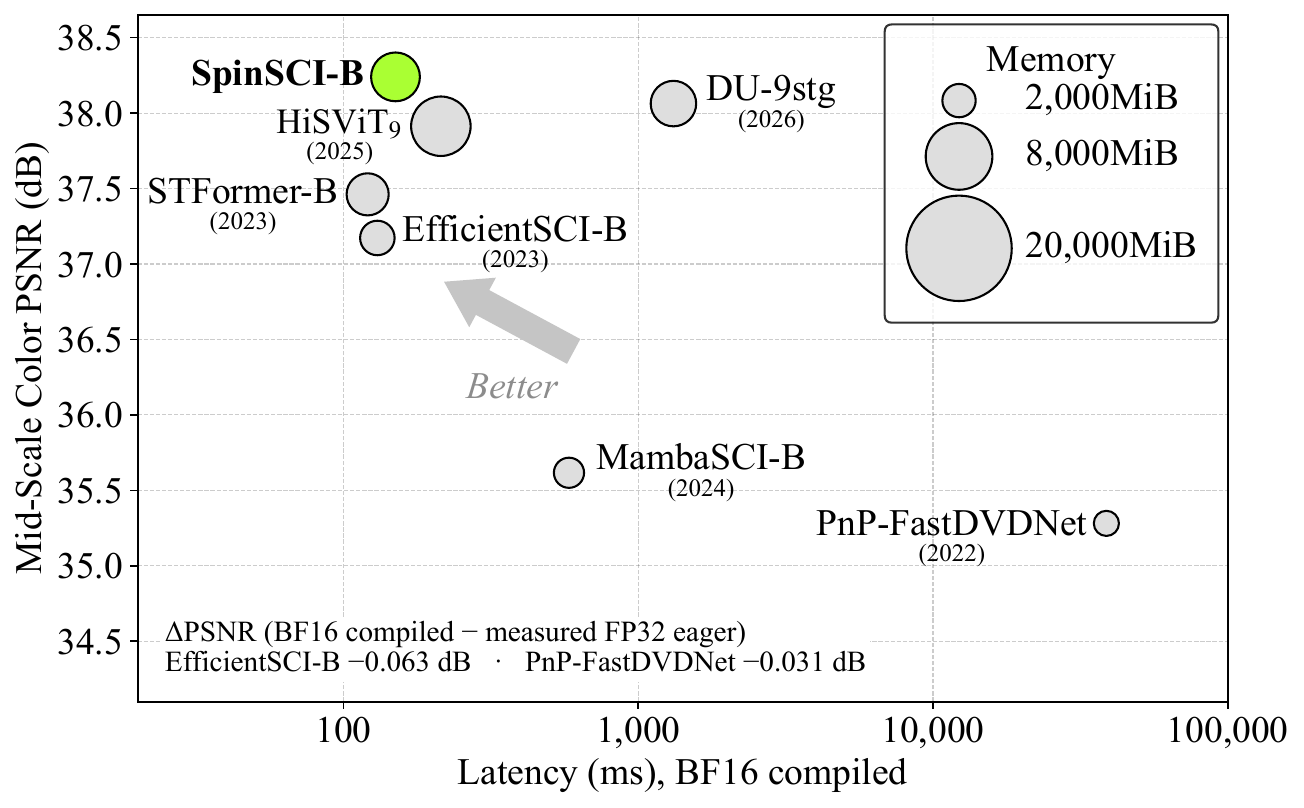}
\caption{Mid-scale color PSNR versus inference latency under BF16 autocast and default Inductor compilation for all seven methods. Circle area is proportional to peak allocated GPU memory. The annotated PSNR differences compare BF16 compiled and FP32 eager checkpoint evaluations on the same six sequences.}
\label{fig:bf16-color}
\end{figure}

\begin{table}[H]
\centering
\caption{Mid-scale color reconstruction under BF16 compiled inference. Each score is a six-sequence average. Latency is per snapshot and memory is peak GPU allocation.}
\label{tab:bf16-color}
\small
\setlength{\tabcolsep}{7pt}
\begin{tabular}{@{}lrrrr@{}}
\toprule
Method & PSNR (dB) & SSIM & Latency (ms) & Memory (MiB) \\
\midrule
PnP-FastDVDNet & 35.28 & 0.958 & 38,732.4 & 1,139 \\
EfficientSCI-B & 37.17 & 0.968 & 130.2 & 2,143 \\
MambaSCI-B & 35.62 & 0.958 & 581.7 & 1,639 \\
STFormer-B & 37.46 & 0.970 & 120.7 & 3,188 \\
HiSViT$_9$ & 37.91 & 0.972 & 213.8 & 6,368 \\
DU-9stg & 38.06 & 0.972 & 1,315.6 & 3,726 \\
SpinSCI-B & 38.24 & 0.973 & 150 & 4,284 \\
\bottomrule
\end{tabular}
\end{table}

\noindent Each timing uses five warm-ups and $20$ timed reconstructions after compilation on an NVIDIA GH200. SpinSCI-B averages $38.243$~dB in FP32 eager inference and $38.237$~dB in BF16 compiled inference. EfficientSCI-B and PnP-FastDVDNet lose $0.063$ and $0.031$~dB, respectively, relative to their measured FP32 eager scores. MambaSCI-B uses native quad-Bayer measurements; the other methods use RGGB.

\noindent\textbf{MC-IRB ablation.} Under BF16 autocast and default Inductor compilation, the backbone, temporal-only, spatial-only, and full variants take $117$, $134$, $135$, and $150$~ms, respectively. Their peak allocations are $4{,}280$, $4{,}282$, $4{,}282$, and $4{,}284$~MiB.

%% file: sections/additional_results.tex
\section{Additional Qualitative Results}\label{app:qualitative}
\numberwithin{figure}{section}

\begin{figure}[!htbp]
    \centering
    \includegraphics[width=\columnwidth,height=0.78\textheight,keepaspectratio]{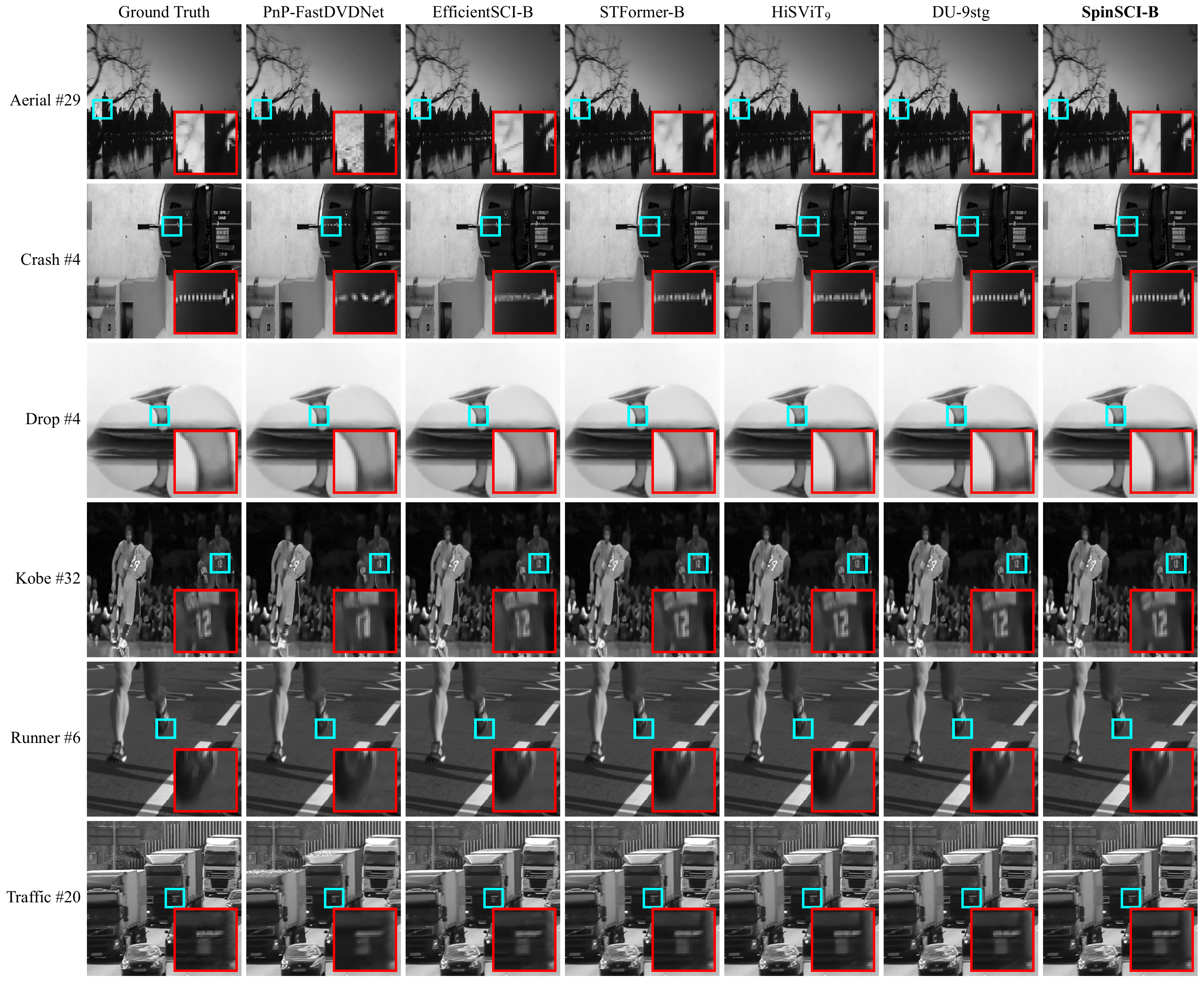}
    \caption{Qualitative comparisons covering all six \textbf{small-scale gray} benchmark sequences: Aerial, Crash, Drop, Kobe, Runner, and Traffic. Nearest-neighbor interpolation is used to zoom in on the cropped regions. $\#$ denotes the frame number.}
    \label{fig:supp-gray}
\end{figure}

\begin{figure}[!htbp]
    \centering
    \includegraphics[width=\columnwidth,height=0.78\textheight,keepaspectratio]{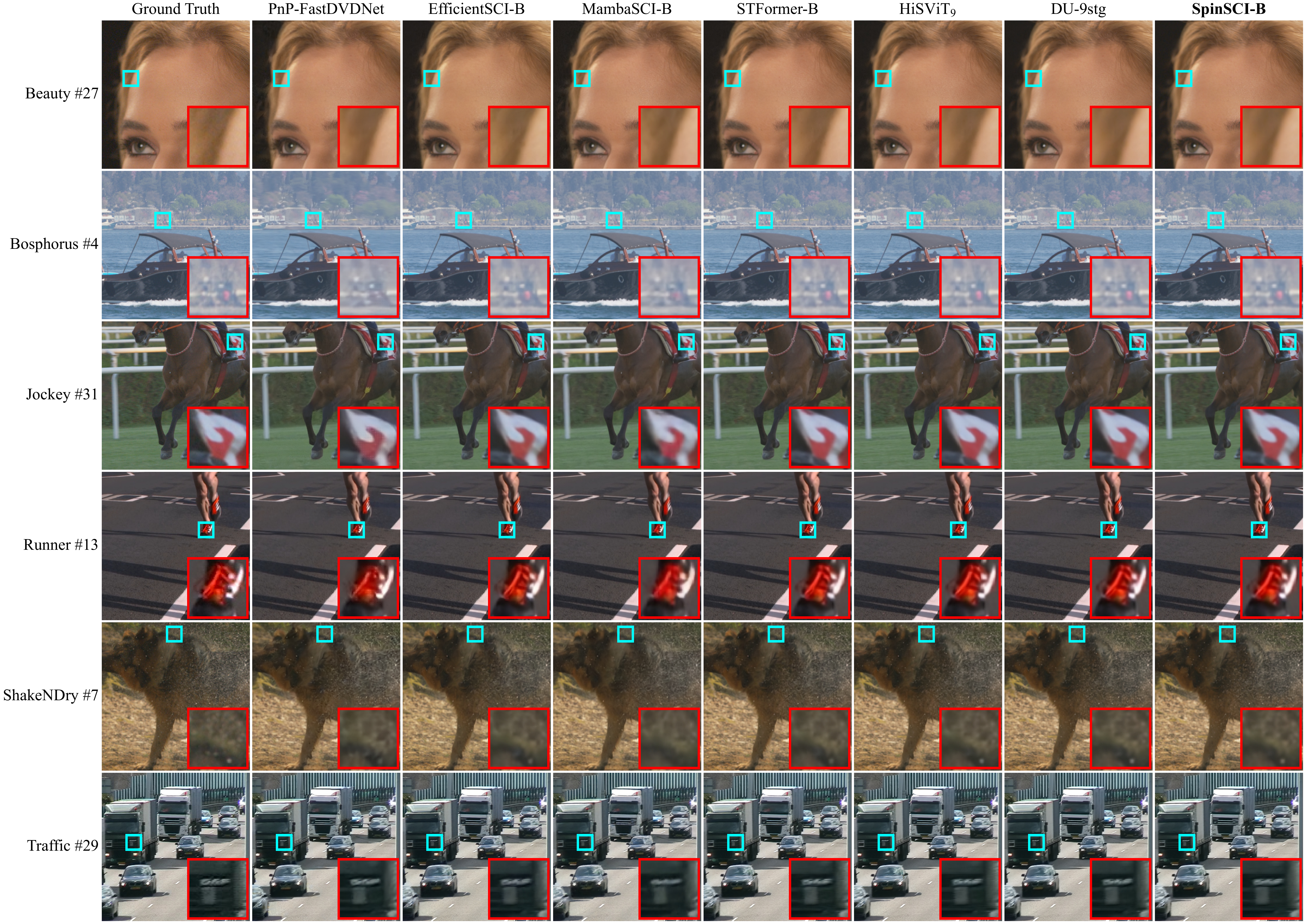}
    \caption{Qualitative comparisons covering all six \textbf{mid-scale color} benchmark sequences: Beauty, Bosphorus, Jockey, Runner, ShakeNDry, and Traffic. Nearest-neighbor interpolation is used to zoom in on the cropped regions. $\#$ denotes the frame number.
    }
    \label{fig:supp-color}
\end{figure}

\begin{figure}[!htbp]
    \centering
    \includegraphics[width=\columnwidth,height=0.78\textheight,keepaspectratio]{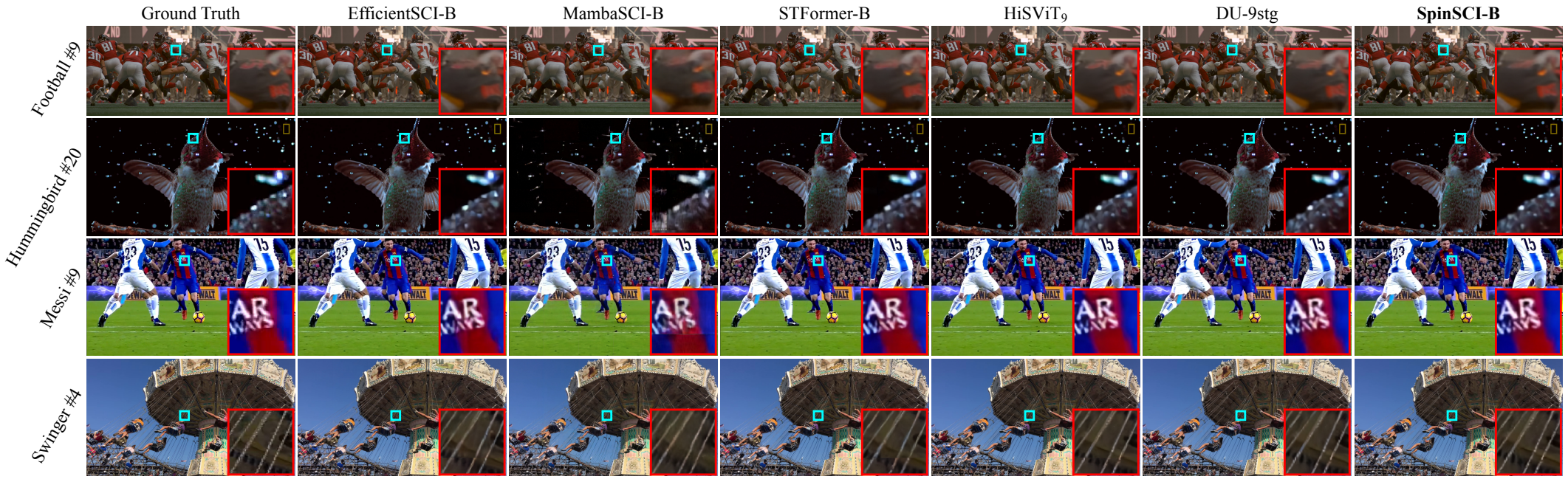}
    \caption{Qualitative comparisons covering all four \textbf{large-scale color} benchmark sequences: Football, Hummingbird, Messi, and Swinger. Nearest-neighbor interpolation is used to zoom in on the cropped regions. $\#$ denotes the frame number.}
    \label{fig:supp-large}
\end{figure}

This section presents full-frame reconstructions for the Domino~\cite{e2ecnn}, Hand~\cite{pnp}, Pendulum Ball, and Water Balloon~\cite{e2ecnn} datasets in the \textbf{real benchmark} at compression rates of $10$ and $50$. Real SpinSCI reconstructions use EMA weights with FP32 eager inference. \Cref{fig:supp-gray,fig:supp-color,fig:supp-large} provide qualitative comparisons covering all sequences in the \textbf{small-scale gray}, \textbf{mid-scale color}, and \textbf{large-scale color} benchmarks, respectively, with one selected frame per sequence. Simulated SpinSCI results use the ordinary model weights evaluated in Tables~\ref{tab:gray}--\ref{tab:large}. Appendix~\ref{app:other-cr} tests how the real-benchmark checkpoints carry to other compression rates.

\subsection{Real Benchmark at Compression Rate 10}\label{app:real-cr10}

Real-benchmark comparisons include only baselines with officially released checkpoints for real-data reconstruction.

\begin{figure}[!htbp]
    \centering
    \includegraphics[width=\columnwidth,height=0.70\textheight,keepaspectratio]{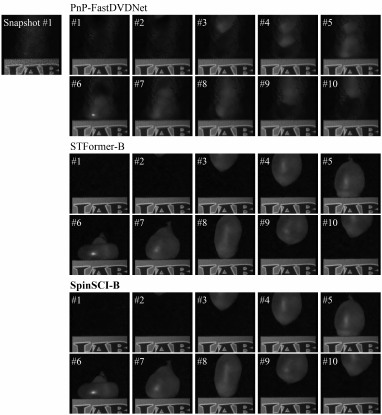}
    \caption{The first column shows snapshots from the Water Balloon dataset~\cite{e2ecnn} ($512\times512\times10$) in the \textbf{real benchmark} at a compression rate of $10$. Columns 2--6 show the corresponding reconstructed frames. Here, $\#$ denotes the frame number.
    }
    \label{fig:supp-waterballoon-cr10}
\end{figure}

\clearpage

\begin{figure}[!htbp]
    \centering
    \includegraphics[width=1.0\columnwidth,height=0.92\textheight,keepaspectratio]{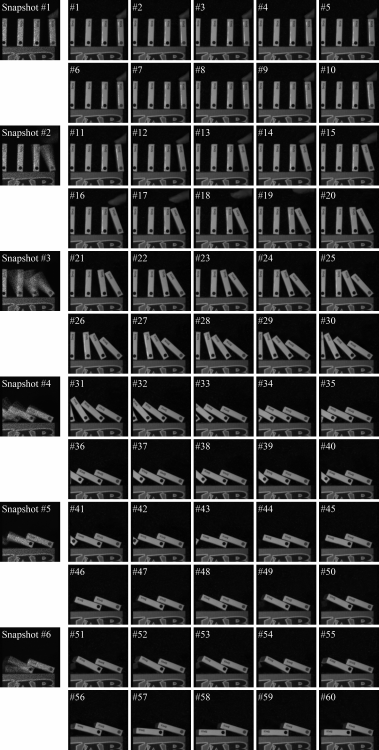}
    \caption{The first column shows snapshots from the Domino dataset~\cite{e2ecnn} ($512\times512\times60$) in the \textbf{real benchmark} at a compression rate of $10$. Columns 2--6 show the corresponding reconstructed frames. Here, $\#$ denotes the frame number.
    }
    \label{fig:supp-domino}
\end{figure}

\clearpage

\begin{figure}[!htbp]
    \centering
    \includegraphics[width=1.0\columnwidth]{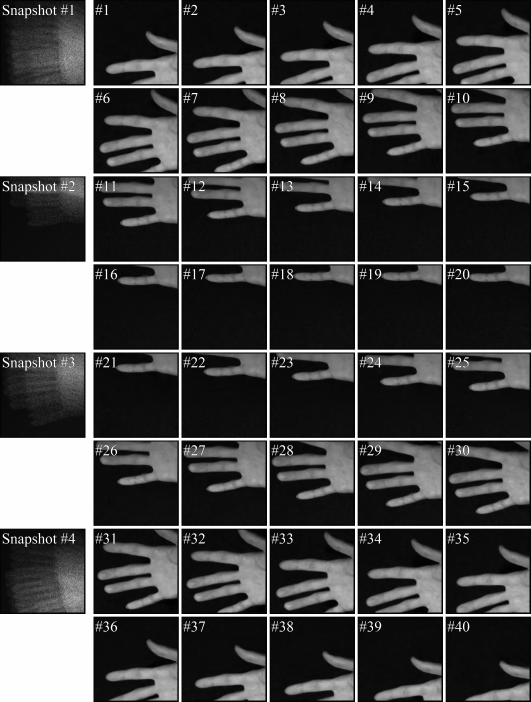}
    \caption{The first column shows snapshots from the Hand dataset~\cite{pnp} ($512\times512\times40$) in the \textbf{real benchmark} at a compression rate of $10$. Columns 2--6 show the corresponding reconstructed frames. Here, $\#$ denotes the frame number.
    }
    \label{fig:supp-hand}
\end{figure}

\clearpage

\begin{figure}[!htbp]
    \centering
    \includegraphics[width=1.0\columnwidth]{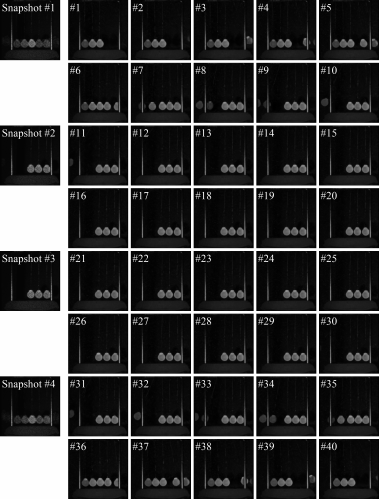}
    \caption{The first column shows snapshots from the Pendulum Ball dataset ($512\times512\times40$) in the \textbf{real benchmark} at a compression rate of $10$. Columns 2--6 show the corresponding reconstructed frames. Here, $\#$ denotes the frame number.
    }
    \label{fig:supp-pendulumball}
\end{figure}

\clearpage

\subsection{Real Benchmark at Compression Rate 50}\label{app:real-cr50}

\begin{figure}[!htbp]
    \centering
    \includegraphics[width=1.0\columnwidth,height=0.88\textheight,keepaspectratio]{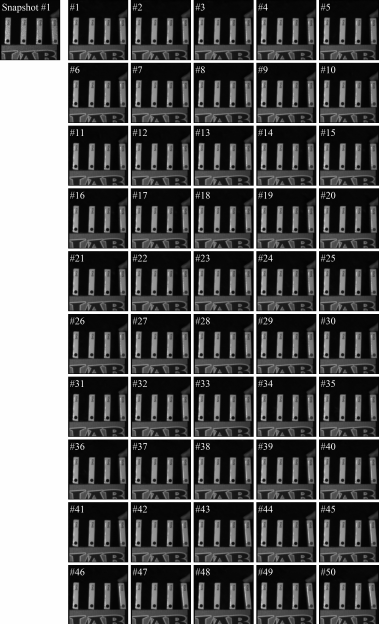}
    \caption{The first column shows snapshots from the Domino dataset~\cite{e2ecnn} ($512\times512\times300$) in the \textbf{real benchmark} at a compression rate of $50$. Columns 2--6 show the corresponding reconstructed frames. Here, $\#$ denotes the frame number.
    }
\label{fig:supp-domino-1-50}
\end{figure}

\clearpage

\begin{figure}[!htbp]
    \centering
    \includegraphics[width=1.0\columnwidth,height=0.92\textheight,keepaspectratio]{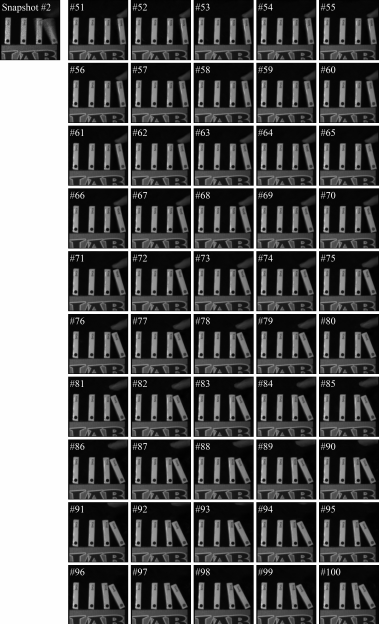}
    \caption{The first column shows snapshots from the Domino dataset~\cite{e2ecnn} ($512\times512\times300$) in the \textbf{real benchmark} at a compression rate of $50$. Columns 2--6 show the corresponding reconstructed frames. Here, $\#$ denotes the frame number.
    }
\label{fig:supp-domino-51-100}
\end{figure}

\clearpage

\begin{figure}[!htbp]
    \centering
    \includegraphics[width=1.0\columnwidth,height=0.92\textheight,keepaspectratio]{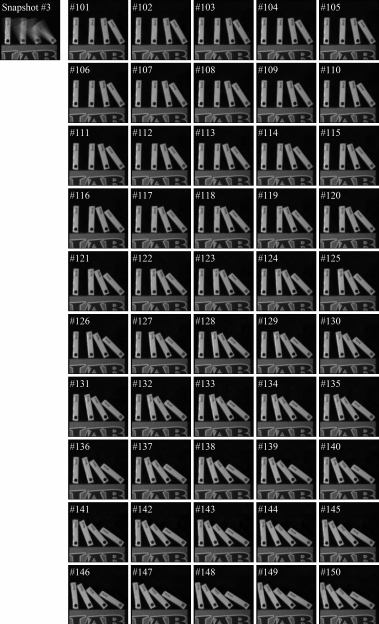}
    \caption{The first column shows snapshots from the Domino dataset~\cite{e2ecnn} ($512\times512\times300$) in the \textbf{real benchmark} at a compression rate of $50$. Columns 2--6 show the corresponding reconstructed frames. Here, $\#$ denotes the frame number.
    }
\label{fig:supp-domino-101-150}
\end{figure}

\clearpage

\begin{figure}[!htbp]
    \centering
    \includegraphics[width=1.0\columnwidth,height=0.92\textheight,keepaspectratio]{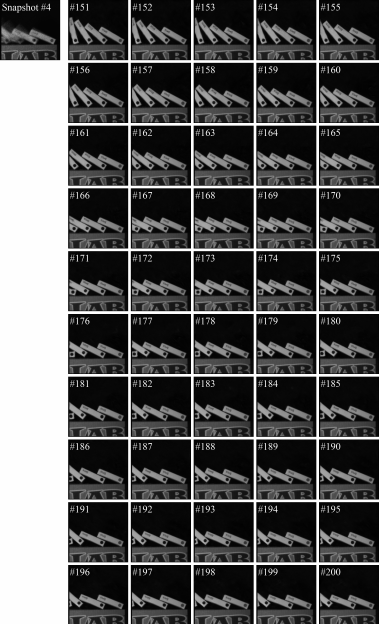}
    \caption{The first column shows snapshots from the Domino dataset~\cite{e2ecnn} ($512\times512\times300$) in the \textbf{real benchmark} at a compression rate of $50$. Columns 2--6 show the corresponding reconstructed frames. Here, $\#$ denotes the frame number.
    }
\label{fig:supp-domino-151-200}
\end{figure}

\clearpage

\begin{figure}[!htbp]
    \centering
    \includegraphics[width=1.0\columnwidth,height=0.92\textheight,keepaspectratio]{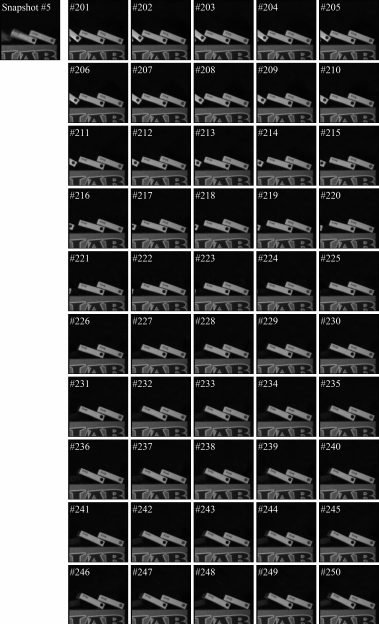}
    \caption{The first column shows snapshots from the Domino dataset~\cite{e2ecnn} ($512\times512\times300$) in the \textbf{real benchmark} at a compression rate of $50$. Columns 2--6 show the corresponding reconstructed frames. Here, $\#$ denotes the frame number.
    }
\label{fig:supp-domino-201-250}
\end{figure}

\clearpage

\begin{figure}[!htbp]
    \centering
    \includegraphics[width=1.0\columnwidth,height=0.92\textheight,keepaspectratio]{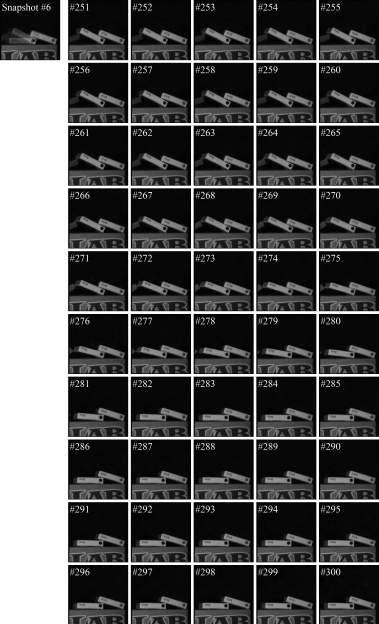}
    \caption{The first column shows snapshots from the Domino dataset~\cite{e2ecnn} ($512\times512\times300$) in the \textbf{real benchmark} at a compression rate of $50$. Columns 2--6 show the corresponding reconstructed frames. Here, $\#$ denotes the frame number.
    }
\label{fig:supp-domino-251-300}
\end{figure}

\clearpage

\begin{figure}[!htbp]
    \centering
    \includegraphics[width=1.0\columnwidth,height=0.92\textheight,keepaspectratio]{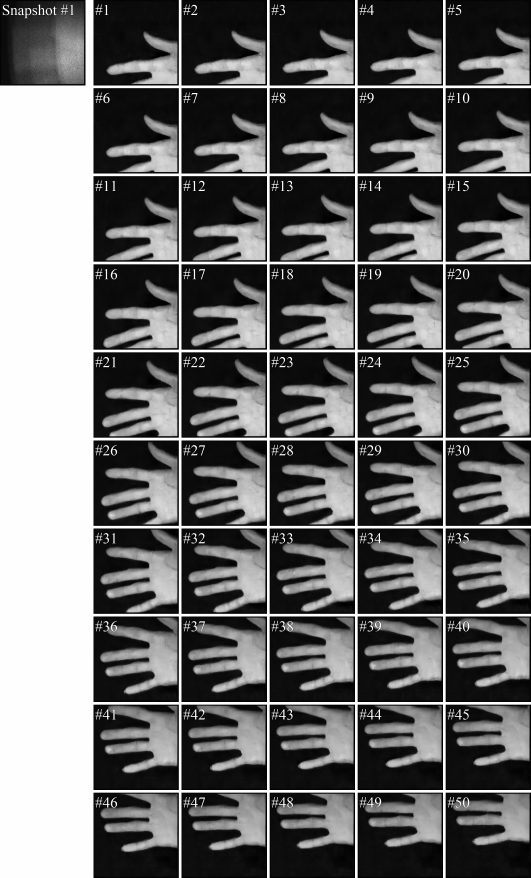}
    \caption{The first column shows snapshots from the Hand dataset~\cite{pnp} ($512\times512\times200$) in the \textbf{real benchmark} at a compression rate of $50$. Columns 2--6 show the corresponding reconstructed frames. Here, $\#$ denotes the frame number.
    }
\label{fig:supp-hand-1-50}
\end{figure}

\clearpage

\begin{figure}[!htbp]
    \centering
    \includegraphics[width=1.0\columnwidth,height=0.92\textheight,keepaspectratio]{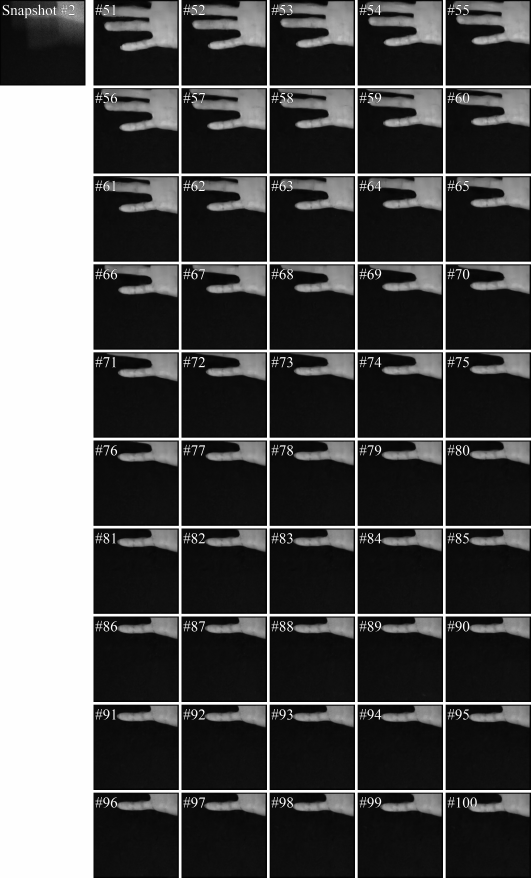}
    \caption{The first column shows snapshots from the Hand dataset~\cite{pnp} ($512\times512\times200$) in the \textbf{real benchmark} at a compression rate of $50$. Columns 2--6 show the corresponding reconstructed frames. Here, $\#$ denotes the frame number.
    }
\label{fig:supp-hand-51-100}
\end{figure}

\clearpage

\begin{figure}[!htbp]
    \centering
    \includegraphics[width=1.0\columnwidth,height=0.92\textheight,keepaspectratio]{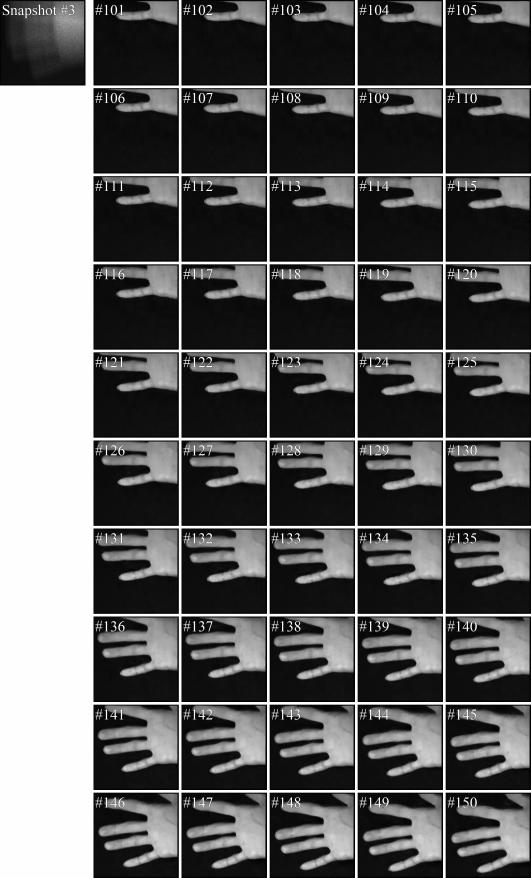}
    \caption{The first column shows snapshots from the Hand dataset~\cite{pnp} ($512\times512\times200$) in the \textbf{real benchmark} at a compression rate of $50$. Columns 2--6 show the corresponding reconstructed frames. Here, $\#$ denotes the frame number.
    }
\label{fig:supp-hand-101-150}
\end{figure}

\clearpage

\begin{figure}[!htbp]
    \centering
    \includegraphics[width=1.0\columnwidth,height=0.92\textheight,keepaspectratio]{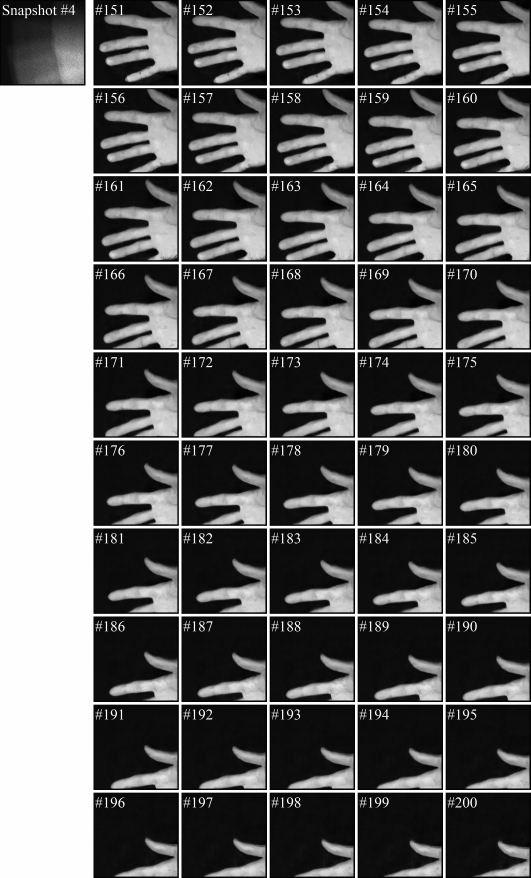}
    \caption{The first column shows snapshots from the Hand dataset~\cite{pnp} ($512\times512\times200$) in the \textbf{real benchmark} at a compression rate of $50$. Columns 2--6 show the corresponding reconstructed frames. Here, $\#$ denotes the frame number.
    }
\label{fig:supp-hand-151-200}
\end{figure}

\clearpage

\begin{figure}[!htbp]
    \centering
    \includegraphics[width=1.0\columnwidth,height=0.92\textheight,keepaspectratio]{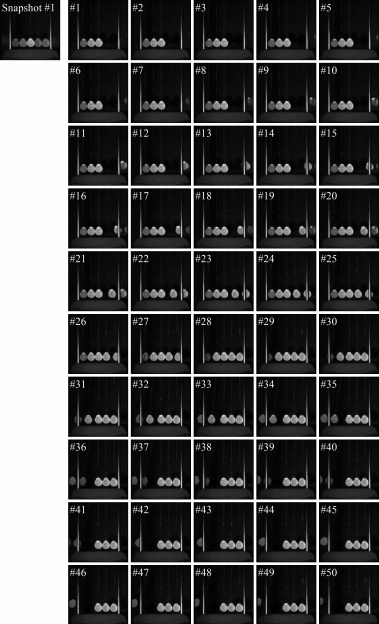}
    \caption{The first column shows snapshots from the Pendulum Ball dataset ($512\times512\times200$) in the \textbf{real benchmark} at a compression rate of $50$. Columns 2--6 show the corresponding reconstructed frames. Here, $\#$ denotes the frame number.
    }
\label{fig:supp-pendulumball-1-50}
\end{figure}

\clearpage

\begin{figure}[!htbp]
    \centering
    \includegraphics[width=1.0\columnwidth,height=0.92\textheight,keepaspectratio]{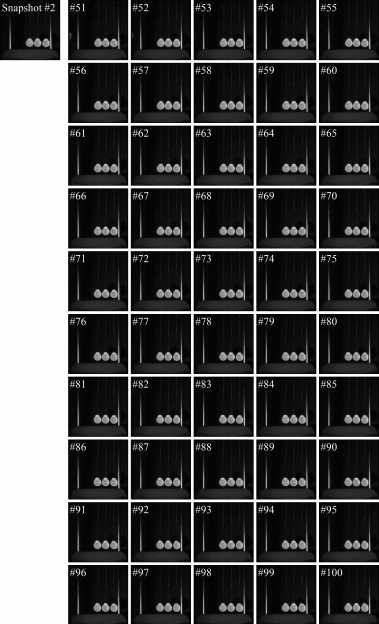}
    \caption{The first column shows snapshots from the Pendulum Ball dataset ($512\times512\times200$) in the \textbf{real benchmark} at a compression rate of $50$. Columns 2--6 show the corresponding reconstructed frames. Here, $\#$ denotes the frame number.
    }
\label{fig:supp-pendulumball-51-100}
\end{figure}

\clearpage

\begin{figure}[!htbp]
    \centering
    \includegraphics[width=1.0\columnwidth,height=0.92\textheight,keepaspectratio]{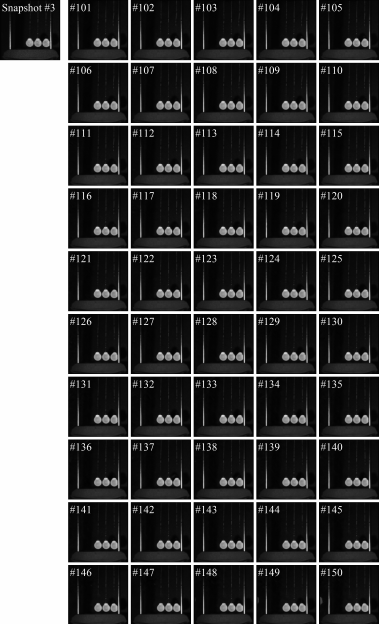}
    \caption{The first column shows snapshots from the Pendulum Ball dataset ($512\times512\times200$) in the \textbf{real benchmark} at a compression rate of $50$. Columns 2--6 show the corresponding reconstructed frames. Here, $\#$ denotes the frame number.
    }
\label{fig:supp-pendulumball-101-150}
\end{figure}

\clearpage

\begin{figure}[!htbp]
    \centering
    \includegraphics[width=1.0\columnwidth,height=0.92\textheight,keepaspectratio]{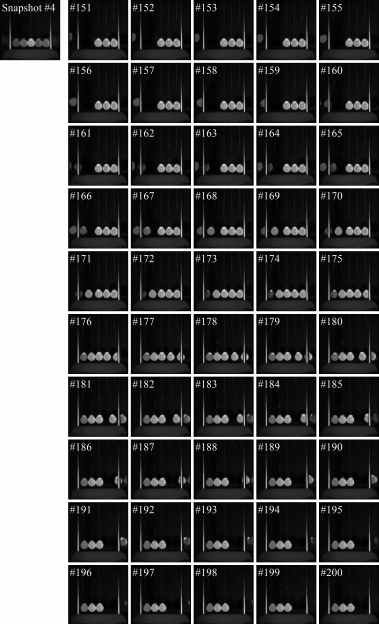}
    \caption{The first column shows snapshots from the Pendulum Ball dataset ($512\times512\times200$) in the \textbf{real benchmark} at a compression rate of $50$. Columns 2--6 show the corresponding reconstructed frames. Here, $\#$ denotes the frame number.
    }
\label{fig:supp-pendulumball-151-200}
\end{figure}

\clearpage

\begin{figure}[!htbp]
    \centering
    \includegraphics[width=1.0\columnwidth,height=0.92\textheight,keepaspectratio]{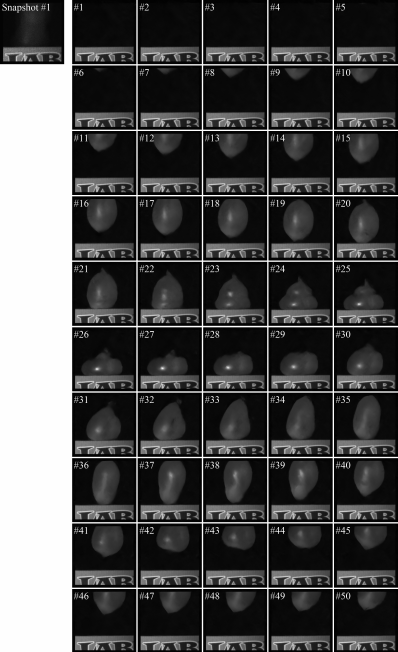}
    \caption{The first column shows snapshots from the Water Balloon dataset~\cite{e2ecnn} ($512\times512\times50$) in the \textbf{real benchmark} at a compression rate of $50$. Columns 2--6 show the corresponding reconstructed frames. Here, $\#$ denotes the frame number.
    }
\label{fig:supp-waterballoon-cr50}
\end{figure}

\clearpage

\subsection{Small-Scale Gray Comparison}\label{app:small-gray}

\clearpage

\subsection{Mid-Scale Color Comparison}\label{app:mid-color}

\subsection{Large-Scale Color Comparison}\label{app:large-color}

\clearpage

\subsection{Other Compression Rates}\label{app:other-cr}

The two real-benchmark checkpoints of SpinSCI-B are trained at compression
rates $10$ and $50$ (Appendices~\ref{app:real-cr10}--\ref{app:real-cr50}).
To test how far one checkpoint carries to other compression rates without
retraining, we simulate measurements of seven UVG videos~\cite{uvg}
(Beauty, Bosphorus, HoneyBee, Jockey, ReadySetGo, ShakeNDry, and
YachtRide; gray, $512\times512$, $600$ frames each except $300$ for
ShakeNDry) with the first $N$ frames of the real-benchmark mask, the mask
both checkpoints were trained with, for $N\in\{10,20,30,40,50\}$. Every
compression rate uses the same frames, except that CR $40$ drops the last
$20$ frames of ShakeNDry. Only the temporal relative-position bias depends
on the number of frames. At an untrained compression rate we keep the
learned bias of every frame displacement that occurs and set the bias of
displacements the checkpoint never saw to zero; this choice was fixed
before any run. Inference uses EMA weights in FP32, as for the real
benchmark. The baselines with official real-data checkpoints run under the
same protocol with their official code and real-data settings: STFormer-B,
released for CR $10$ only, whose temporal bias table is extended with the
same rule, and PnP-FastDVDNet, whose pretrained denoiser is not trained for
a specific compression rate. On the small-scale gray benchmark, this
evaluation reproduces the values of \Cref{tab:gray} for both baselines
within $0.18$~dB.

Among the methods with official real-data checkpoints, only SpinSCI-B
reconstructs CR $50$ well (\Cref{tab:app-other-cr}): its CR-$50$
checkpoint reaches $34.73$~dB, PnP-FastDVDNet reaches $22.56$~dB, and
STFormer-B releases no CR-$50$ checkpoint. Without retraining, the same
SpinSCI-B checkpoint loses only $0.03$ and $0.53$~dB at CR $40$ and $30$,
so one checkpoint trained at a long exposure also serves shorter ones; it
loses $1.90$~dB at CR $20$ and $4.40$~dB at CR $10$. The CR-$10$
checkpoint does not carry to longer exposures: it reaches $31.21$~dB at
CR $20$ and stays below $23$~dB from CR $30$ on. The CR-$10$ checkpoint of
STFormer-B reaches $34.37$~dB at CR $10$ and fails at every longer exposure
($10.52$~dB at CR $20$). A $20$-frame build of the model reproduces its CR-$10$ result on
Beauty, so the extended bias table is not the cause. PnP-FastDVDNet
degrades gradually, from $31.11$~dB at CR $10$ to $22.56$~dB at CR $50$.
With the better of its two checkpoints, SpinSCI-B stays $4.91$--$12.17$~dB
above PnP-FastDVDNet at every compression rate. The best quality therefore
needs one checkpoint per compression rate, as for STFormer~\cite{stformer},
MetaSCI~\cite{metasci} and RevSCI~\cite{revsci}. Resampling the bias table
of SpinSCI-B linearly instead changes its PSNR by $-1.18$ to $+1.16$~dB and
leaves these conclusions unchanged.

\begin{table}[H]
\centering
\caption{Average PSNR (dB) / SSIM at other compression rates, on seven UVG
videos (gray, $512\times512$) simulated with the real-benchmark mask, for
the real-benchmark checkpoints of SpinSCI-B and the baselines with official
real-data checkpoints. Parentheses: the compression rate each checkpoint
was trained at (bold). PnP-FastDVDNet is not trained for a specific
rate.}
\label{tab:app-other-cr}
\footnotesize
\setlength{\tabcolsep}{4pt}
\begin{tabular}{@{}lrrrrr@{}}
\toprule
Method & CR $10$ & CR $20$ & CR $30$ & CR $40$ & CR $50$ \\
\midrule
PnP-FastDVDNet & 31.11 / 0.859 & 27.92 / 0.799 & 25.86 / 0.746 & 24.48 / 0.710 & 22.56 / 0.633 \\
STFormer-B (CR $10$) & \textbf{34.37 / 0.912} & 10.52 / 0.621 & 5.23 / 0.335 & 2.08 / 0.198 & $-$0.19 / 0.122 \\
\midrule
SpinSCI-B (CR $10$) & \textbf{39.48 / 0.960} & 31.21 / 0.871 & 22.77 / 0.647 & 18.04 / 0.462 & 16.28 / 0.336 \\
SpinSCI-B (CR $50$) & 30.33 / 0.852 & 32.83 / 0.901 & 34.20 / 0.914 & 34.70 / 0.913 & \textbf{34.73 / 0.910} \\
\bottomrule
\end{tabular}
\end{table}

%% file: sections/related_work_extended.tex
\section{Extended Related Work}\label{app:related}

This section extends \Cref{sec:related_work}. For each line of work, it states how MC-IRB or SpinSCI differs.

\noindent\textbf{Unrolled Mean-Field Inference.} CRF-as-RNN~\cite{crfasrnn} and the Deep Parsing Network~\cite{dpn} unroll mean-field updates of conditional random fields as network layers for semantic segmentation. Their potentials act on per-pixel label distributions, and the unrolled updates produce the segmentation output. MC-IRB applies a fixed number of mean-field updates to one scalar latent spin per post-attention token. Its couplings are built from pre-softmax attention logits after the sensing penalty in Eq.~\eqref{eq:sensing-penalty}, row-centering, and symmetrization, and its magnetizations only set the strength of a residual feature correction. The block output therefore remains a feature map.

\noindent\textbf{Energy-Based Views of Attention.} Modern Hopfield networks~\cite{hopfield} interpret softmax attention as an update of a continuous associative-memory energy. The Energy Transformer~\cite{energytransformer} designs a network whose forward pass descends an energy over token representations, and deep equilibrium models~\cite{deq} solve for layer fixed points. In these models, the energy or fixed point governs the token representations themselves. In MC-IRB, the Ising energy in Eq.~\eqref{eq:ising-energy} is defined over auxiliary binary spins, and its mean-field magnetizations act only as gates. The attention output is computed as usual and remains the base of the residual update.

\noindent\textbf{Signed and Gated Attention.} Differential Transformer~\cite{difftransformer} subtracts two softmax attention maps, so that the effective attention weights can be negative. Gated Attention~\cite{gatedattention} multiplies the attention output by a sigmoid gate computed from each token alone. Both change how values are weighted or scaled. MC-IRB leaves the attention weights and values unchanged. Its signed couplings act between gates: through the mean-field updates, the gate of each token depends on the states of the other tokens in its group, and the couplings and fields are conditioned on the coding masks, the normalized initial estimate, and the exposure normalizer. In \Cref{tab:ablation_irb}, the field-only sigmoid gate and the $\mathbf{J}=\mathbf{0}$ control are the closest analogues of an independently gated output.

\noindent\textbf{Deep Unfolding for SCI.} Deep unfolding methods unroll an iterative solver and replace its prior step with a learned network, alternating data-fidelity updates with learned denoising. Examples include Deep Tensor ADMM-Net~\cite{admm}, Tensor FISTA-Net~\cite{fista}, the dense deep unfolding network of \citet{dun}, the unfolding framework of \citet{deepunf}, the convolution--Transformer mixture prior of \citet{ctm}, and the degradation-aware unfolding network of \citet{dadun}. Differential Unfolding~\cite{du} reduces cost by alternating full stages with lightweight stages that refine changes in intermediate representations. In these methods, the mask enters each stage through the measurement operator. SpinSCI is not unrolled. It uses the measurement model once for initialization and once for final data consistency, and additionally supplies the mask to every MC-IRB as a conditioning signal.

\noindent\textbf{Other Learned Video SCI.} BIRNAT~\cite{birnat} reconstructs frames with bidirectional recurrent networks and adversarial training, and RevSCI~\cite{revsci} uses reversible 3D convolutions to reduce training memory for large-scale inputs. MadyGraph~\cite{madygraph} passes messages among space--time pixels with a graph neural network whose neighbors are selected by motion-aware dynamic sampling. Its graph is built from features alone, whereas MC-IRB conditions its signed couplings on the masks through the sensing penalty in Eq.~\eqref{eq:sensing-penalty}. \citet{effipp} combine CNN and Transformer components for efficient large-scale reconstruction. EfficientSCI~\cite{efficient}, STFormer~\cite{stformer}, MambaSCI~\cite{mamba}, and HiSViT~\cite{hisvit} are compared in \Cref{sec:comparison}. In these networks, the mask mainly enters through the input formed from the measurement. SpinSCI also forms its input from the measurement and mask, and additionally conditions each refinement block on them.

\noindent\textbf{Mask-Conditioned Reconstruction.} We call a reconstruction network mask-conditioned when the coding masks condition its intermediate operations, so that one set of trained weights adapts to the mask supplied at test time without per-mask fine-tuning or mask optimization. Most prior video SCI methods use the mask in other ways: to form the network input, in the data-fidelity steps of deep unfolding, for per-mask parameter adaptation in MetaSCI~\cite{metasci}, or for joint learning of the mask and the reconstruction in Res2former~\cite{res2former}. In spectral SCI, MST~\cite{mst} derives a mask-guided modulation for its spectral-wise attention, and DAUHST~\cite{dauhst} estimates degradation parameters from the mask and measurement to guide its unfolding stages. In video SCI, the degradation-aware unfolding network of \citet{dadun} likewise uses the physical mask in the estimates that guide its iterations. These methods are mask-conditioned in the above sense, but their conditioning modulates individual features or stages rather than pairwise interactions. To our knowledge, SpinSCI is the first video SCI method in which the masks condition pairwise interactions inside a refinement module, through the sensing penalty in Eq.~\eqref{eq:sensing-penalty} and the sensing features in the unary fields. Removing these explicit mask channels lowers PSNR by $0.13$--$0.19$~dB on the training mask and by $0.30$--$0.38$~dB on an unseen binary mask, although the mask-dependent initialization and final data consistency are kept (\Cref{tab:ablation_mask}).

%% file: sections/hardware_cost.tex
\section{First-Order Hardware Cost Estimate}\label{app:hw-ppa}

We estimate what each method would cost on a dedicated accelerator under two cost models, reported separately. The first uses only published per-operation energies and areas at 45nm. The second is a cost model calibrated on a place-and-route implementation of an integer transformer accelerator at 22nm, with published DRAM constants. Both apply the same rules to every network and are first-order, so ratios between networks are more reliable than absolute values.

\noindent\textbf{Workload.} A PyTorch dispatch-level trace of each released model, at the resolutions of Tables~\ref{tab:gray}--\ref{tab:large}, counts every matrix product and convolution as multiply--accumulates (MACs), including attention products and padded positions, and one element-op per element of the largest operand of every other arithmetic operator. Unfused traffic assumes that each operator reads its inputs and writes its outputs once. The live peak, the largest total of live 16-bit tensors plus the weights in the reference operator order, is the on-chip memory that a forward pass needs without DRAM. Large scale uses 24 tiles of $272\times320$ for every method, as in \Cref{tab:large}; its memory figures are per tile. Costs are per reconstructed frame, one eighth of a measurement.

\noindent\textbf{Public 45nm model.} Following \citet{horowitz2014energy}, an FP16 MAC or element-op costs 1.5\,pJ; a 64-bit access costs 20\,pJ from a 32\,KB SRAM bank, 100\,pJ from a 1\,MB bank and 1.3\,nJ from DRAM; and SRAM leaks about 100\,mW per MB. An FP16 MAC unit occupies 3{,}000\,$\mu$m$^2$ \citep[45nm synthesis;][]{han2017efficient} and an SRAM bit cell 0.346\,$\mu$m$^2$ \citep{mistry2007nm45}. Every network runs on 1{,}024 MAC lanes at 1\,GHz with full utilization, and energy is action counts times per-action energies \citep{wu2019accelergy}. On chip, each 64-bit word of traffic costs one 1\,MB-bank access, the time is compute-bound, and the SRAM holds either the GPU peak memory of the tables at FP16, which illustrates the memory wall, or the live peak. With data in DRAM behind a 1\,MiB buffer, each word costs one DRAM and one buffer access, and the time is the larger of the compute time and the transfer time over one 64-bit DDR4-2400 channel at 19.2\,GB/s \citep{jedec2012ddr4}. A 16-bit DRAM word then costs 217 FP16 MACs, close to the 200$\times$ of a 65nm accelerator \citep{chen2016eyeriss}.

\begin{table}[H]
\centering
\caption{Public 45nm model, per reconstructed frame, for SpinSCI-B (S) and HiSViT$_9$ (H). Saving: how many times less SpinSCI-B needs than HiSViT$_9$. Middle rows: all data on chip. Bottom rows: data in DRAM, where both designs occupy 5.97\,mm$^2$.}
\label{tab:hw-45nm}
\footnotesize
\setlength{\tabcolsep}{3.6pt}
\begin{tabular}{@{}l rrr rrr rrr@{}}
\toprule
& \multicolumn{3}{c}{Small-scale gray} & \multicolumn{3}{c}{Mid-scale color} & \multicolumn{3}{c}{Large-scale color} \\
\cmidrule(lr){2-4}\cmidrule(lr){5-7}\cmidrule(l){8-10}
& S & H & Saving & S & H & Saving & S & H & Saving \\
\midrule
MACs ($10^{9}$) & 141 & 203 & 1.44 & 565 & 814 & 1.44 & 4{,}510 & 6{,}830 & 1.51 \\
Traffic, unfused (GB) & 10.9 & 16.0 & 1.47 & 42.8 & 64.0 & 1.49 & 346 & 748 & 2.16 \\
\midrule
SRAM at GPU peak (GiB) & 0.903 & 2.39 & 2.65 & 3.49 & 9.50 & 2.72 & 1.18 & 3.18 & 2.68 \\
SRAM at live peak (GiB) & 0.628 & 1.24 & 1.97 & 2.42 & 4.89 & 2.02 & 0.818 & 1.74 & 2.13 \\
Area at live peak (mm$^2$) & 1{,}870 & 3{,}680 & 1.97 & 7{,}190 & 14{,}500 & 2.02 & 2{,}430 & 5{,}180 & 2.13 \\
Energy at live peak (J) & 9.30 & 25.9 & 2.78 & 139 & 404 & 2.91 & 383 & 1{,}230 & 3.20 \\
\midrule
Time with DRAM (s) & 0.568 & 0.833 & 1.47 & 2.23 & 3.33 & 1.49 & 18.0 & 38.9 & 2.16 \\
Energy with DRAM (J) & 2.07 & 3.03 & 1.46 & 8.14 & 12.1 & 1.49 & 65.6 & 138 & 2.10 \\
\bottomrule
\end{tabular}
\end{table}

\noindent\textbf{SpinSCI-B versus HiSViT$_9$, 45nm.} SpinSCI-B needs 1.44 times fewer MACs than HiSViT$_9$ at small and mid scale (\Cref{tab:hw-45nm}), more than the FLOP ratio of 1.27 in the tables: the trace adds the attention products that the module-level counter omits, and it charges SpinSCI-B's stride-2 transposed convolution at its actual cost. With data in DRAM, both networks are DRAM-bound at about 3.6\,W, and SpinSCI-B needs 1.46--1.49 times less energy and time than HiSViT$_9$. On chip, both networks hit the memory wall: even the live peak is 0.63--2.4\,GiB for SpinSCI-B and 1.2--4.9\,GiB for HiSViT$_9$. Leakage is then at least 96\% of the energy, and SpinSCI-B's saving without it is 1.45--1.77 times. At large scale, its saving rises to 2.10 times for energy and 2.16 times for time, mainly because HiSViT$_9$ pads the $136\times160$ feature map of each tile to $256\times256$ in its pooled attention; at the mid-scale cost per pixel, it is 1.49 times. SpinSCI-B's DRAM energy saving stays within 1.45--1.50 times at small and mid scale and 1.93--2.14 times at large scale under FP32 or INT8, 2.6\,nJ DRAM accesses, MAC-operator traffic only, MAC-only time, module-level FLOPs or 7nm energies \citep{jouppi2021tpuv4i}.

\noindent\textbf{22nm calibrated model.} The engine has 4{,}096 MAC lanes and spends 1.68\,pJ per executed MAC (sensitivity band 1.03--1.91\,pJ); on-chip memory, DRAM and leakage are charged separately. LPDDR4 costs 6.6\,pJ/bit \citep{gans2016lpddr} plus 2.7\,pJ/bit for the host interface \citep{chowdhury2025phy}, with background and refresh power from a vendor data sheet \citep{micron2023lpddr4}. Stacked DRAM costs 0.49 and 0.88\,pJ/bit to read and write \citep{wang2023sedram,bai2020sedram}. The model fuses each network into passes over the frame, keeps rows of activations in an on-chip line buffer and sends only pass boundaries to DRAM. Schedule (a) runs one pass per block with LPDDR4; (b) uses a smaller buffer, two passes per transformer block and stacked DRAM; the DRAM-free (c) and (d) keep the whole frame or 256-pixel crops on chip. Each schedule is built by hand for both networks under the same rules, and a design fits if its logic die stays within 164\,mm$^2$ (a $12.8\times12.8$\,mm die) and a power density of 30\,mW/mm$^2$. HiSViT$_9$'s frame-global channel attention makes each pass store a block's input and its pre-gate output. Its cosine-attention normalization and channel-attention mean use the normalization unit, its sigmoid gates and softmax the lookup-table unit, and its layout operators address generation. Both networks get SpinSCI-B's measured array utilization of 66.5\%, which is optimistic for HiSViT$_9$.

\begin{table}[t]
\centering
\caption{22nm calibrated model, per reconstructed frame, at 1.68\,pJ/MAC. Saving: how many times less SpinSCI-B needs than HiSViT$_9$; band: energy saving over 1.03--1.91\,pJ/MAC. Area: logic die; $^\dagger$: exceeds the 164\,mm$^2$ die limit, so the design does not fit. Schedule (d) applies at mid scale only.}
\label{tab:hw-22nm}
\footnotesize
\setlength{\tabcolsep}{4pt}
\begin{tabular}{@{}ll rr rrr rr@{}}
\toprule
& & \multicolumn{2}{c}{Energy (J)} & \multicolumn{3}{c}{Saving} & \multicolumn{2}{c}{Area (mm$^2$)} \\
\cmidrule(lr){3-4}\cmidrule(lr){5-7}\cmidrule(l){8-9}
Scale & Schedule & S & H & energy & band & time & S & H \\
\midrule
Small & (a) Streaming, LPDDR4 & 0.249 & 0.374 & 1.50 & 1.50--1.53 & 1.46 & 39.3 & 74.5 \\
 & (b) Streaming, stacked DRAM & 0.235 & 0.350 & 1.49 & 1.48--1.51 & 1.41 & 27.9 & 56.9 \\
 & (c) DRAM-free, whole frame & 0.253 & 0.369 & 1.46 & 1.44--1.53 & 1.35 & 110 & 302$^\dagger$ \\
\midrule
Mid & (a) Streaming, LPDDR4 & 1.01 & 1.53 & 1.53 & 1.52--1.56 & 1.46 & 59.3 & 128 \\
 & (b) Streaming, stacked DRAM & 0.948 & 1.46 & 1.54 & 1.53--1.59 & 1.41 & 38.4 & 95.0 \\
 & (c) DRAM-free, whole frame & 1.13 & 1.91 & 1.69 & 1.65--1.87 & 1.35 & 392$^\dagger$ & 1{,}140$^\dagger$ \\
 & (d) DRAM-free, 256-px crops & 1.01 & 1.48 & 1.46 & 1.45--1.53 & 1.35 & 112 & 304$^\dagger$ \\
\midrule
Large & (a) Streaming, LPDDR4 & 7.97 & 12.6 & 1.58 & 1.57--1.61 & 1.53 & 44.3 & 88.0 \\
 & (b) Streaming, stacked DRAM & 7.53 & 11.8 & 1.57 & 1.56--1.59 & 1.48 & 30.5 & 66.4 \\
 & (c) DRAM-free, whole frame & 8.20 & 12.8 & 1.56 & 1.55--1.65 & 1.42 & 141 & 395$^\dagger$ \\
\bottomrule
\end{tabular}
\end{table}

\noindent\textbf{SpinSCI-B versus HiSViT$_9$, 22nm.} With the streaming schedules (a) and (b), SpinSCI-B needs 1.49--1.58 times less energy and 1.41--1.53 times less time than HiSViT$_9$, and both draw 2.3--2.7\,W (\Cref{tab:hw-22nm}). SpinSCI-B's energy saving stays within 1.48--1.61 times over the engine-energy band, and charging HiSViT$_9$'s extra normalization and lookup-table operations at their unit cost raises it to 1.53--1.65 times for (a). SpinSCI-B's area saving of 1.9--2.5 times depends on the line buffer we derived for HiSViT$_9$; with SpinSCI-B's buffer, it falls to 1.02--1.05 times and the energy saving stays at 1.46--1.55 times. Without DRAM, SpinSCI-B's whole frame fits at small scale and per large-scale tile but not at mid scale, and HiSViT$_9$'s fits nowhere. At mid scale, 256-pixel crops fit for SpinSCI-B (112\,mm$^2$) but not for HiSViT$_9$ (304\,mm$^2$), with 1.46 times less energy for SpinSCI-B; measured separately, such crops cost 0.04\,dB of PSNR for SpinSCI-B and 0.05\,dB for HiSViT$_9$. For SpinSCI-B, the DRAM-free designs cost 1--12\% more energy than schedule (a). Sending every operator's traffic to LPDDR4, as the public model does, gives SpinSCI-B savings of 1.47, 1.56 and 2.02 times: most of the public model's 2.10 at large scale comes from unfused traffic around HiSViT$_9$'s padded attention.

\begin{figure}[t]
\centering
\includegraphics[width=\linewidth]{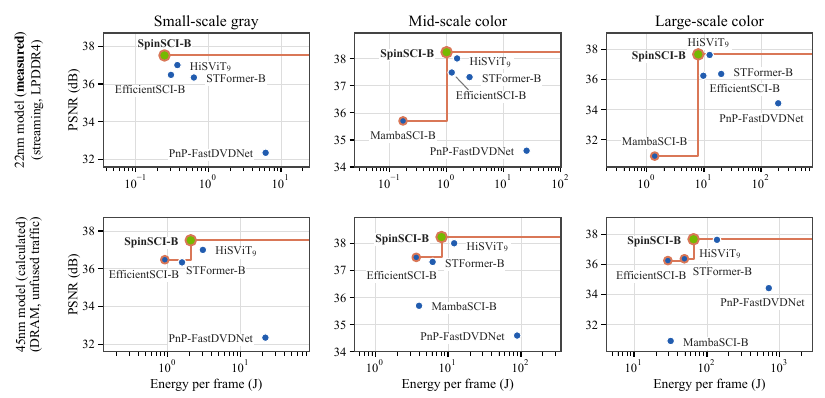}
\caption{PSNR versus estimated energy per frame: 22nm model, calibrated on measurements of a place-and-route implementation, schedule (a) (top); 45nm model, calculated from published per-operation energies, data in DRAM, with unfused traffic that works against SpinSCI-B relative to the cheaper methods (bottom). Green: SpinSCI-B; blue: the other methods. Orange line and rings: the Pareto front, on which no other method has both lower energy and higher PSNR.}
\label{fig:hw-frontier}
\end{figure}

\begin{table}[t]
\centering
\caption{All methods, per frame; PSNR from Tables~\ref{tab:gray}--\ref{tab:large}. 45nm: data in DRAM; live peak. 22nm: schedule (a); DRAM-free capacity ($^\dagger$: no fit). $\times$S: ratio to SpinSCI-B. $^\ast$: Pareto-optimal.}
\label{tab:hw-frontier}
\footnotesize
\setlength{\tabcolsep}{4pt}
\begin{tabular}{@{}ll rr rrr rrr@{}}
\toprule
& & & & \multicolumn{3}{c}{Public 45nm model} & \multicolumn{3}{c}{22nm calibrated model} \\
\cmidrule(lr){5-7}\cmidrule(l){8-10}
Scale & Method & PSNR & MACs & Energy & $\times$S & Live peak & Energy & $\times$S & DRAM-free \\
& & (dB) & ($10^{9}$) & (J) & & (GiB) & (J) & & (MiB) \\
\midrule
Small & PnP-FastDVDNet & 32.35 & 3{,}700 & 21.8 & 10.53 & 0.191 & 6.15 & 24.70 & 20.0 \\
 & EfficientSCI-B & 36.48 & 178 & 0.914$^\ast$ & 0.44 & 0.266 & 0.305 & 1.23 & 26.8 \\
 & STFormer-B & 36.34 & 372 & 1.57 & 0.76 & 0.568 & 0.632 & 2.54 & 57.1 \\
 & HiSViT$_9$ & 37.00 & 203 & 3.03 & 1.46 & 1.24 & 0.374 & 1.50 & 127$^\dagger$ \\
 & SpinSCI-B & 37.51 & 141 & 2.07$^\ast$ & 1.00 & 0.628 & 0.249$^\ast$ & 1.00 & 65.2 \\
\midrule
Mid & PnP-FastDVDNet & 34.60 & 15{,}000 & 88.9 & 10.92 & 1.01 & 25.3 & 25.12 & 109 \\
 & EfficientSCI-B & 37.49 & 714 & 3.66$^\ast$ & 0.45 & 1.02 & 1.24 & 1.23 & 109$^\dagger$ \\
 & MambaSCI-B & 35.70 & 102 & 3.99 & 0.49 & 1.33 & 0.174$^\ast$ & 0.17 & 143$^\dagger$ \\
 & STFormer-B & 37.32 & 1{,}470 & 6.13 & 0.75 & 2.07 & 2.54 & 2.52 & 221$^\dagger$ \\
 & HiSViT$_9$ & 38.01 & 814 & 12.1 & 1.49 & 4.89 & 1.53 & 1.53 & 517$^\dagger$ \\
 & SpinSCI-B & 38.24 & 565 & 8.14$^\ast$ & 1.00 & 2.42 & 1.01$^\ast$ & 1.00 & 261$^\dagger$ \\
\midrule
Large & PnP-FastDVDNet & 34.43 & 120{,}000 & 709 & 10.80 & 0.353 & 199 & 25.04 & 37.8 \\
 & EfficientSCI-B & 36.26 & 5{,}690 & 29.1$^\ast$ & 0.44 & 0.348 & 9.78 & 1.23 & 36.0 \\
 & MambaSCI-B & 30.92 & 813 & 31.8 & 0.48 & 0.453 & 1.38$^\ast$ & 0.17 & 47.4 \\
 & STFormer-B & 36.38 & 11{,}800 & 49.2$^\ast$ & 0.75 & 0.718 & 20.1 & 2.52 & 73.9 \\
 & HiSViT$_9$ & 37.64 & 6{,}830 & 138 & 2.10 & 1.74 & 12.6 & 1.58 & 184$^\dagger$ \\
 & SpinSCI-B & 37.68 & 4{,}510 & 65.6$^\ast$ & 1.00 & 0.818 & 7.97$^\ast$ & 1.00 & 86.7 \\
\bottomrule
\end{tabular}
\end{table}

\noindent\textbf{All methods.} \Cref{tab:hw-frontier} and \Cref{fig:hw-frontier} cost every method of Tables~\ref{tab:gray}--\ref{tab:large} under both models. On the 22nm model, networks without a hand-built schedule stream through DRAM once per block with SpinSCI-B's line buffer scaled to their widest tensor, and their DRAM-free capacity is their live peak scaled by SpinSCI-B's ratio of planned to live memory. The public model charges every operator's traffic to DRAM, and SpinSCI-B moves 3.6 times more bytes per MAC than EfficientSCI-B (0.076 against 0.021 at mid scale), so EfficientSCI-B, MambaSCI-B and STFormer-B cost 0.44--0.76 times its energy. This unfused traffic works against SpinSCI-B: counting only the traffic of matrix products and convolutions, as if all other operators were fused into them, raises these three methods to 0.63--1.19 times SpinSCI-B's energy, while HiSViT$_9$ changes from 1.46--2.10 to 1.48--1.93 times. On the 22nm model, energy follows MACs: only MambaSCI-B costs less than SpinSCI-B, at 0.17 times, with its selective scan charged as MACs for lack of a matching unit, which favors it. EfficientSCI-B costs 1.23 times as much as SpinSCI-B, HiSViT$_9$ 1.50--1.58 times, STFormer-B 2.5 times and PnP-FastDVDNet 25 times; DU-9stg, which is not in the tables, costs 2.7--2.9 and 1.5--1.6 times as much under the two models. SpinSCI-B is on the PSNR--energy frontier in every panel because it has the highest PSNR. The cheaper frontier points are EfficientSCI-B, and STFormer-B at large scale, under the public model, and MambaSCI-B, 2.5--6.8\,dB lower in PSNR, under the 22nm model. Without DRAM, SpinSCI-B needs more on-chip memory than EfficientSCI-B, MambaSCI-B, STFormer-B and PnP-FastDVDNet under both models, and less than HiSViT$_9$.

\noindent\textbf{Measured GPU energy.} \Cref{tab:gpu-energy} reports measured energy per reconstructed frame on the GH200 node used for \Cref{tab:gray,tab:color,tab:large}, with the same inputs, adapters and FP32 eager settings; the power limit (700\,W enforced) and clocks are left unchanged. The only NVML energy counter on this system covers the whole Grace Hopper module, including the CPU, so we integrate the GPU-level instantaneous power, sampled every 20\,ms, over timed blocks of at least 20\,s after warm-up; integrating the module-level power in the same way reproduces the module counter to within 1.3\%. Net values subtract the idle power of the GPU with the model loaded, measured over 30\,s before each block. Measured SpinSCI-B and HiSViT$_9$ latencies are within 2.4\% of the tables. SpinSCI-B uses 1.16--1.73$\times$ less energy per frame than HiSViT$_9$ (1.15--1.67$\times$ net); this saving is smaller than its 1.20--1.96$\times$ speedup because SpinSCI-B draws more power while it runs. EfficientSCI-B, STFormer-B and MambaSCI-B use 0.35--0.84 times the energy of SpinSCI-B, at 0.75--6.8\,dB lower PSNR. BF16 or FP16 autocast lowers the energy of SpinSCI-B by 29--36\% at a mean-PSNR cost of at most 0.01~dB, and that of HiSViT$_9$ by 20--27\% at up to 0.12~dB.

\begin{table}[!htbp]
\centering
\caption{Measured GPU energy per reconstructed frame (J) on an NVIDIA GH200, median of three runs, each in a fresh process: GPU-level energy, with the value net of idle power in parentheses. FP32 PSNR (dB) is from \Cref{tab:gray,tab:color,tab:large}; BF16/FP16 PSNR is measured with each method's evaluation code under \texttt{torch.autocast}. Large-scale energy covers all $24$ tiles of a $1080\times1920$ snapshot. The last rows give SpinSCI-B's energy saving: how many times less energy it uses than HiSViT$_9$.}
\label{tab:gpu-energy}
\footnotesize
\setlength{\tabcolsep}{3.5pt}
\begin{tabular}{@{}llrrrrrr@{}}
\toprule
 & & \multicolumn{2}{c}{Small-scale gray} & \multicolumn{2}{c}{Mid-scale color} & \multicolumn{2}{c}{Large-scale color} \\
\cmidrule(lr){3-4}\cmidrule(lr){5-6}\cmidrule(l){7-8}
Method & Prec. & PSNR & J/frame & PSNR & J/frame & PSNR & J/frame \\
\midrule
PnP-FastDVDNet & FP32 & 32.35 & 90.0 (68.7) & 34.60 & 977 (297) & 34.43 & 7{,}470 (2{,}329) \\
EfficientSCI-B & FP32 & 36.48 & 3.95 (3.13) & 37.49 & 17.6 (14.1) & 36.26 & 128 (102) \\
STFormer-B & FP32 & 36.34 & 6.95 (5.58) & 37.32 & 30.4 (24.3) & 36.38 & 218 (175) \\
MambaSCI-B & FP32 & -- & -- & 35.70 & 38.2 (21.8) & 30.92 & 282 (155) \\
HiSViT$_9$ & FP32 & 37.00 & 13.2 (10.2) & 38.01 & 57.7 (45.0) & 37.64 & 580 (444) \\
SpinSCI-B & FP32 & 37.51 & 10.4 (8.23) & 38.24 & 49.9 (39.3) & 37.68 & 334 (266) \\
\midrule
HiSViT$_9$ & BF16 & 36.88 & 9.64 (7.03) & 37.90 & 43.5 (32.7) & 37.55 & 458 (333) \\
HiSViT$_9$ & FP16 & 36.99 & 9.82 (7.22) & 38.01 & 44.2 (33.4) & 37.64 & 463 (338) \\
SpinSCI-B & BF16 & 37.50 & 6.70 (5.08) & 38.24 & 34.4 (26.4) & 37.67 & 217 (165) \\
SpinSCI-B & FP16 & 37.51 & 6.89 (5.27) & 38.24 & 35.5 (27.3) & 37.68 & 222 (171) \\
\midrule
\multicolumn{2}{@{}l}{SpinSCI-B saving, FP32} &  & 1.27$\times$ (1.24$\times$) &  & 1.16$\times$ (1.15$\times$) &  & 1.73$\times$ (1.67$\times$) \\
\multicolumn{2}{@{}l}{SpinSCI-B saving, BF16} &  & 1.44$\times$ (1.39$\times$) &  & 1.26$\times$ (1.24$\times$) &  & 2.11$\times$ (2.01$\times$) \\
\multicolumn{2}{@{}l}{SpinSCI-B saving, FP16} &  & 1.42$\times$ (1.37$\times$) &  & 1.25$\times$ (1.22$\times$) &  & 2.08$\times$ (1.98$\times$) \\
\bottomrule
\end{tabular}
\end{table}

\noindent\textbf{Reduced precision.} Both cost models assume reduced-precision arithmetic, so we measured its effect on accuracy (\Cref{tab:hw-precision}). Every multiply--accumulate operand of both networks, attention included, is rounded to the listed format; accumulation and every mask-dependent input stay in FP32, and no setting is tuned per network. BF16, FP16 and W12A16 with min--max ranges change PSNR by at most 0.02\,dB, and SpinSCI-B stays ahead of HiSViT$_9$ in every condition. Eight-bit post-training quantization with per-tensor activation ranges costs both networks 0.8--1.8\,dB, less for SpinSCI-B in all four cases, so neither network tolerates this plain eight-bit scheme.

\makeatletter\setlength{\@fptop}{0pt}\makeatother
\begin{table}[t]
\centering
\caption{Reconstruction PSNR (dB) at reduced precision. All multiply--accumulate operands (weights and activations of every convolution, linear layer and matrix product, including attention) are rounded to the listed format; accumulation and all other operators stay FP32, and every mask-dependent input stays FP32. W8A8 and W12A16 use per-channel symmetric weights and per-tensor activation ranges calibrated on four DAVIS~2017 training clips. $\Delta$: change from FP32. H$-$S: HiSViT$_9$ minus SpinSCI-B. Autocast rows run the released code under GPU mixed precision.}
\label{tab:hw-precision}
\footnotesize
\setlength{\tabcolsep}{3.2pt}
\begin{tabular}{@{}l rrrrr rrrrr@{}}
\toprule
& \multicolumn{5}{c}{Small-scale gray} & \multicolumn{5}{c}{Mid-scale color} \\
\cmidrule(lr){2-6}\cmidrule(l){7-11}
Format & S & $\Delta$S & H & $\Delta$H & H$-$S & S & $\Delta$S & H & $\Delta$H & H$-$S \\
\midrule
FP32 & 37.51 & -- & 37.00 & -- & $-$0.51 & 38.24 & -- & 38.01 & -- & $-$0.23 \\
BF16 & 37.49 & $-$0.02 & 36.98 & $-$0.02 & $-$0.51 & 38.22 & $-$0.02 & 38.00 & $-$0.01 & $-$0.22 \\
FP16 & 37.51 & 0.00 & 37.00 & 0.00 & $-$0.51 & 38.24 & 0.00 & 38.01 & 0.00 & $-$0.23 \\
W12A16, min--max & 37.51 & 0.00 & 36.99 & $-$0.01 & $-$0.52 & 38.24 & 0.00 & 38.01 & 0.00 & $-$0.23 \\
W12A16, 99.99th pct. & 36.28 & $-$1.23 & 35.92 & $-$1.08 & $-$0.36 & 38.15 & $-$0.09 & 37.45 & $-$0.56 & $-$0.70 \\
W8A8, min--max & 36.64 & $-$0.87 & 35.58 & $-$1.42 & $-$1.06 & 36.99 & $-$1.25 & 36.51 & $-$1.50 & $-$0.48 \\
W8A8, 99.99th pct. & 35.94 & $-$1.57 & 35.19 & $-$1.81 & $-$0.75 & 37.41 & $-$0.83 & 36.77 & $-$1.24 & $-$0.64 \\
\midrule
BF16 autocast & 37.50 & $-$0.01 & 36.88 & $-$0.12 & $-$0.62 & 38.24 & 0.00 & 37.90 & $-$0.11 & $-$0.34 \\
FP16 autocast & 37.51 & 0.00 & 36.99 & $-$0.01 & $-$0.52 & 38.24 & 0.00 & 38.01 & 0.00 & $-$0.23 \\
\bottomrule
\end{tabular}
\end{table}

\noindent\textbf{Limitations.} The public model's traffic comes from unfused reference code, including layout copies that hardware would replace by address generation, and the live peak assumes the reference operator order without recomputation. Both models give every network the same array utilization, which ignores how well each network maps onto a MAC array and favors the other networks. MambaSCI-B's selective scan is costed as one fused operator; its reference implementation, which stores every scan step, would cost 3.6 times SpinSCI-B under the public model. Single literature values replace a circuit-level memory model such as CACTI \citep{muralimanohar2007cacti}. None of these choices is tuned per network, but each can change the absolute values substantially.